\documentclass[10pt,twocolumn,letterpaper]{article}

\usepackage{iccv}
\usepackage{times}
\usepackage{epsfig}
\usepackage{graphicx}
\usepackage{amsmath}
\usepackage{enumitem}
\usepackage{amssymb}
\usepackage{caption}
\usepackage{subcaption}

\usepackage[pagebackref=true,breaklinks=true,letterpaper=true,colorlinks,bookmarks=false]{hyperref}

 \iccvfinalcopy 

\def\iccvPaperID{****} 

\ificcvfinal\fi
\begin{document}

\title{Symbolic Segmentation Using Algorithm Selection}

\author{Martin Lukac\\
SST\\
Nazarbayev University, Astana, Kazakhstan\\
{\tt\small martin.lukac@nu.edu.kz}
\and
Kamila Abdiyeva\\
SST\\
Nazarbayev University, Astana, Kazakhstan\\
{\tt\small kabdiyeva@nu.edu.kz}
\and
Michitaka Kameyama\\
GSIS\\
Tohoku University, Sendai, Japan\\
{\tt\small kameyama@ecei.tohoku.ac.jp}
}

\maketitle

\begin{abstract}
In this paper we present an alternative approach to symbolic segmentation; instead of implementing a new method we approach symbolic segmentation as an algorithm selection problem. That is, let there be $n$ available algorithms for symbolic segmentation, a selection mechanism forms a set of input features and image attributes and selects on a case by case basis the best algorithm. The selection mechanism is demonstrated from within an algorithm framework where the selection is done in a set of various algorithm networks. Two sets of experiments are performed and in both cases we demonstrate that the algorithm selection allows to increase the result of the symbolic segmentation by a considerable amount. 
\end{abstract}

\section{Introduction}

The research field of computer vision contains currently several very hard open issues. One of the problems being investigated is the problem of the symbolic segmentation; in this task the algorithm must segment images into meaningful regions and then detect objects present in the image. Both segmentation and object recognition have been extensively studied using various approaches. For instance, for segmentation in various contexts several dedicated resources exists~\cite{martin:01,gelasca:08,everingham:10}. Similarly algorithms for various contexts have been developed such as for natural images~\cite{malik:99,shi:00,arbalez:06,maire:08}, for medical images~\cite{sharma:10,lathen:10,mharib:12,ali:14} or for biological images~\cite{ali:11,meijering:12}. The object recognition are received even more attention due to very high interest in object recognition from the industry. Some of the recent approach to object recognition and detection include~\cite{lowe:99,heikkila:04,jung:06,felzenswalb:10,ciresan:12}. 

The combination of both segmentation and recognition is however more difficult and relatively smaller number of studies and approaches have been proposed.For instance semantic segmentation has been implemented as a combination of segmentation and recognition~\cite{carreira:12}, probabilistic models~\cite{tu:02,ladicky:10}, convolutional networks~\cite{bharath:14} or other approaches for either specific conditions~\cite{perera:06}, a unified framework~\cite{li:09} or interleaved recognition and segmentation~\cite{leibe:08}. Some of the main difficulties of semantic segmentation are:
\begin{enumerate}[label=\alph*]
	\item The segmentation by humans depends on recognition and higher level information~\cite{zavitz:14}
	\item The recognition is directly depending on features and regions from which the features are extracted.
	\item The context of the image strongly modulate segmentation and object recognition.
\end{enumerate}
Consequently the symbolic segmentation is very complex due to the mutual influences of recognition and segmentation and the designed algorithms have generally high specificity to some particular features or context. 

As can be seen in computer science and other fields requiring algorithms it happens very often that several algorithms are implemented to solve similar or same problem in some varying contexts, environments or different types of inputs. The reason for such diversity and specificity is the fact that real-world problems are much more complex and dynamical than the current state of art software and hardware can handle. Consequently several approaches used the algorithm selection approach to improve the algorithms for various problems.

In this paper we propose the algorithm selection approach to the problem of symbolic segmentation. We base our work on the previously proposed platform for algorithm selection in~\cite{lukac:13}. We show that using algorithm selection and high level reasoning about the results of algorithm processing allows to iteratively improve result of semantic segmentation. We analyze two different approaches for algorithm selection using either Bayesian Network (BN) or Support Vector Machine (SVM). The main contributions of this paper are:
\begin{enumerate}
	\item Analysis of an iterative algorithm selection framework in the context of symbolic segmentation
	\item Evaluation of two different machine learning approaches for semantic segmentation algorithms
	\item Demonstration of the fact that despite the low precision of the algorithm selector the resulting semantic segmentation is improved
\end{enumerate}

This paper is organized as follows. Section~\ref{sec:bck} introduces related and previous works and Section~\ref{sec:algosel} introduces the algorithm selection framework. Section~\ref{sec:exp} describes the experimentation and the results and Section~\ref{sec:con} concludes the paper and discusses future extensions.
\section{Previous Work and Background}
\label{sec:bck}
The algorithm selection have been used previously in the area of image processing as well as in certain applications to computer vision. The general idea behind the algorithm selection is to select a unique algorithm for a particular set of properties, attributes and features  extracted from the data or obtained prior to processing. The algorithm selection was originally proposed by Rice~\cite{rice:76} for the problem of operating system scheduler selection. Since then the algorithm selection have been used in various problems but has never become a main stream of problem solving.

The reason for which algorithm selection is not a mainstream is dual: on one hand it is necessary to find distinctive features and on the other hand the problem studied should be difficult enough that extracting additional features from the input data is computationally advantageous. 

\begin{figure}[bht]
	\centering
	\includegraphics[width=0.9\linewidth]{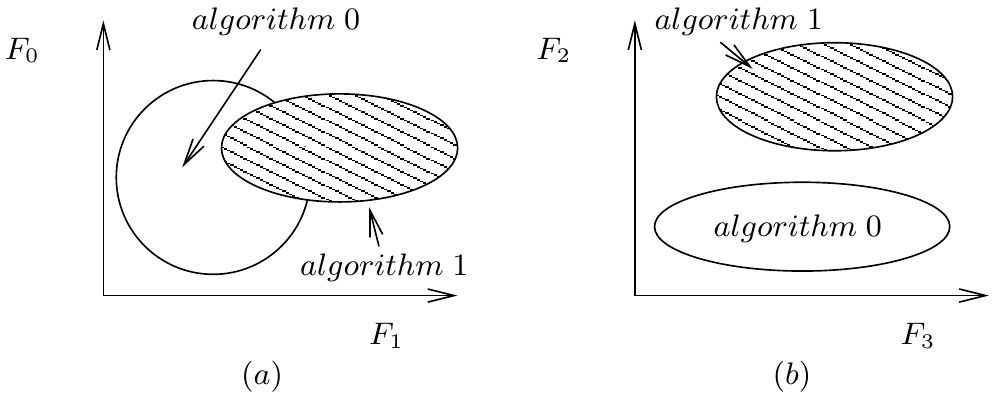}
	\caption{\label{fig:feats} Example illustrating (a) non-distinctive and (b) distinctive features}
\end{figure}

The distinctive features might be too expensive (computationally) to obtain and thus algorithm selection requires the selection of such features that provide the highest quality of algorithm selection using the least amount of features. This idea is illustrated in Figure~\ref{fig:feats}. Figure~\ref{fig:feats}a shows that when features are not well identified the algorithm selection does not allow to uniquely determine the best algorithm because the features are non-distinctive for the available algorithms. Counter example using distinctive features is shown in Figure~\ref{fig:feats}b.

The ratio of computational effort that is required to extract additional features to the whole computation of the result can be estimated by comparing their respective computational time. In~\cite{lukac:12b} it was shown that for the task of image segmentation the algorithm selection is directly proportional to the size of the processed region of the image. If the region of segmentation is too small, the resulting segmentation of the tested algorithms results in very similar f-values and thus selecting fastest/computational least expensive algorithm. For regions of larger size up to regions having the size of the input image, algorithm selection is both advantageous due to computational advantages as well as due to the increased quality of the result.

In computer vision and image processing the algorithm selection was previously on various levels of algorithmic processing. For instance, image segmentation of artificial~\cite{yong:03} or biological images~\cite{takemoto:09} was successfully implemented using algorithm selection approach. A set of features was found sufficient and allowed to clearly separate the area of performance of different algorithms. These two approaches however focused to separate the available algorithms only with respect to noise present in the image. Moreover, the algorithms used were single level line detectors such as Canny or the Prewitt. More complex algorithms for image segmentations were studied in~\cite{lukac:11d,lukac:12b}. Similarly to~\cite{yong:03,takemoto:09} a method using machine learning for algorithm selection for the segmentation of natural real-world images was developed. Other approaches have been studying the parameter selection or improving image processing algorithms using either machine learning or analytical methods but their approach is in general contained within a single algorithm~\cite{kolmogorov:07,peng:08,price:10}.

Methods and algorithms aimed at understanding of real world images have in general quite limited extend of their application. Currently there is a large amount of work combining segmentation and recognition and some of them are~\cite{ladicky:10,carreira:12}. In~\cite{leibe:08} uses an interleaved object recognition and segmentation in such manner that the recognition is used to seed the segmentation and obtain more precise detected objects contours. In~\cite{arbelaez:12} objects are detected by combining part detection and segmentation in order to obtain better shapes of objects. More general approaches such as~\cite{li:09} build a list of available objects and categories by learning them from data samples and reducing them to relevant information using some dictionary tool. However this approach does not scale to arbitrary size because the labels are not structured and ultimately require complete knowledge of the whole world.

In~\cite{hoiem:08} uses depth information to estimate whole image properties such as occlusions, background and foreground isolation and point of view estimation to determine type of objects in the image. All the modules of this approach are processed in parallel and integrated in a final single step. An airport apron analysis is performed in~\cite{ferryman:05} where the authors use motion tracking and understanding inspired by cognitive vision techniques. Finally, the image understanding can also be approached from a more holistic approach such as for instance in~\cite{oliva:01} where the intent is only to estimate the nature of the image and distinguish between mostly natural or artificial content.

\section{Algorithm Selection for Symbolic Segmentation}
\label{sec:algosel}
\begin{figure}[tbh]
\begin{center}
\fbox{
   \includegraphics[width=0.9\linewidth]{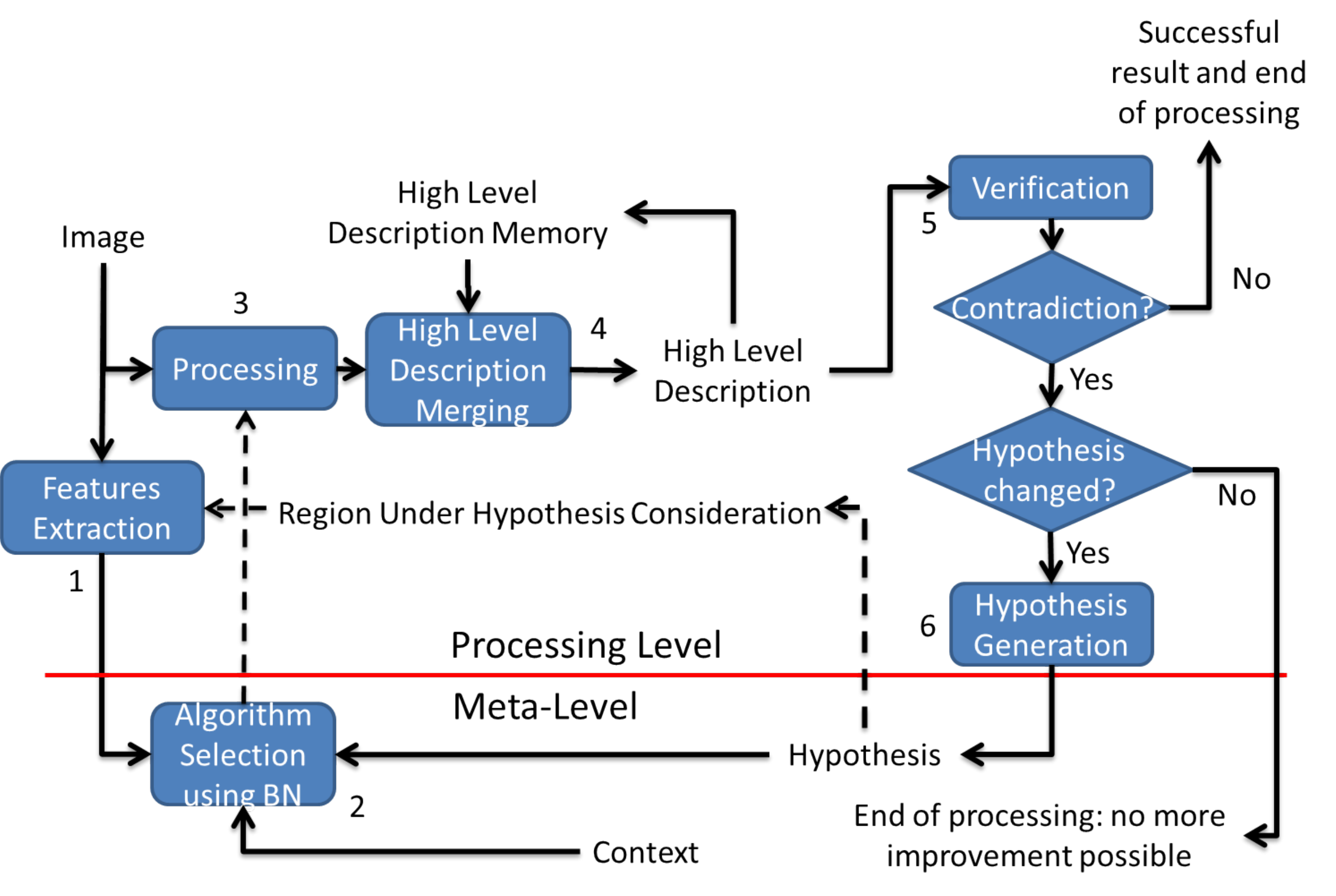}
}
\end{center}
   \caption{\label{fig:system}Algorithm Selection Platform}
\end{figure}

The framework used in this experiments was originally introduced in~\cite{lukac:13}. The schematic representation is shown in Figure~\ref{fig:system}. The processing start by extracting features (1) from the input image which are used by the algorithm selector (2) to determine the most appropriate algorithm. The input image is processed by the selected network of algorithms (3) which results in symbolic segmentation of the input image. The symbolic segmentation result is interpreted by constructing a multi-relational graph representing the high level description. The high-level description is analyzed for symbolic contradiction (5).

The contradiction is obtained using a contradiction which is based on co-occurrence statistics obtained from training data. If a contradiction is detected a new hypothesis about a region containing a contradiction is generated by the largest co-occurrence statistics given the symbolic segmentation for all but one regions being fixed.  Once the new hypothesis is generated it is used as an additional input to the algorithm selector. Finally, features are extracted from the region of the contradiction. This new set of features values and hypothesis attributes are used for a new algorithm selection.

The newly selected algorithm processes the whole image and generates a new symbolic segmentation. The region that before contained the contradiction is now extracted and is merged with the original high-level description (4). The new high-level description is analyzed and the cycle begins again. The processing stops when for a given input there are no more contradictions or when no more algorithms can be selected. This amounts to either have no more errors in the high level description or when no more new hypotheses can be generated. This platform will be referred to as Iterative Analysis (IA) as it incrementally changes the high level description of the input image.
\begin{figure}[bh]
\begin{center}
\fbox{
   \includegraphics[width=0.9\linewidth]{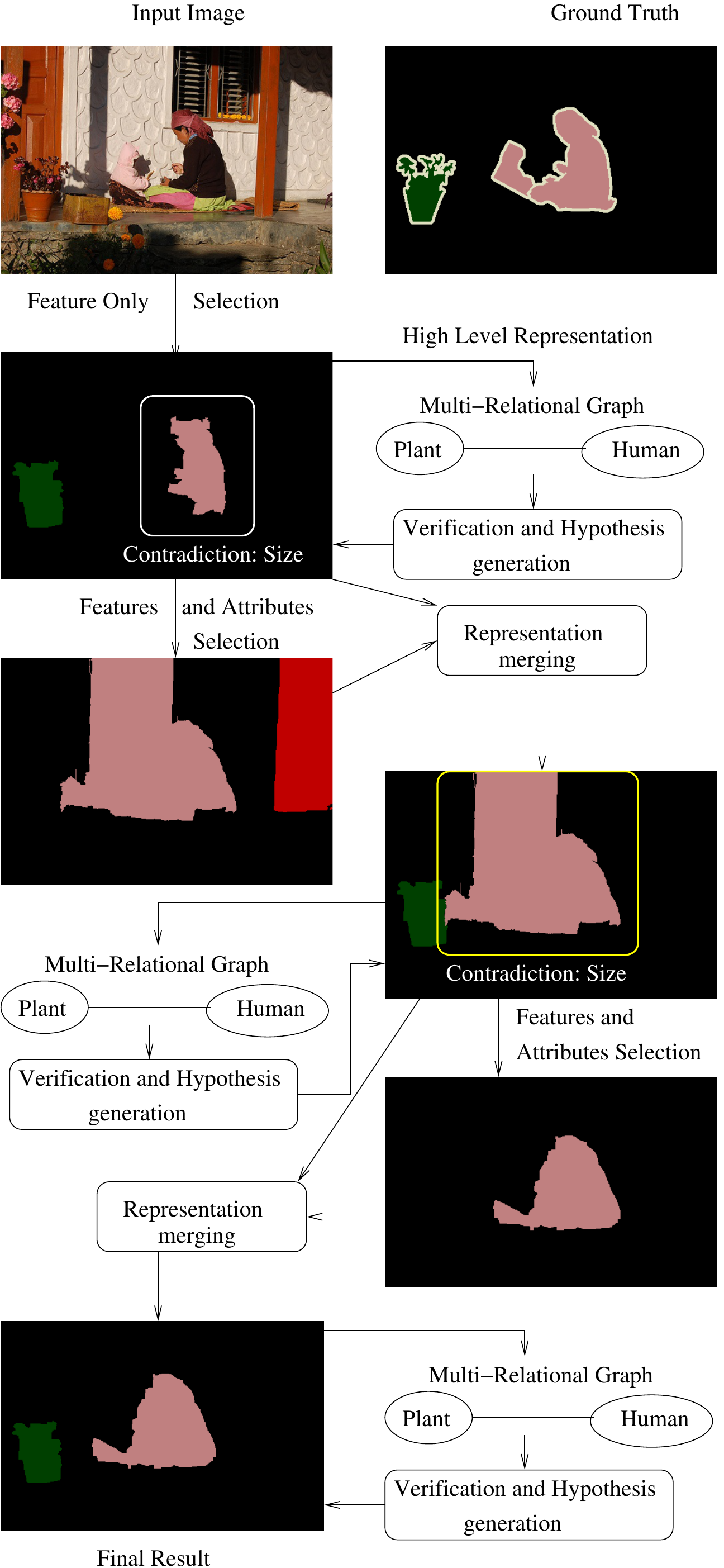}
}
\end{center}
   \caption{\label{fig:exm}Exemplar processing of an input image by the IA platform}
\end{figure}

The output of symbolic segmentation algorithm is a set of labeled regions. The high-level interpretation (description) consists of building a multi-relational graph, that specifies relations between the labeled regions in the resulting image. Using this graph the result is checked for contradiction and a hypothesis about the recognized objects' relations is generated if necessary. Currently, the IA platform uses co-occurrence statistics obtained from training data to estimate the contradiction and to propose most viable hypothesis. The estimated relations are the relative position (left, right, above, below), relative size (larger, smaller, same), background/foreground (in front, back) and one single object property which is the  shape (Hough transform). Each of these properties are applied to either a pair of objects or individual objects and the probability of contradiction is generated as a cumulative normalized product of all individual scores. An example of IA processing an image is shown in Figure~\ref{fig:exm}.

The verification is intended as an additional source of information; the reasoning over the recognized regions is performed only on relational level and thus only if two or more regions are detected our method is applicable.
\section{Experiments}
\label{sec:exp}
To evaluate the proposed framework we used the VOC2012 data and three different algorithms for symbolic segmentation~\cite{ladicky:10,carreira:12,bharath:14}. Each of the algorithms use similar or none preprocessing, different segmentation and similar classification machine learning based object recognition. All three algorithms have been evaluated and tested on the VOC2012 data set. 

As introduced in Section~\ref{sec:algosel} the high level verification requires multiple objects detections in one image. Consequently the testing and the training of the IA platform was carried only on images that contain more than one distinct object. The training set requires that not only the input contains more than one objects in the ground truth but also that at least one of the algorithms used is able to detect at least two objects in the image. Failing to do so the verification procedure will not be triggered and the iterative process of high level understanding improvement could not be started.
\begin{figure}
	\centering
        \begin{subfigure}[b]{0.5\linewidth}
		\includegraphics[width=\textwidth]{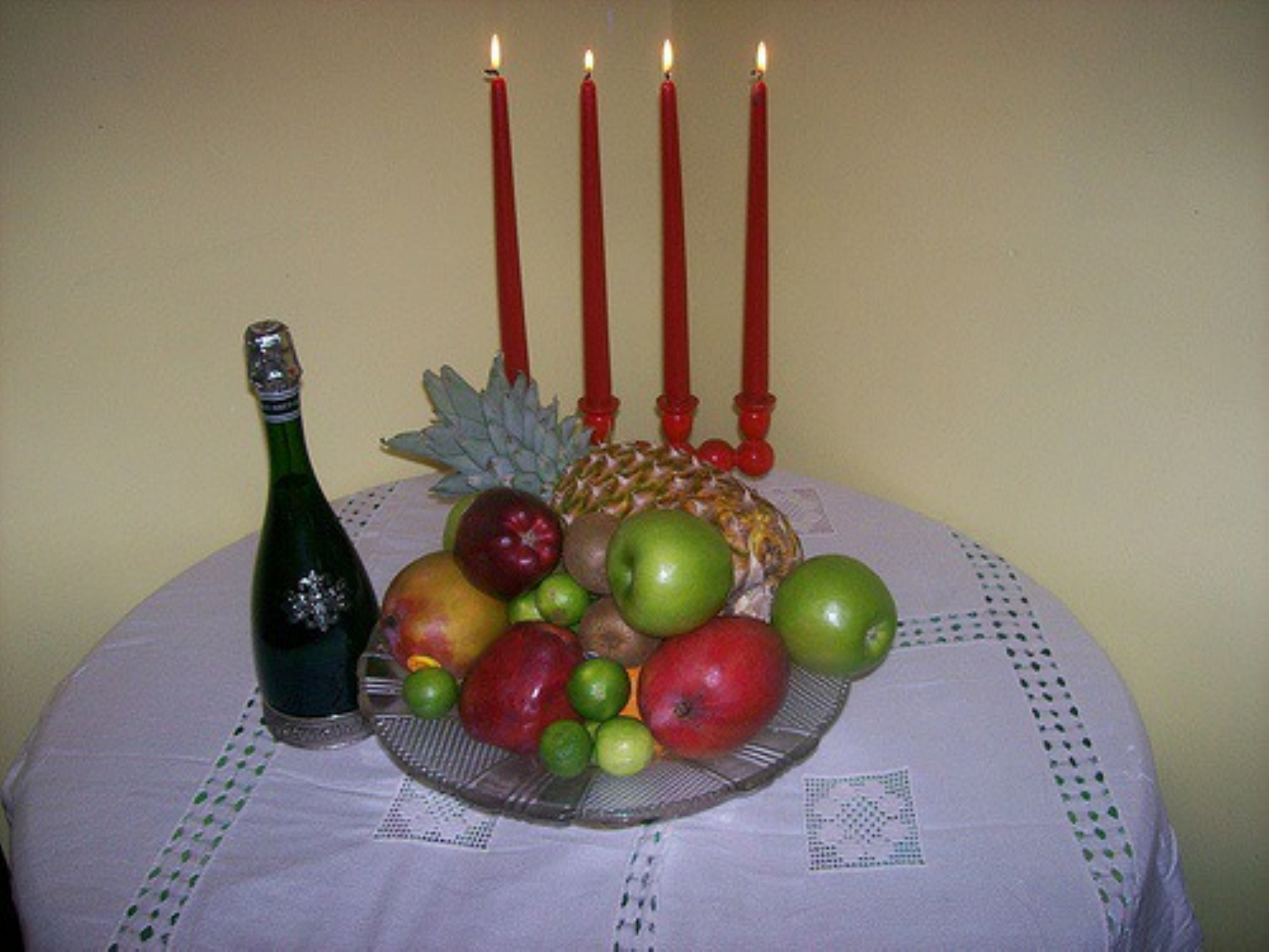}
		\caption{Input/Original Image}
		\label{fig:input}
	\end{subfigure}%
	~ 
	\begin{subfigure}[b]{0.5\linewidth}
		\includegraphics[width=\textwidth]{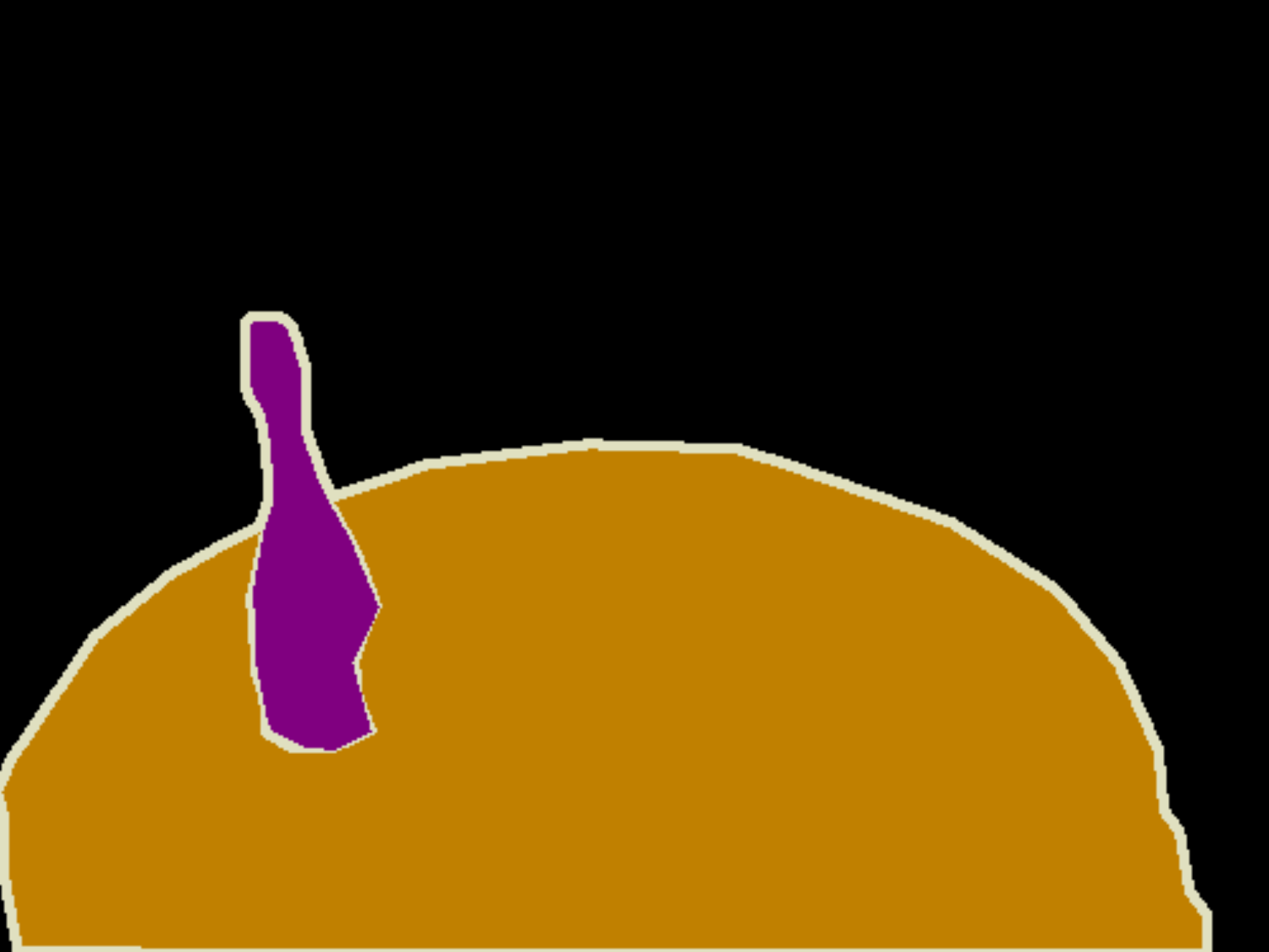}
		\caption{Ground Truth}
		\label{fig:gt}
	\end{subfigure}
        \begin{subfigure}[b]{0.7\linewidth}
		\includegraphics[width=\textwidth]{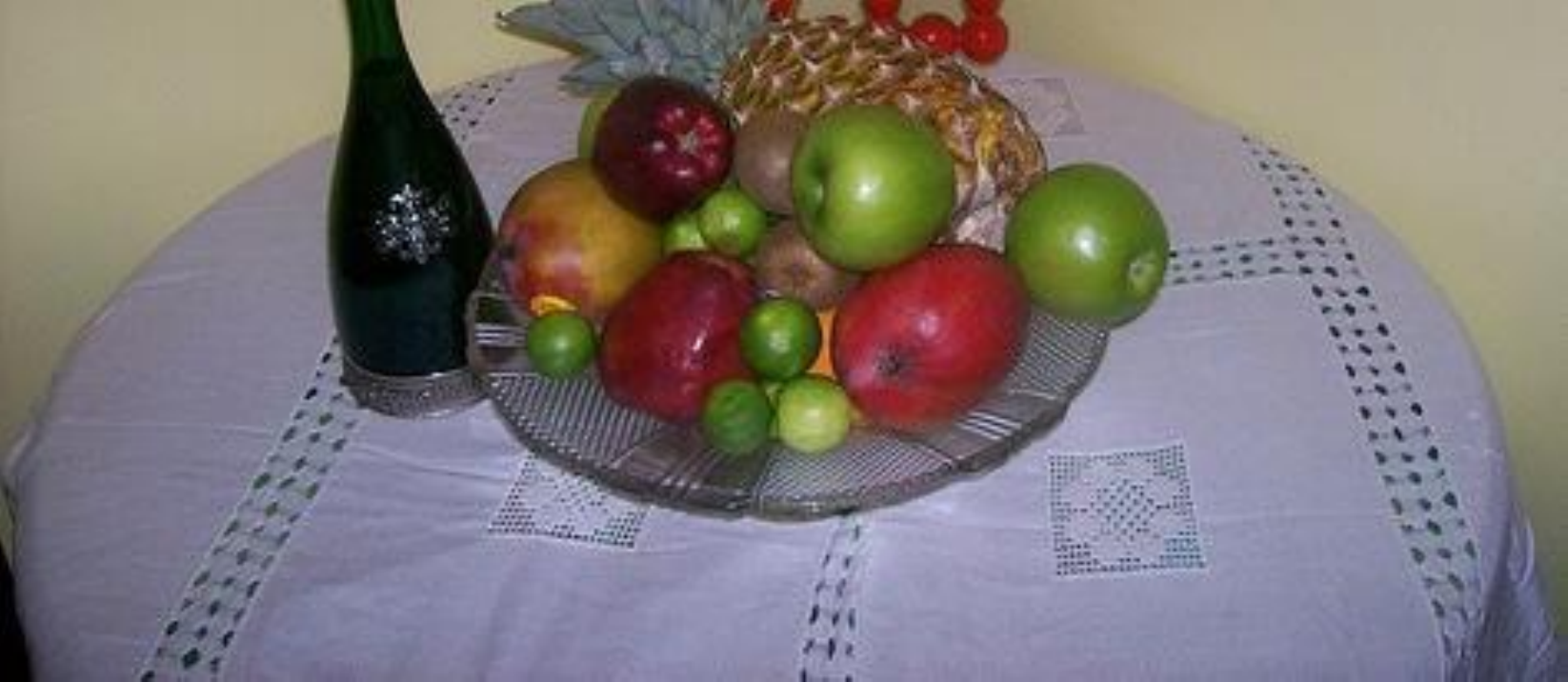}
		\caption{Table BB}
		\label{fig:obj1}
	\end{subfigure}%
	~ 
	\begin{subfigure}[b]{0.3\linewidth}
		\centering
		\includegraphics[width=.3\textwidth]{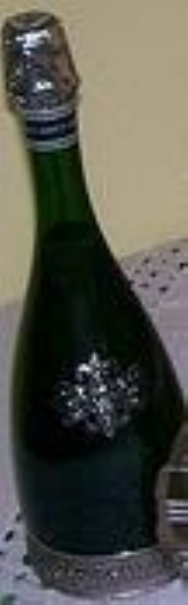}
		\caption{Bottle BB}
		\label{fig:obj2}
	\end{subfigure}
\end{figure}
The experiments are carried over various features' set and terminating conditions. We evaluate two different algorithm selection algorithms: a Bayesian Network (BN) and support vector machine (SVM). The motivation for using these two different methods is one hand given by the ability of using hierarchy of information and thus to reduce the complexity of learning and on the other hand the simplicity and in general good learning results of BN and SVM respectively.

\subsection{Training of the Algorithm Selector}

For SVM algorithm selection two SVM are trained: one for the selection of algorithms from image features only SVM$_f$ and one for algorithm selection using features and hypothesis attributes SVM$_a$. Such approach is used as a solution to the problem of missing values in the inputs of SVM~\cite{pelckmans:05} and is one of the possible solutions~\cite{mallison:03}. Initially two separate SVM machines have been used: one for the initial algorithm selection using only image features and another one for selection using features and hypothesis attributes. However it was shown experimentally that patching approach~\cite{mallison:03} outperformed the two separate SVMs. Using the patching approach, whenever the attributes of an image could not be obtained (hypithesis was not generated, or it is unknown) the attributes values were generated by the average of the available values.

The first training data set $T_f$ is equivalent to the VOC2012 training data set. In the case of SVM$_f$ only features are extracted. The feature vector contains all together 7856 feature values composed from histograms of various features. The features used are brightness, fft, gabor, wavelets, rgb intensity, acutance, and so on. 
The second training data set $T_a$ is created from bounding boxes of around the semantic segmentations in the training set of VOC2012 data set. Same features as in $T_f$ but additionally a set of attributes extracted from the region corresponding to the region of the correct semantic segmentation is extracted using the Matlab regionprops function.

In the case of the BN only $T_a$ is used for training as the BN is well suited to handle missing input values. However the BN approach requires deterministic input values - observations. Because most of the features extracted are continuous values within a certain range it is necessary to cluster the data to discrete values. The clusterization is done using an equivalent ranges for each value given by~(\ref{eq:ranges}). 
\begin{equation}
	r_i = ](max_f-min_f)/k*(i-1), (max_f-min_f)/k*(i)]
	\label{eq:ranges}
\end{equation}

\begin{figure}[tbh]
\begin{center}
\fbox{
   \includegraphics[width=0.9\linewidth]{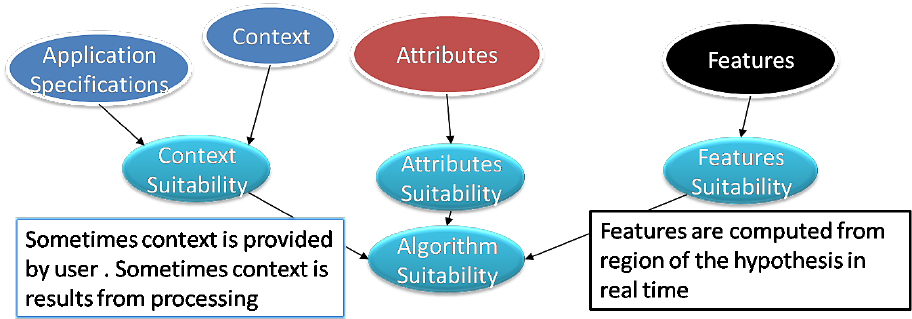}
}
\end{center}
   \caption{\label{fig:bn}Bayesian Network used for Algorithm Selection in some experiments}
\end{figure}

The BN structure is shown in Figure~\ref{fig:bn} and the inputs are specified by three categories: application specifications, hypothesis attributes and image features. 
The application specification represents input information about the target application and other application related information that are constant in the framework of this study. The attributes are regional properties extracted regionprops command in Matlab and represent the attributes for each of the available hypotheses. The hypotheses are the available labels for used in region labeling. Here the labels corresponds to the 20 classes and a background from the VOC database. Each attribute for a class is calculated as an average of the values extracted from all objects of that class encountered in the training data set. The extracted features from the image are together with the attributes clustered as described in next subsection.

Both the training and the testing data however are fairly imbalance as can be seen in Table~\ref{tab:datbal}.
\begin{table}[bht]
	\centering
	\caption{\label{tab:datbal} The distribution of samples representing each of the algorithms}
	\begin{tabular}{|c|c|c|}
		\hline
		ALE~\cite{ladicky:10}&COMP6~\cite{bharath:14}&CPMC~\cite{carreira:12}\\
		\hline
		35\%&42\%&23\%\\
		964&1133&633\\
		\hline
	\end{tabular}
\end{table}

The creation of the training sets follows different principles depending whether the training set is $T_f$  or $T_a$. For the $T_f$ data set, each sample image is evaluated as $\frac{1}{C_I}\sum_{c\in C_I} F_c$ with $C_I$ being all labels present in the ground truth of image I, and $F_c$ is the f-value of the symbolic segmentation of class $c$ in image I. In $T_a$ the evaluation of each algorithm is done only with respect to the region representing a single label fully enclosed in the bounding box provided by the VOCdevkit. 

Finally, the experimental results have shown that using all data for learning the algorithm selection is not well suited because many images have relatively close results of processing by more than one algorithm. Let, $I_j$ be an input image and $F_{nj}$ are f-values calculated on the output of each algorithm $n$ applied to $I_j$, let $F_j=\{F^0\geq F^1\geq\ldots\geq F^n\}$ be the ordered set of $F_{nj}$ then a $I_j$ is used for learning if $\vert F^0-F^1\vert \geq\theta$. In most of the experiments in this paper $\theta$ was set to $0.5$.

\subsection{Testing of the Platform}

The testing of the system was done over a subset of images from the VOC2012 validation data set; images that contain at least two objects in the ground truth. At first we evaluate the algorithm selector ability to learn to classify the images according to which algorithm results in best symbolic segmentation. To evaluate the classification power of both algorithms we analyzed results both for binary classification (with two different algorithms for semantic segmentation) and for multi-class classification (using all three available semantic segmenters). Then the whole system is analyzed by looking at the resulting data. 

First we evaluated the BN for various levels of data clustering. Intuitively, the size of the BN is directly and inversely proportional to the number of values on the input observations; the conditional probabilities tables in the nodes where the observable inputs are connected to grows according to $k^n$ with $k$ being the number of observable values of the input variables and $n$ being the number of input variables connected to this node. The experimentation using the BN was carried in Matlab using the BNT~\cite{murphy:01} package and the learning of the BN was performed using the EM algorithm. The results of evaluating the BN classification power on the $T_a$ data set with respect to the number of observed data values is shown in Table~\ref{tab:bneval}
\begin{table}[bht]
	\centering
	\caption{\label{tab:bneval} BN classification results with respect to the number of data values}
	\begin{tabular}{|c|c|}
		\hline
		Clusters&BN Classification Error\\
		\hline
		3&53\%\\
		5&53\%\\
		6&42\%\\
		7&92\%\\
		\hline
	\end{tabular}
\end{table}
Notice that for $k=7$ the EM algorithm used for BN learning results in very high error rate of classification and for any $k>7$ the EM does not converge.
The BN is fairly limited in the number of input nodes as well. Because the conditional probability table in each node of the BN grows using~\ref{eq:ranges}. Consequently using the BNT Matlab package we were able to experiment with a BN having at best 10 sextenary input feature variables. 

Because the BN requires the best features for high quality of classification we performed two different experiments of classification with BN: (a) search for best features for BN and (b) using clustered PCA features. The results using the $T_a$ data set are shown in Table~\ref{tab:bnres}.
\begin{table}[bht]
	\centering
	\caption{\label{tab:bnres} BN classification results}
	\begin{tabular}{|c|c|c|c|}
		\hline
		Task&Clusters&Number of Features&BN Error\\
		\hline
		2-class&3&8 PCA& 50\%\\
		2-class&3&5&49\%\\
		3-class&2&21&48\%\\
		3-class&3&11 PCA&49\%\\
		\hline
	\end{tabular}
\end{table}
Contrary to the BN the SVM uses continuous features values and only normalization is required. Moreover SVM works well with large input vectors that are in general reduced using PCA for increased speed and accuracy of classification. The results of testing of the SVM classification using the $T_f$ and $T_a$ training data is shown in Table~\ref{tab:svmeval}. The evaluation was done using two data sets; one data set contained image regions (bounding boxes with individual semantic segmentations) and another data set contained full images (denoted FI in Table~\ref{tab:svmeval}).
\begin{table}[bht]
	\centering
	\caption{\label{tab:svmeval} SVM classification results using the 3 most significant PCA features. (FI) means that the training and testing data is using whole images.}
	\begin{tabular}{|c|c|c|}
		\hline
		Task&Data set&SVM Classification Error\\
		\hline
		2-class&Train $T_a$&25\%\\
		2-class&Test $T_a$&27\%\\
		2-class&Train $T_f$&34\%\\
		2-class&Test $T_f$&37\%\\
		\hline
		3-class&Train $T_a$&47\%\\
		3-class&Test $T_a$&51\%\\
		3-class&Train $T_f$&50\%\\
		3-class&Test $T_f$&53\%\\
		3-class&(FI)Train $T_a$&42\%\\
		3-class&(FI)Test $T_a$&46\%\\
		3-class&(FI)Train $T_f$&47\%\\
		3-class&(FI)Test $T_f$&54\%\\
		\hline
	\end{tabular}
\end{table}
The main result that can be seen in Table~\ref{tab:svmeval} is that the error rate on classification is significantly smaller than when the SVM is using only features. Moreover all experiments where no attributes are used, the SVM is given mean values of the attributes. When the SVM was used completely without the attributes and was trained exclusively in he features the results have even lower accuracy of selection. Consequently all experiments on the IA platform were done using a single SVM that was either given only features and mean values of attributes or features and hypothesis attributes. Moreover, as can be expected the error rate of classification is significantly lower for two algorithms. 

We can see that both the learning of the whole images as well as the learning of segments performs relatively poor with both the SVM and the BN. However the IA  platform uses high level verification and thus it was tested with the best of the algorithm selector, the SVM. 

To evaluate the IA platform data from the VOC2012 trainval set was used. The average precision of the of the three algorithms and the iterative analysis approach is shown in Table~\ref{tab:plateval}.
\begin{table}[bht]
	\centering
	\caption{\label{tab:plateval} The f-measures for all three algorithms and for the presented approach}
	\begin{tabular}{|c|c|}
		\hline
		Algorithm&average f-value\\
		\hline
		COMP6&44.469\%\\
		IA&43.554\%\\
		ALE&43.144\%\\
		CPMC&32.060\%\\
		\hline
	\end{tabular}
\end{table}

\begin{figure*}
	\centering
	\begin{tabular}{cccccc}
		\includegraphics[width=0.15\textwidth]{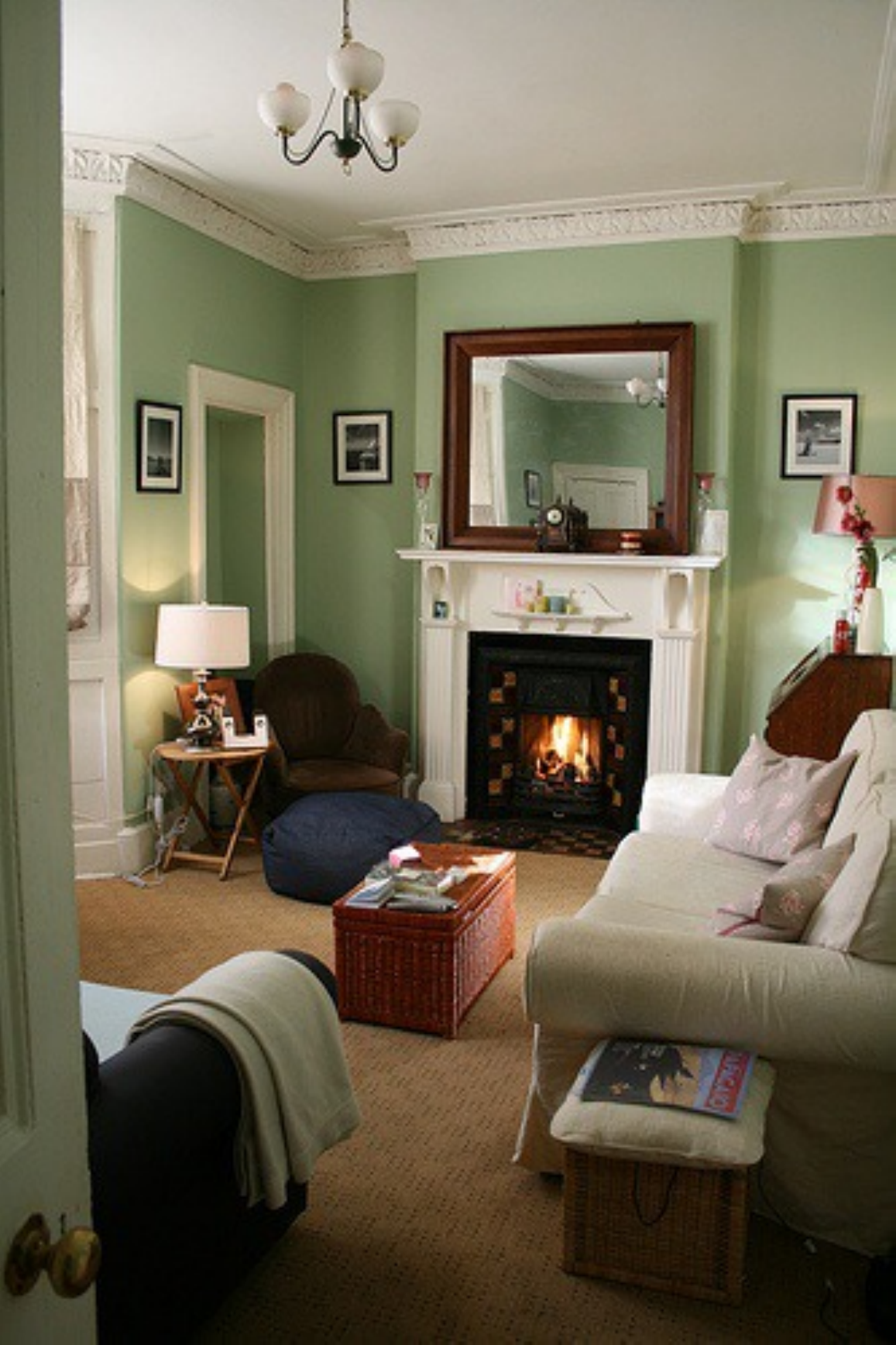}&\includegraphics[width=0.15\textwidth]{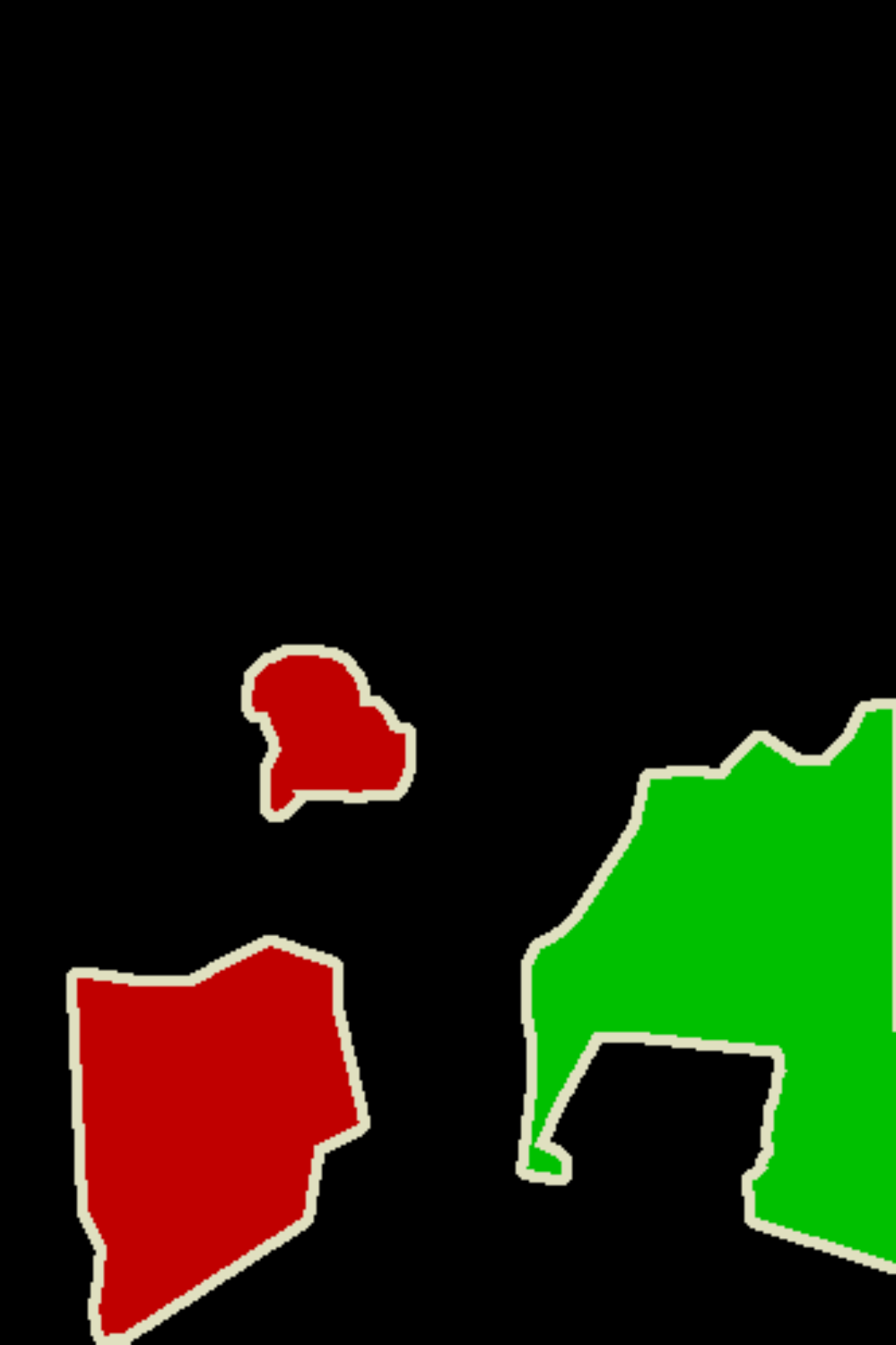}&\includegraphics[width=0.15\textwidth]{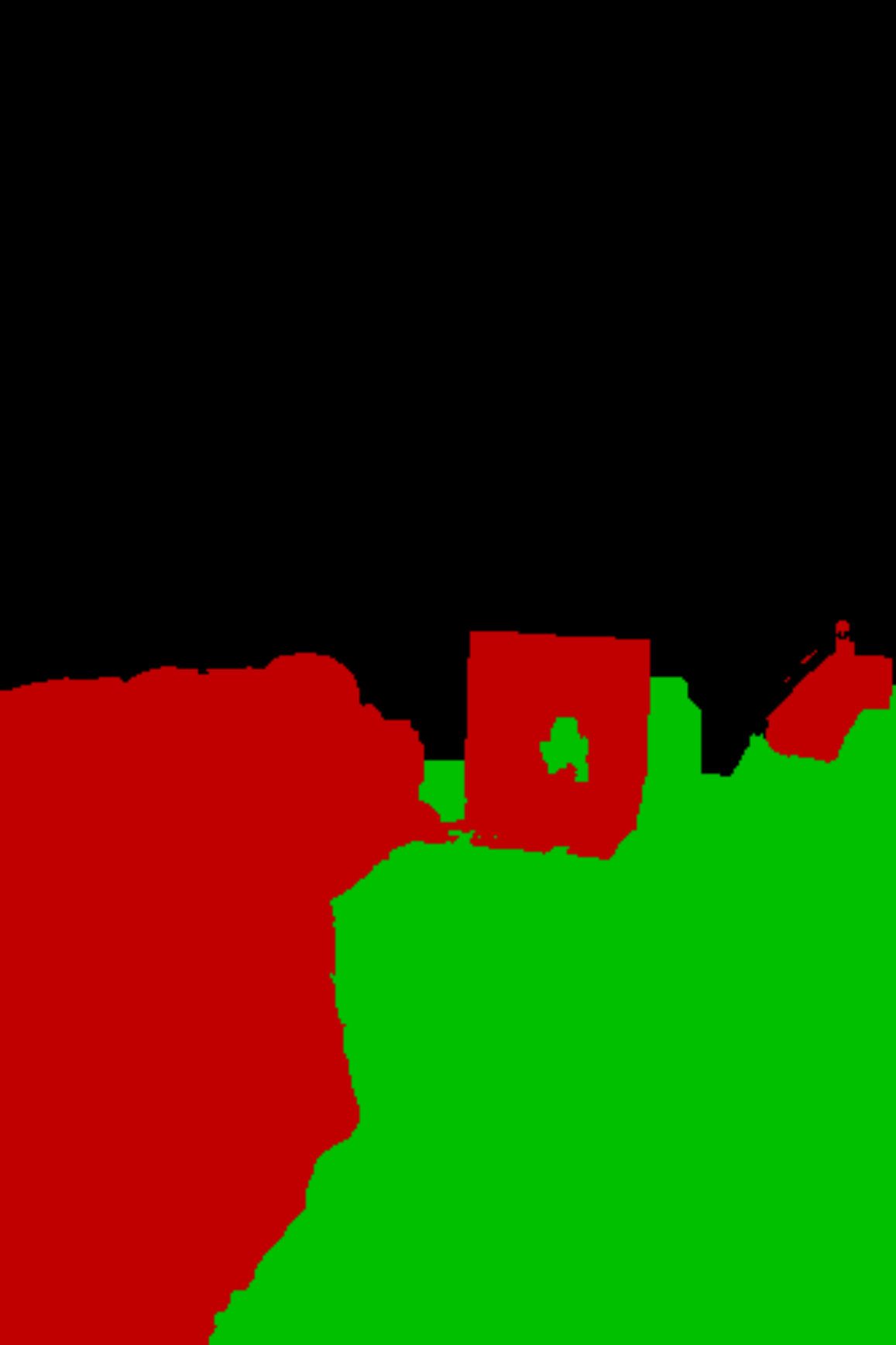}&\includegraphics[width=0.15\textwidth]{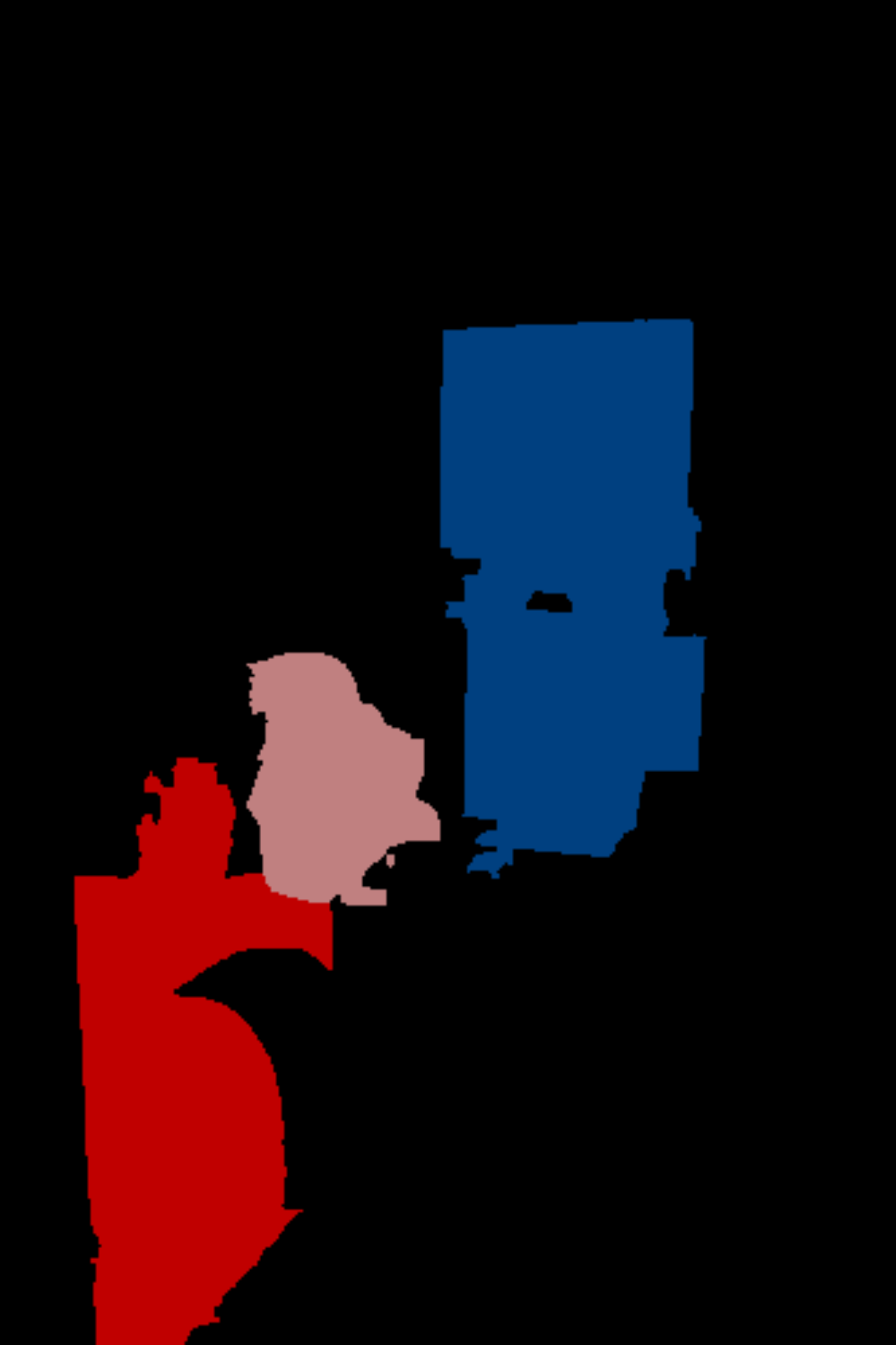}&\includegraphics[width=0.15\textwidth]{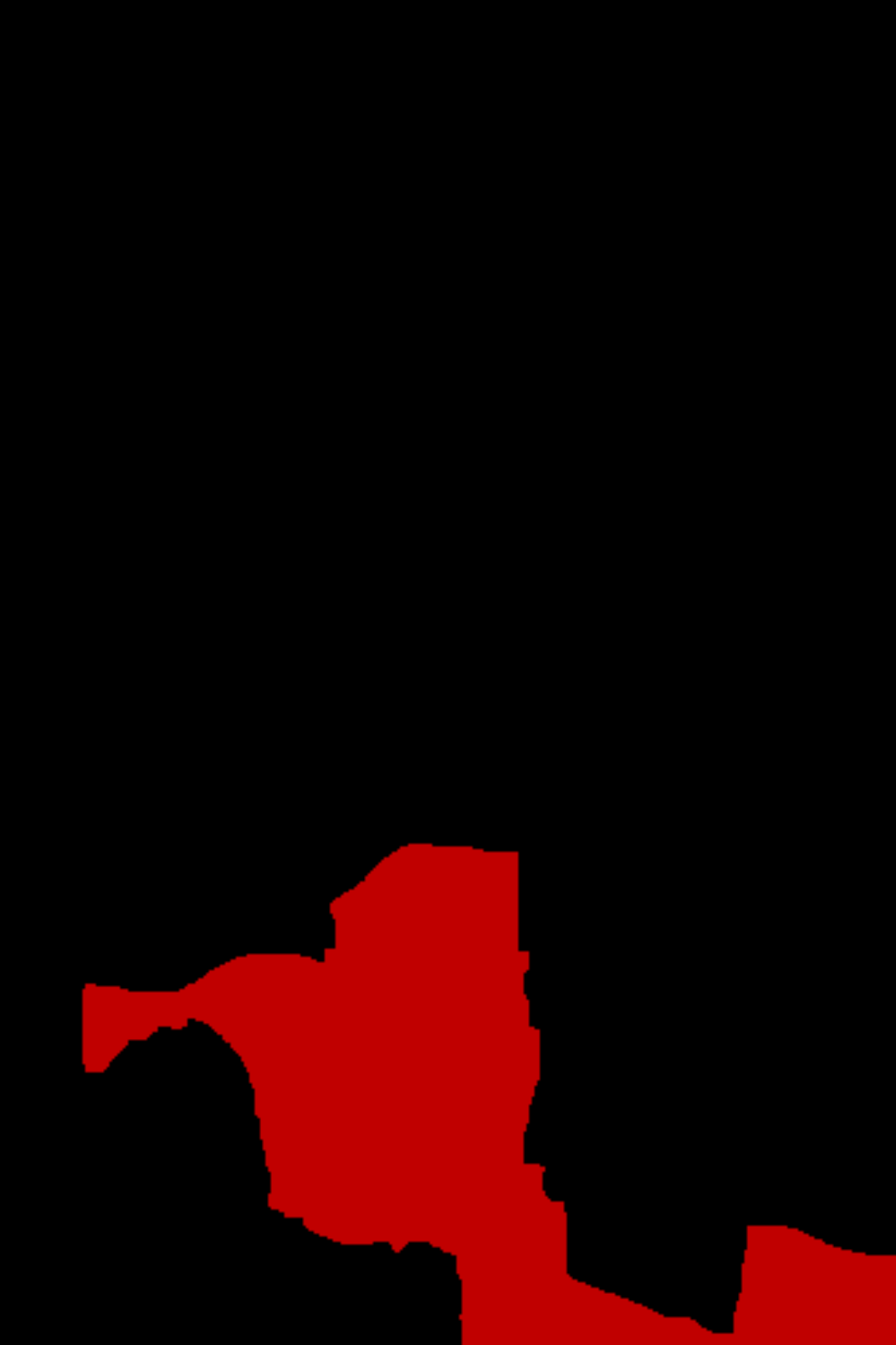}&\includegraphics[width=0.15\textwidth]{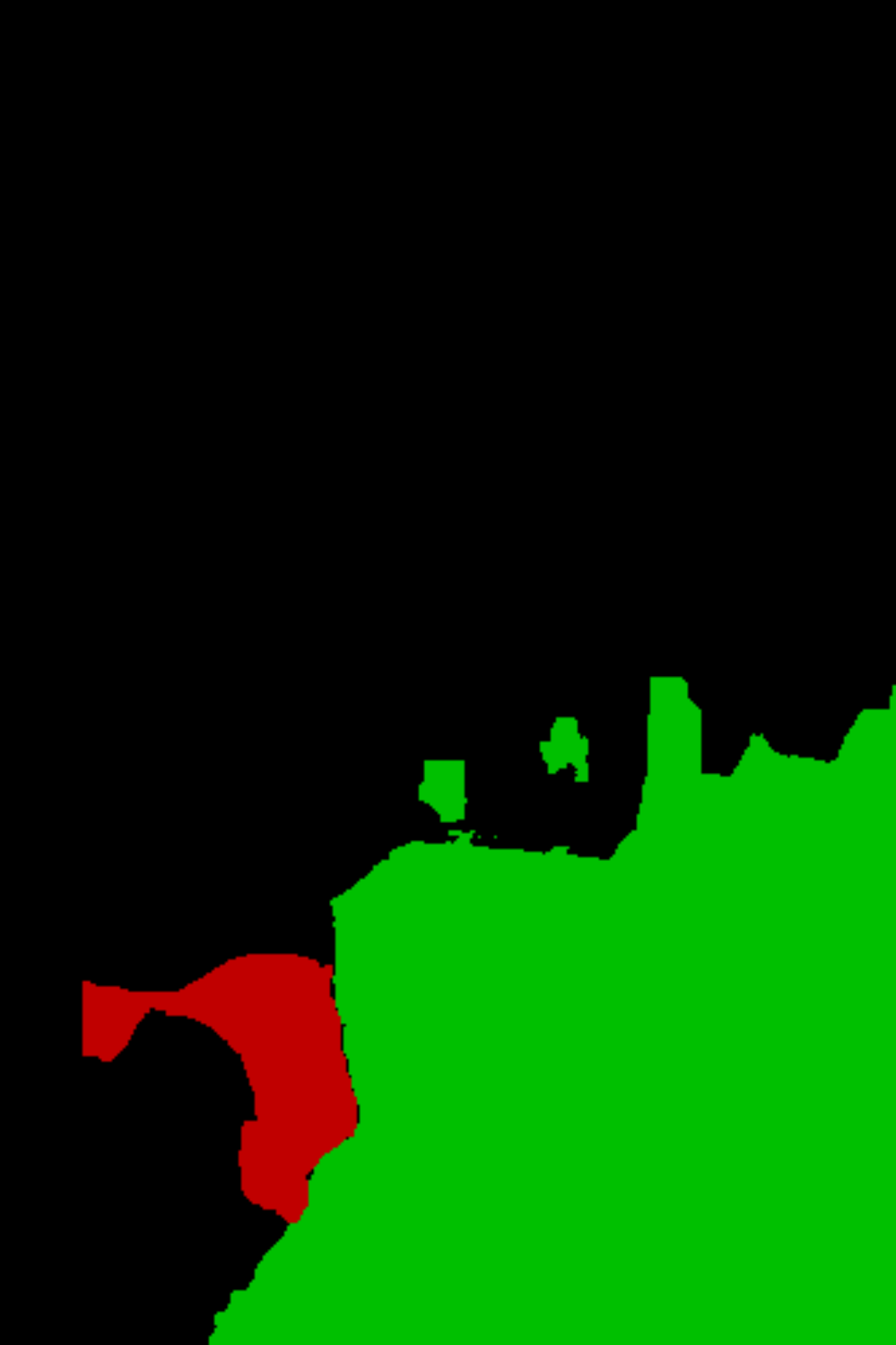}\\
		\includegraphics[width=0.15\textwidth]{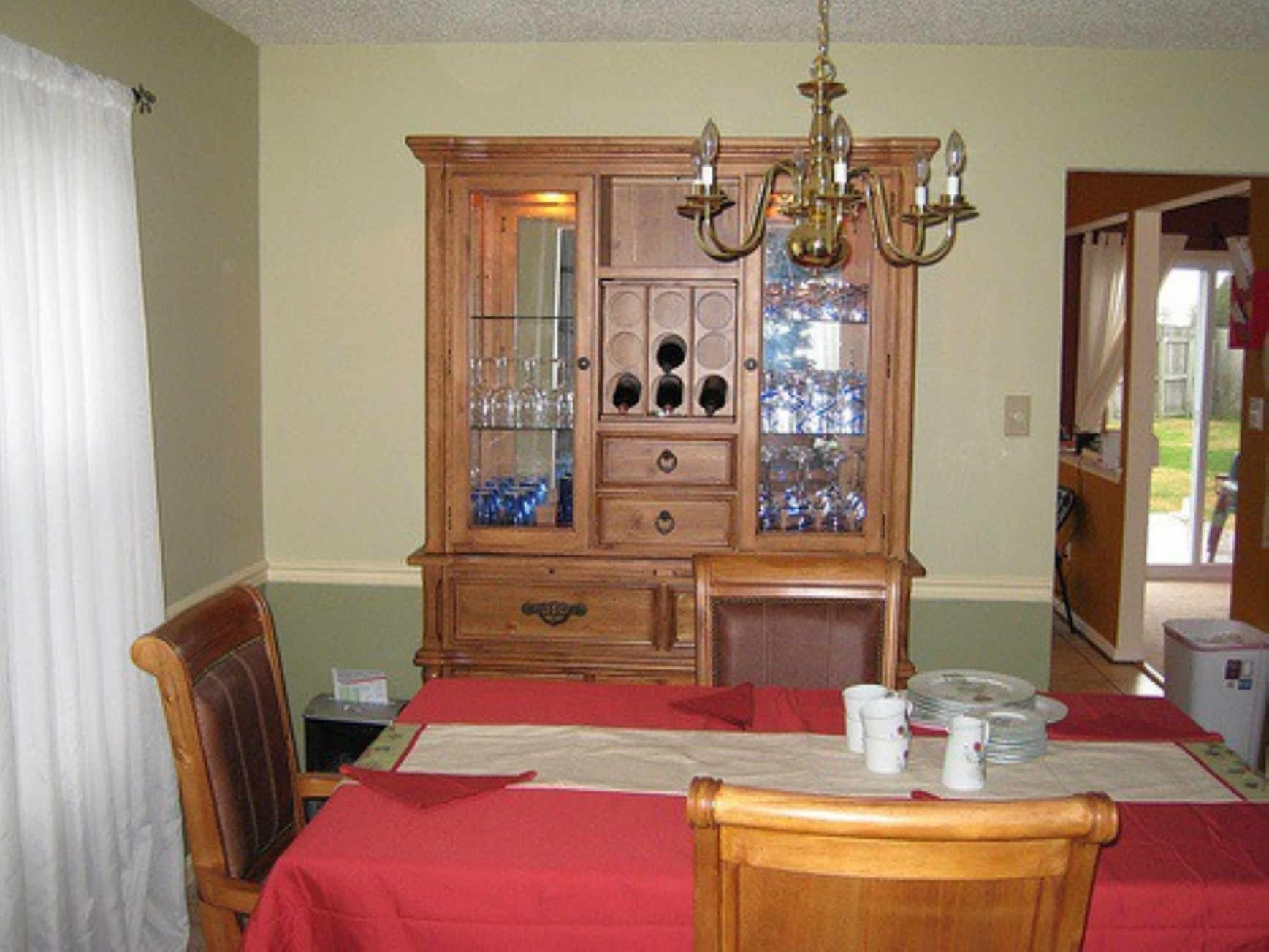}&\includegraphics[width=0.15\textwidth]{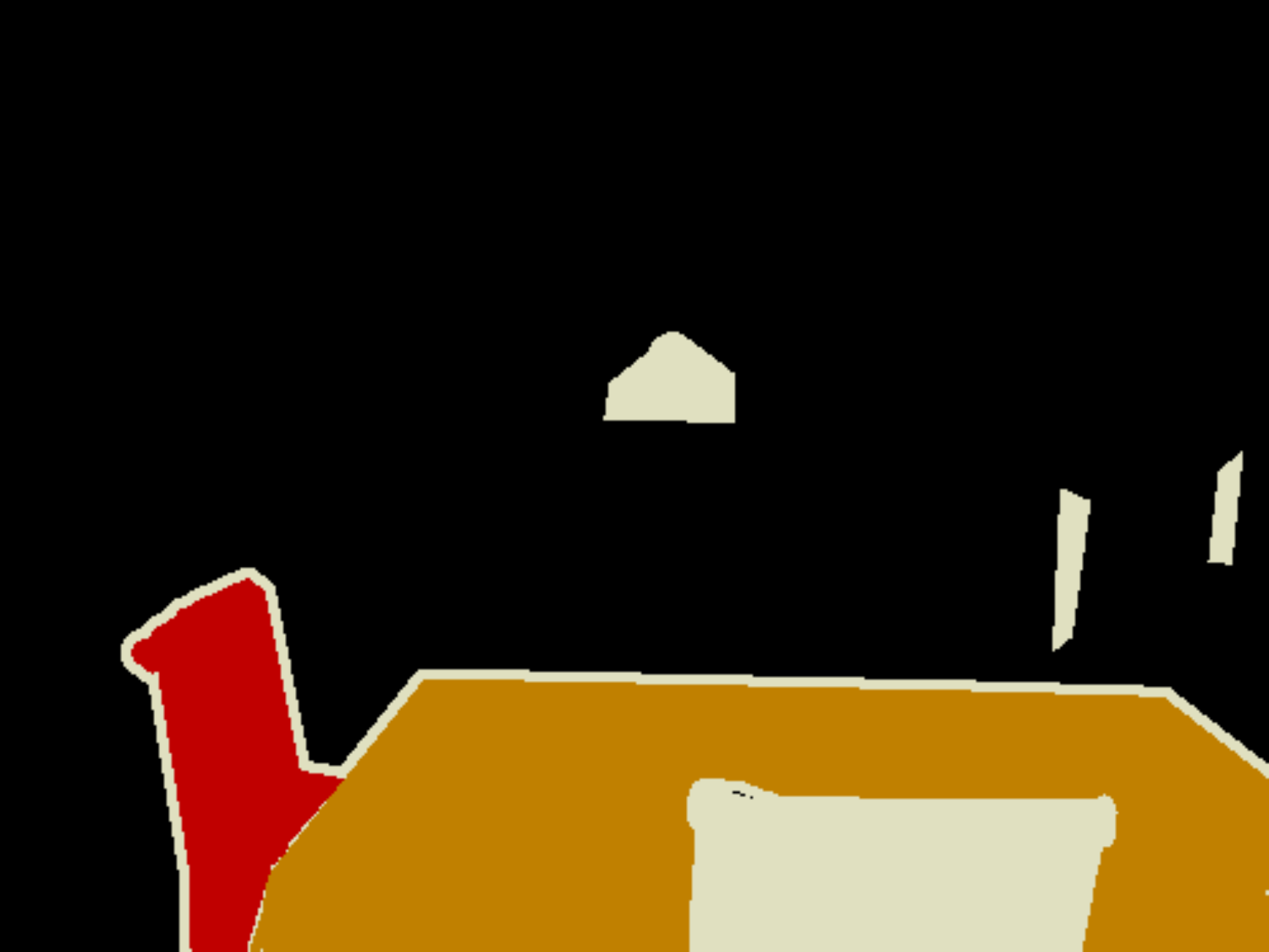}&\includegraphics[width=0.15\textwidth]{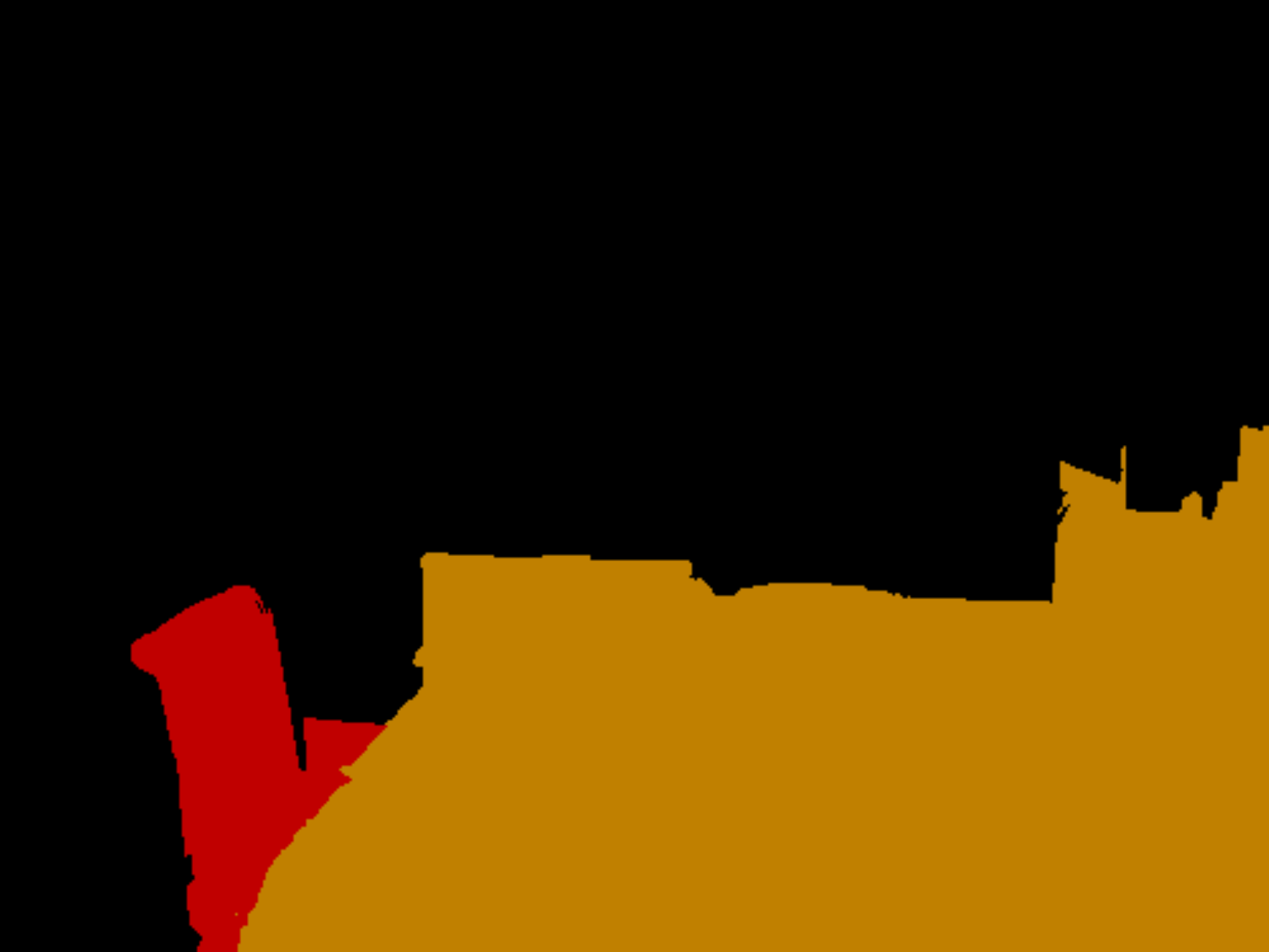}&\includegraphics[width=0.15\textwidth]{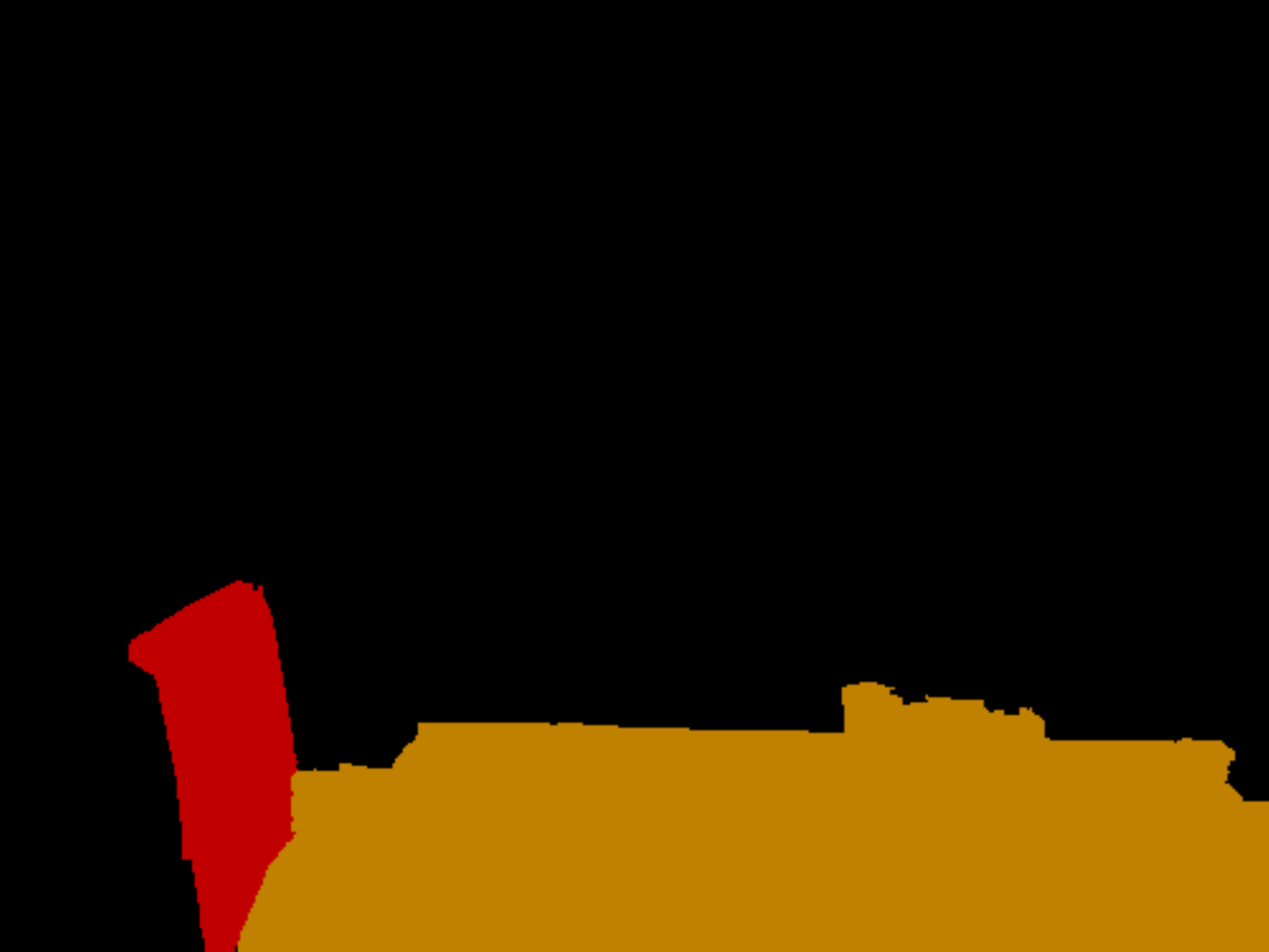}&\includegraphics[width=0.15\textwidth]{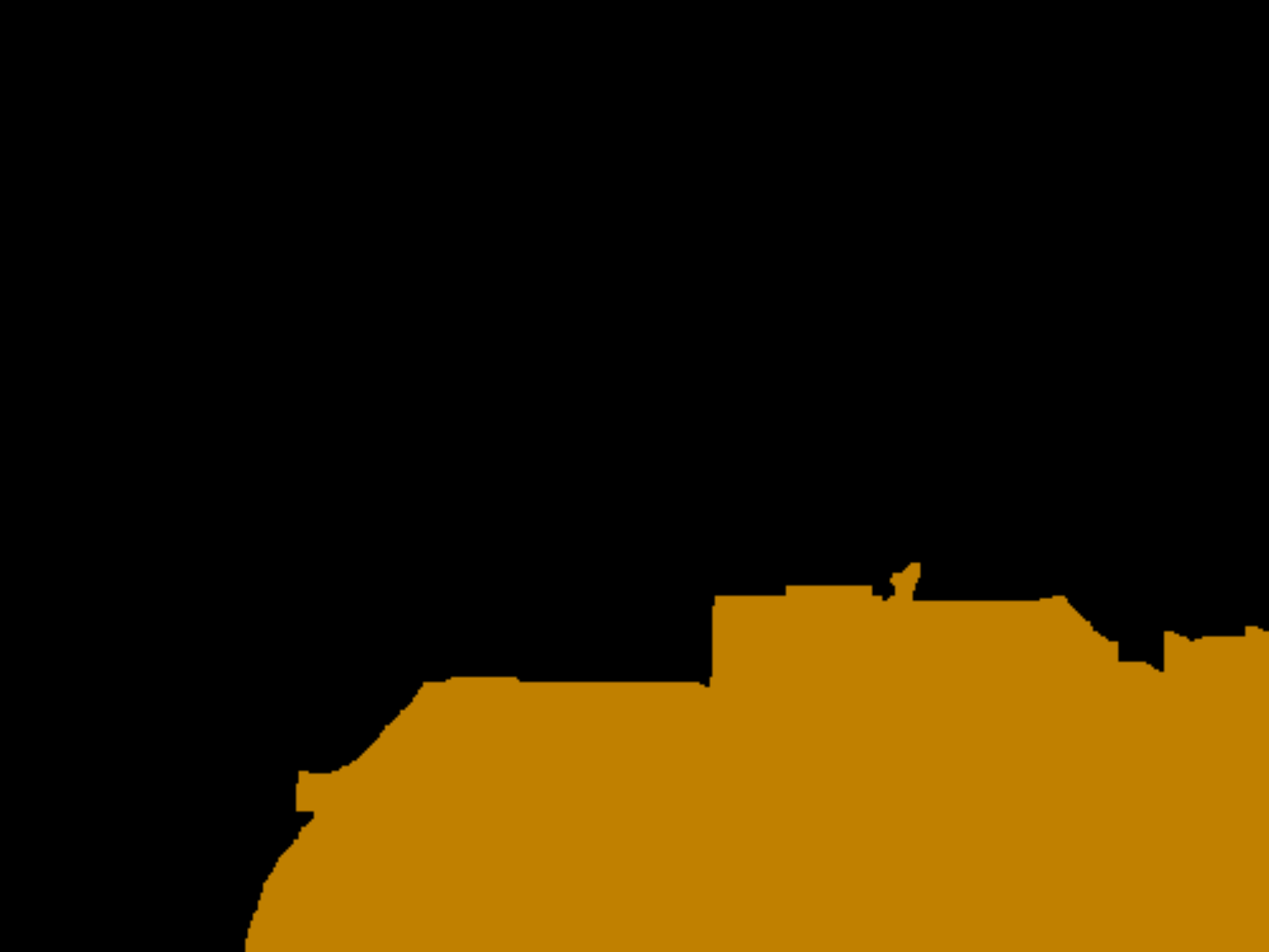}&\includegraphics[width=0.15\textwidth]{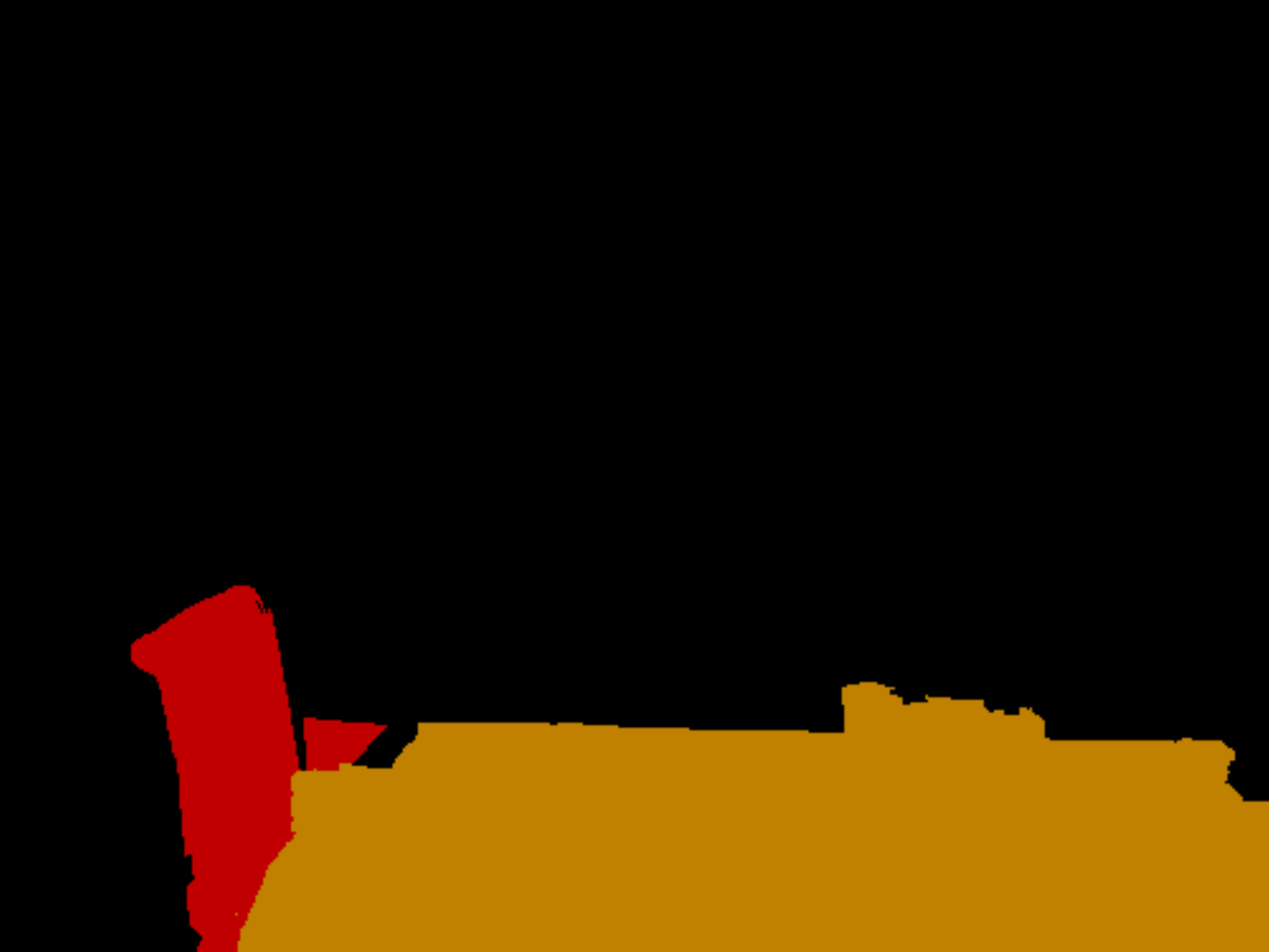}\\
		\includegraphics[width=0.15\textwidth]{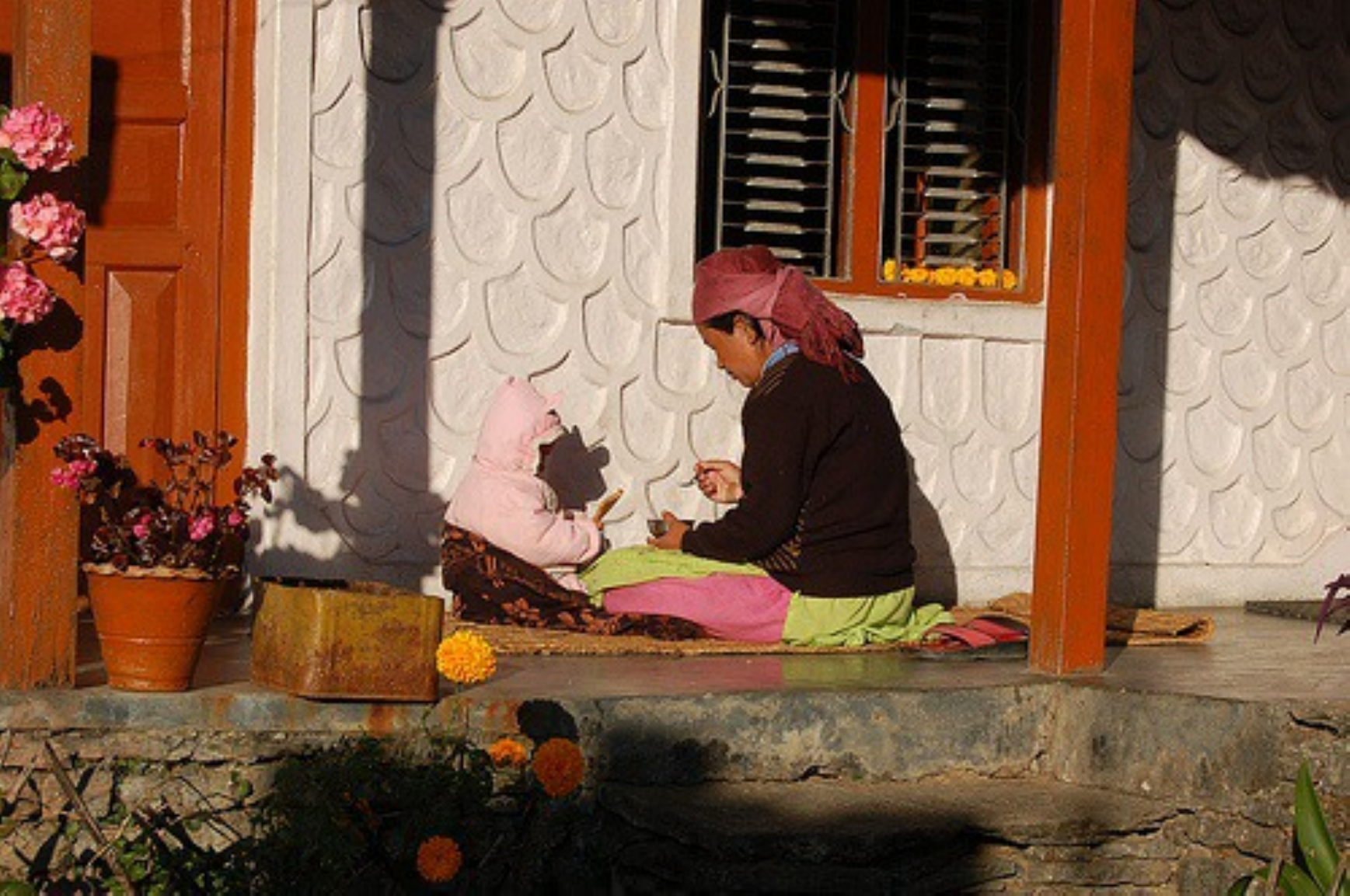}&\includegraphics[width=0.15\textwidth]{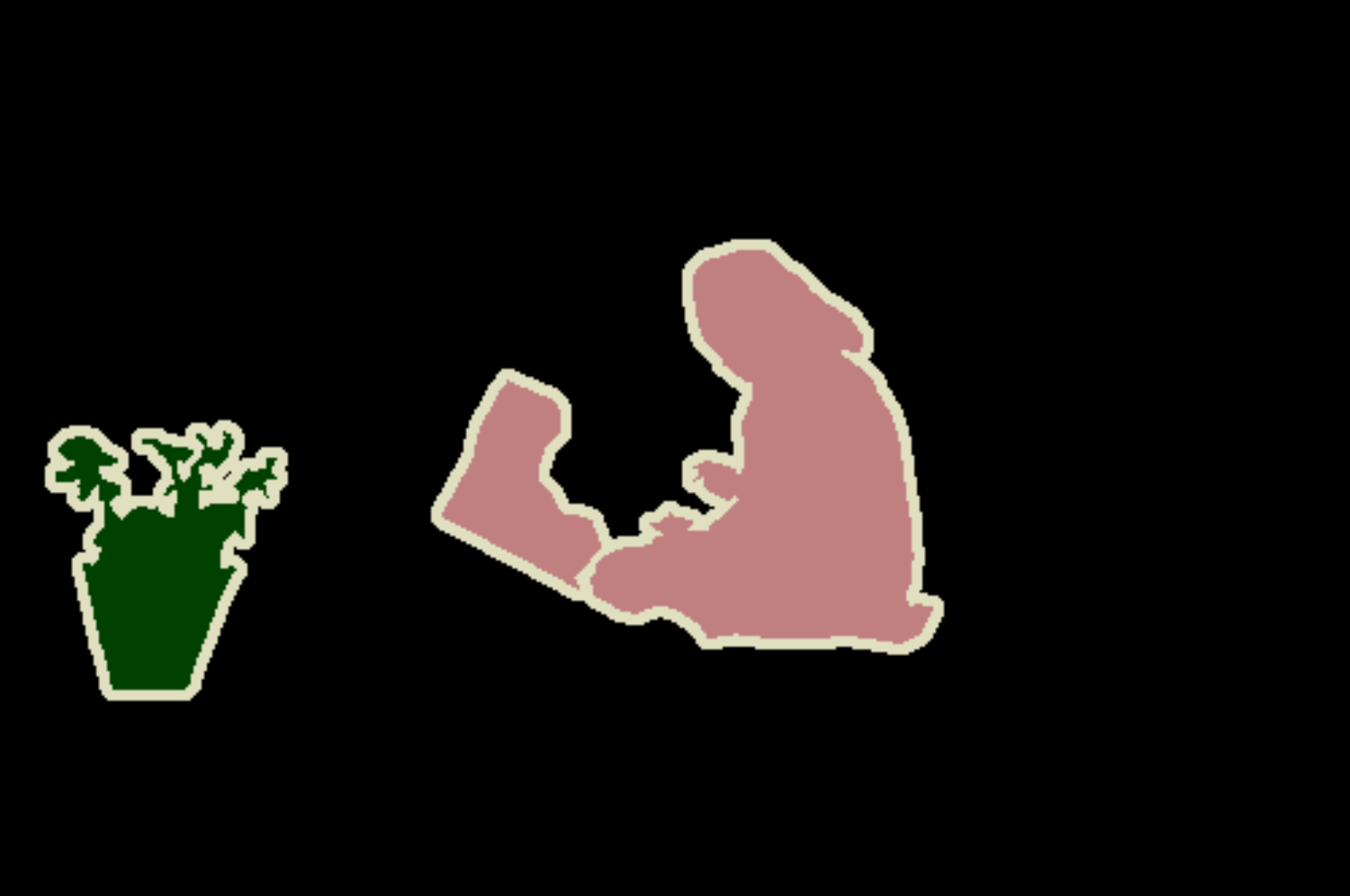}&\includegraphics[width=0.15\textwidth]{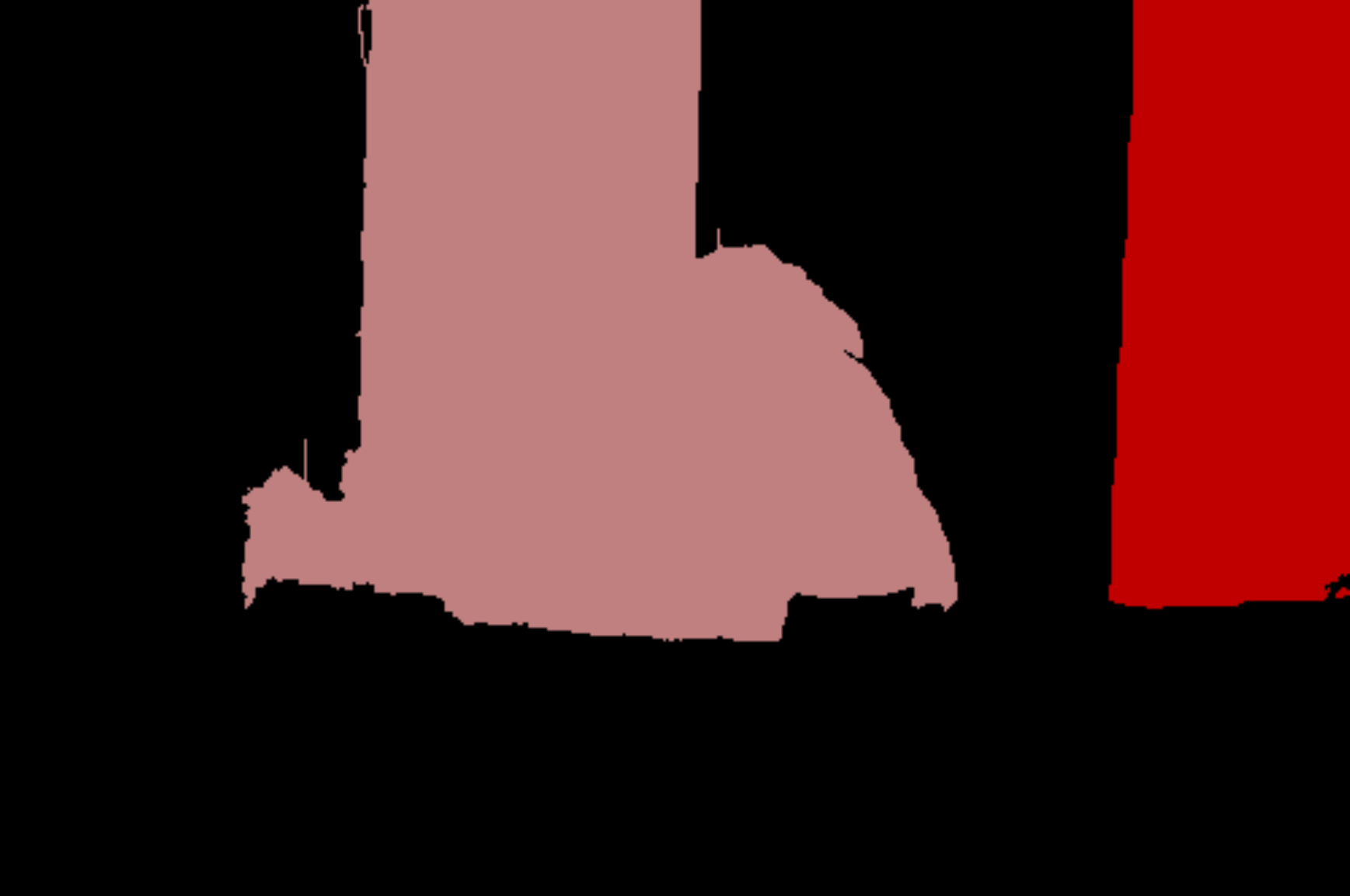}&\includegraphics[width=0.15\textwidth]{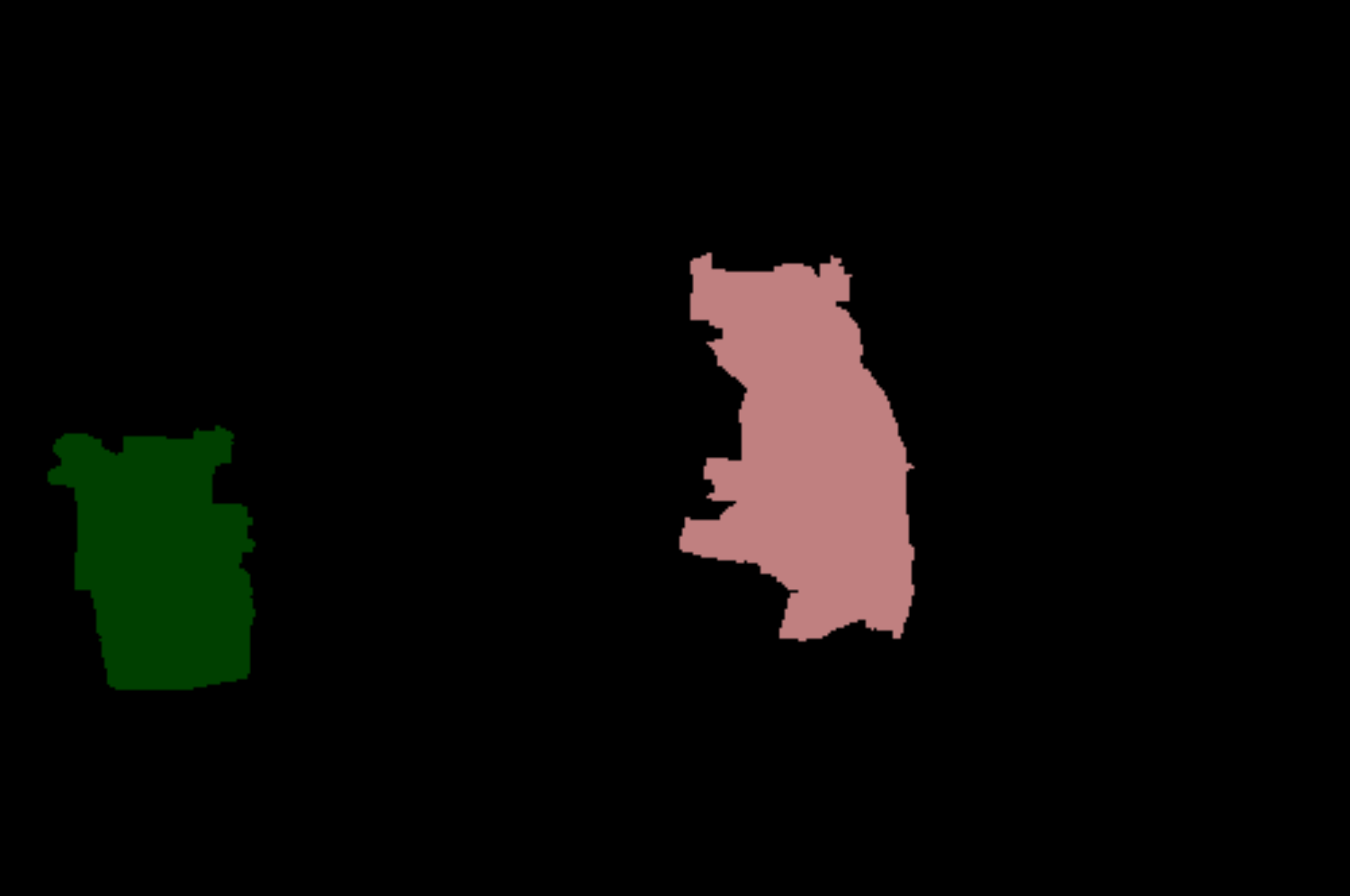}&\includegraphics[width=0.15\textwidth]{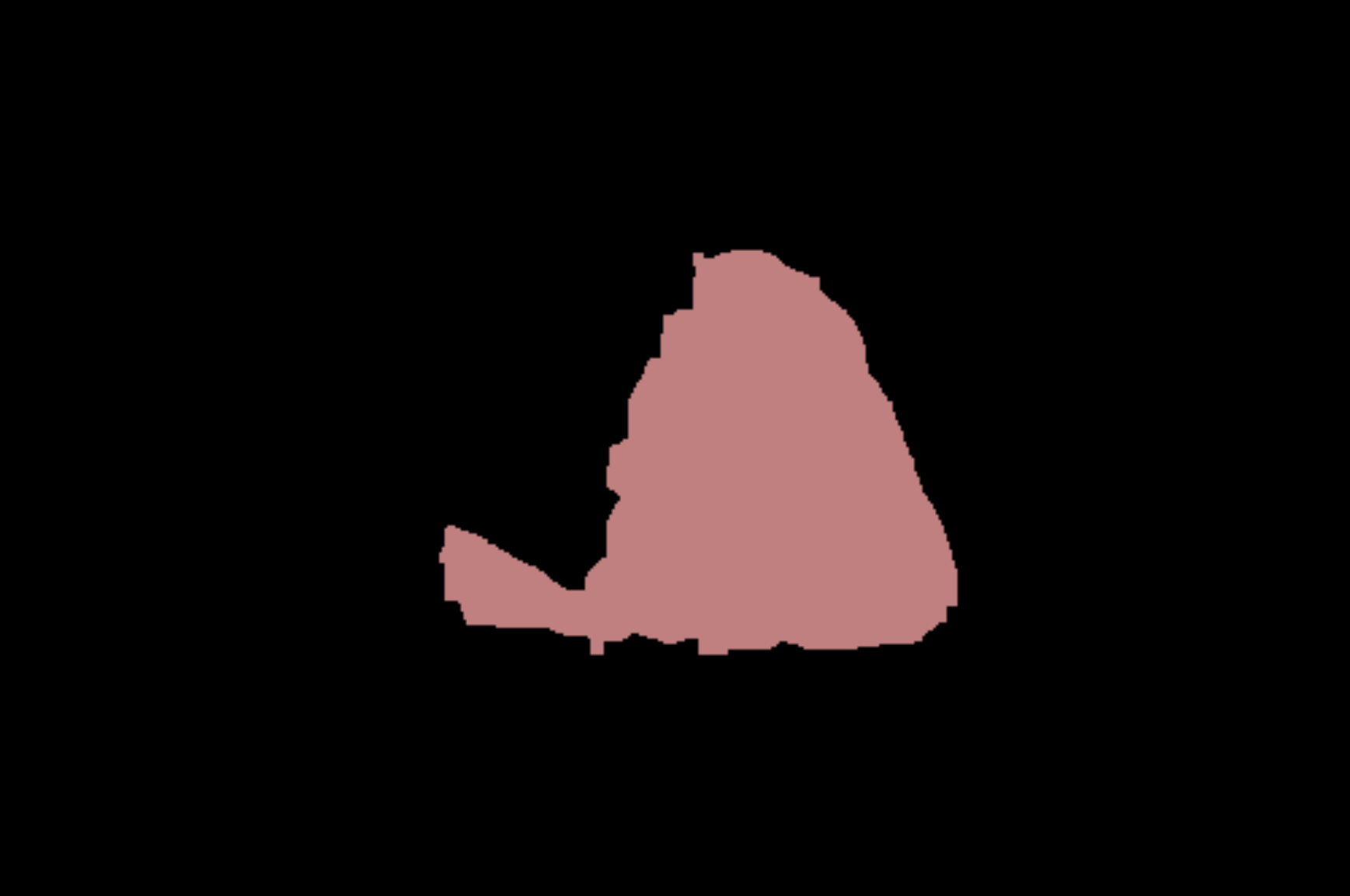}&\includegraphics[width=0.15\textwidth]{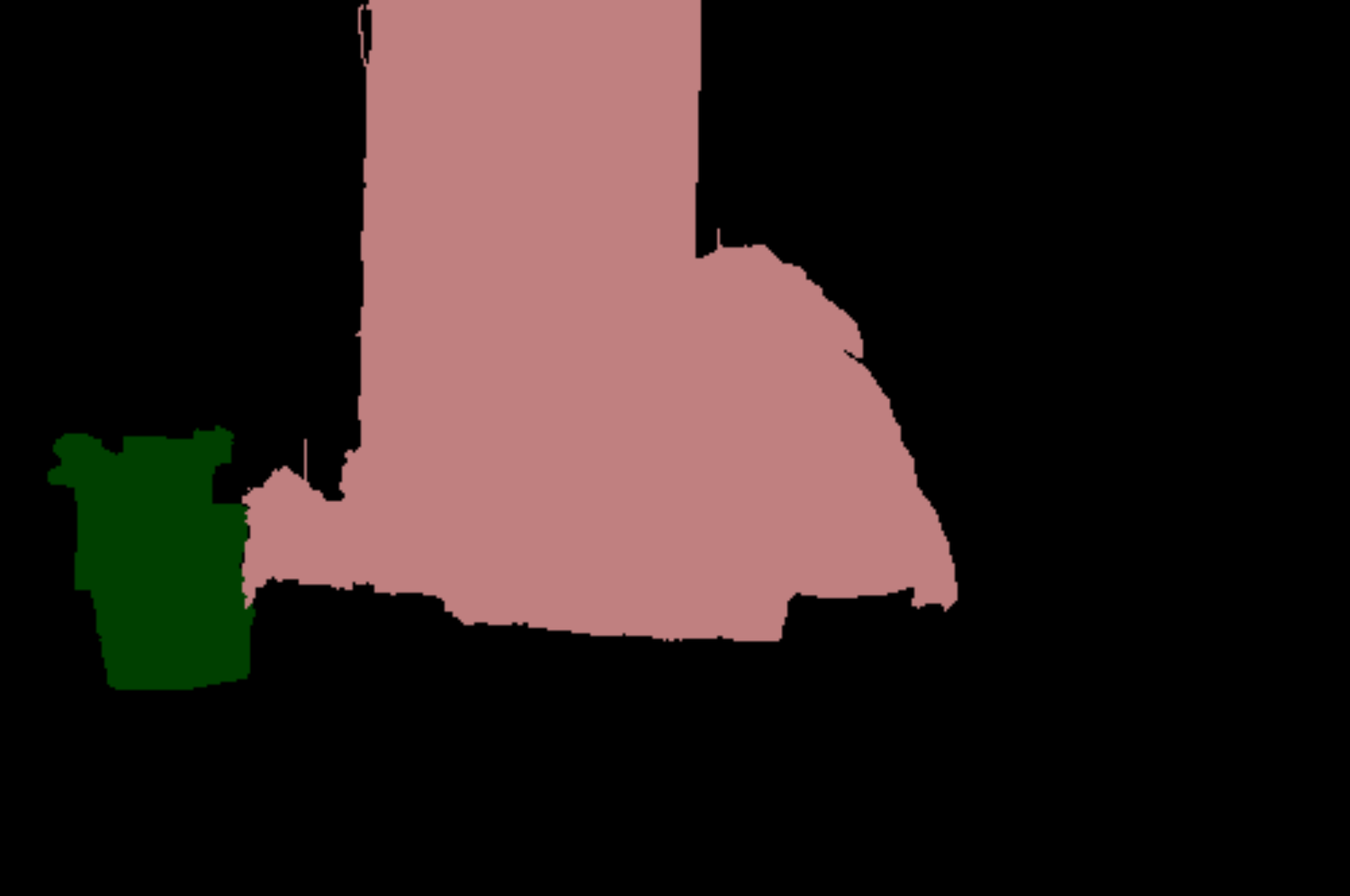}\\
		\includegraphics[width=0.15\textwidth]{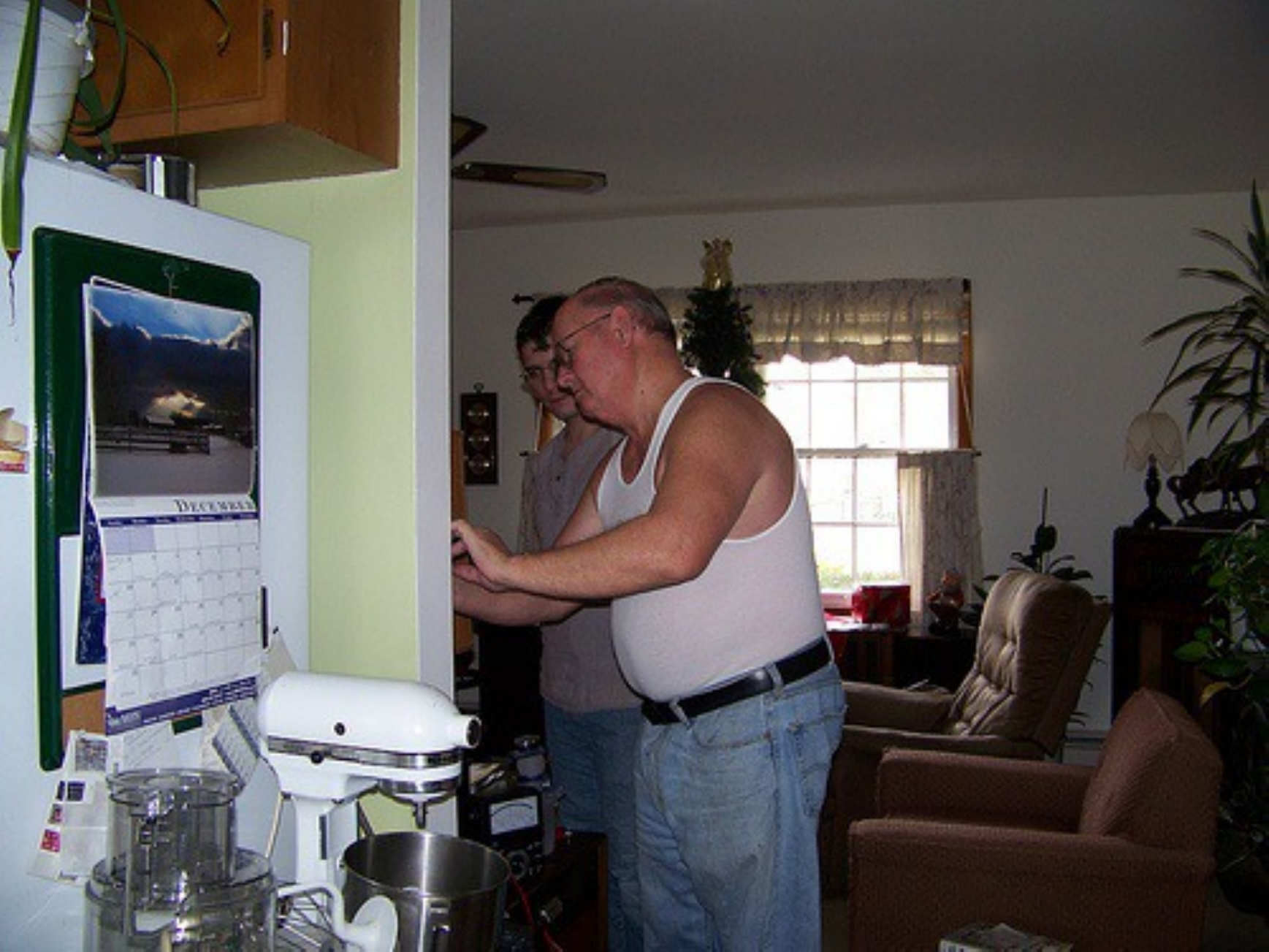}&\includegraphics[width=0.15\textwidth]{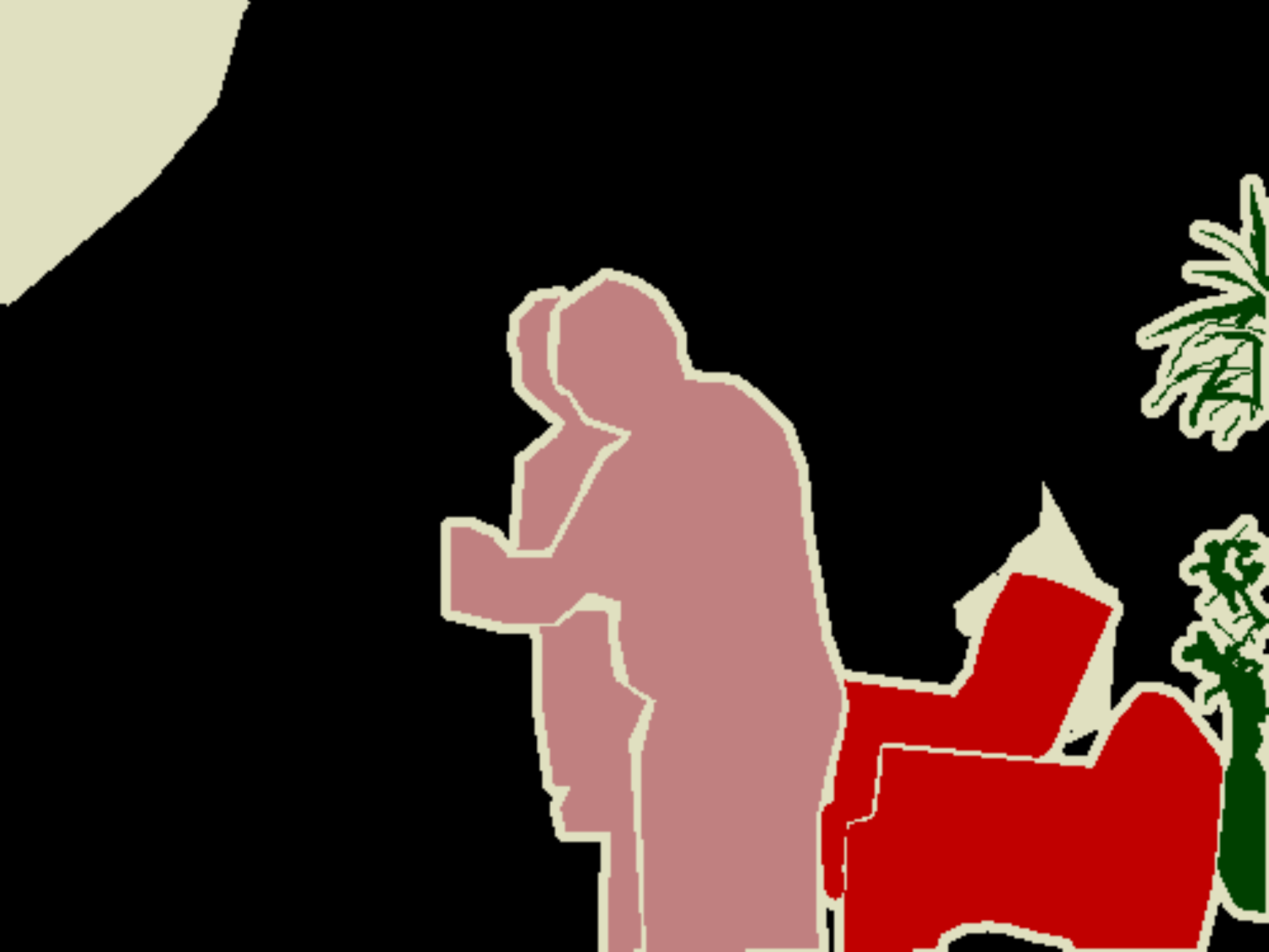}&\includegraphics[width=0.15\textwidth]{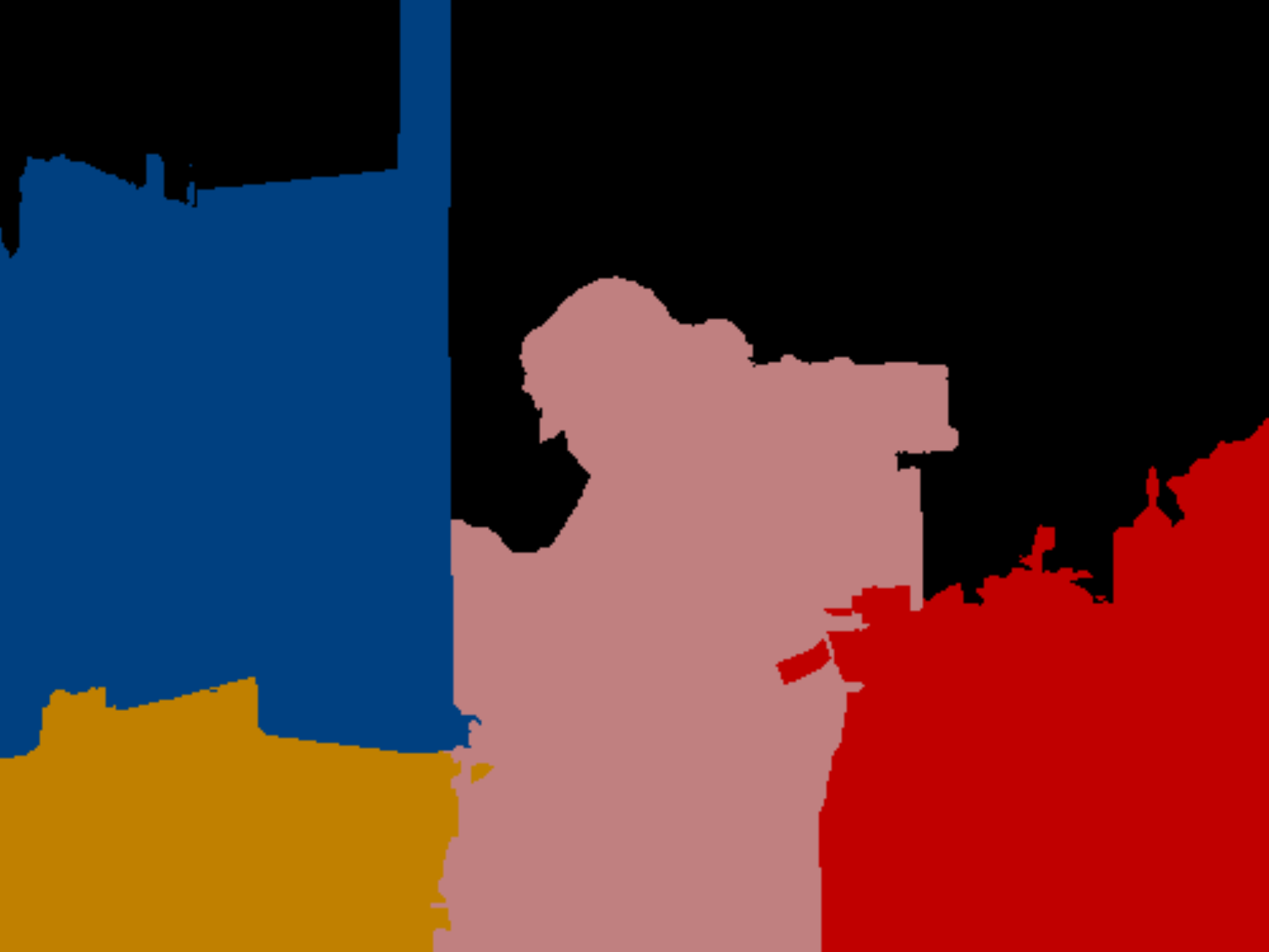}&\includegraphics[width=0.15\textwidth]{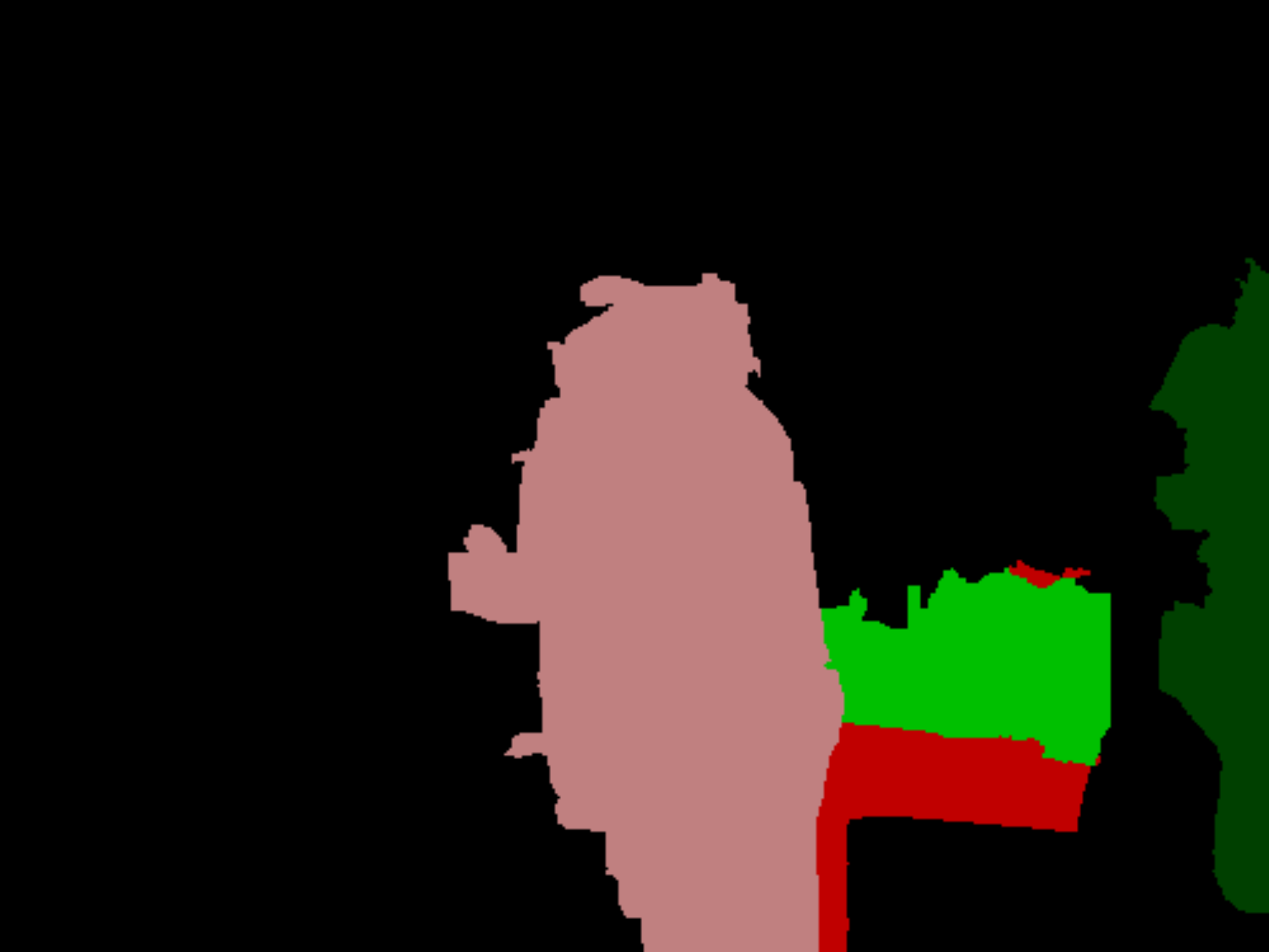}&\includegraphics[width=0.15\textwidth]{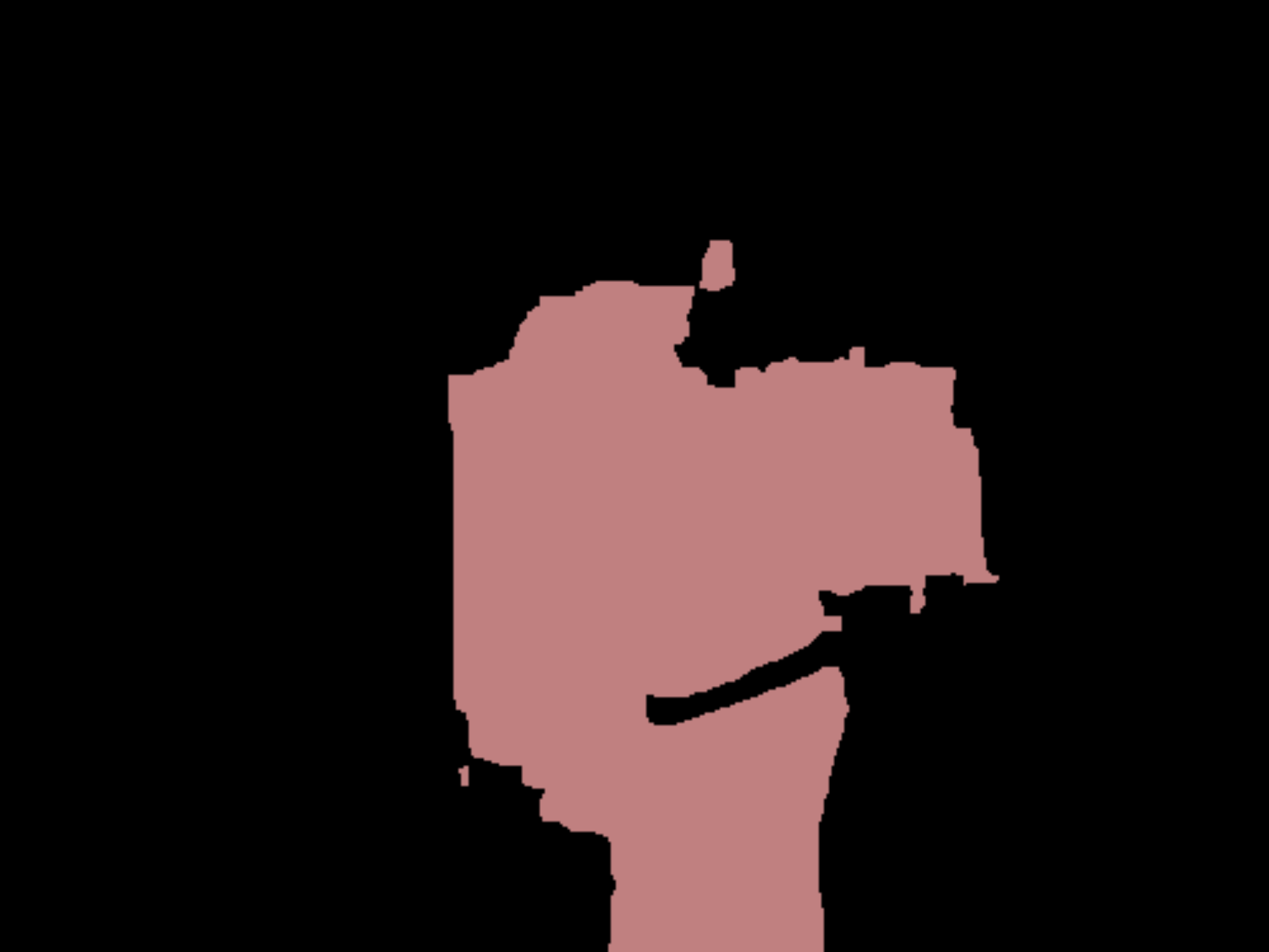}&\includegraphics[width=0.15\textwidth]{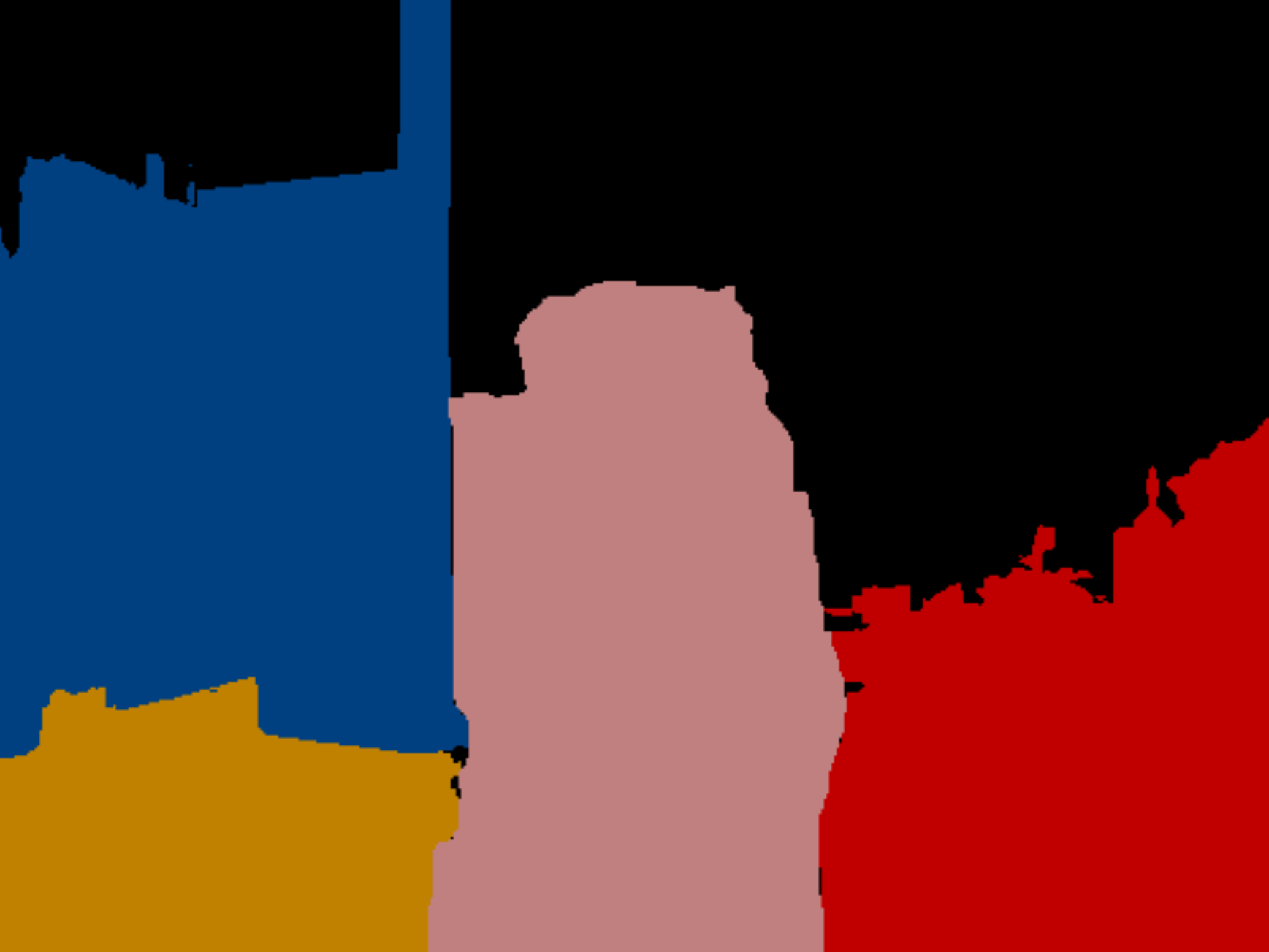}\\
		\includegraphics[width=0.15\textwidth]{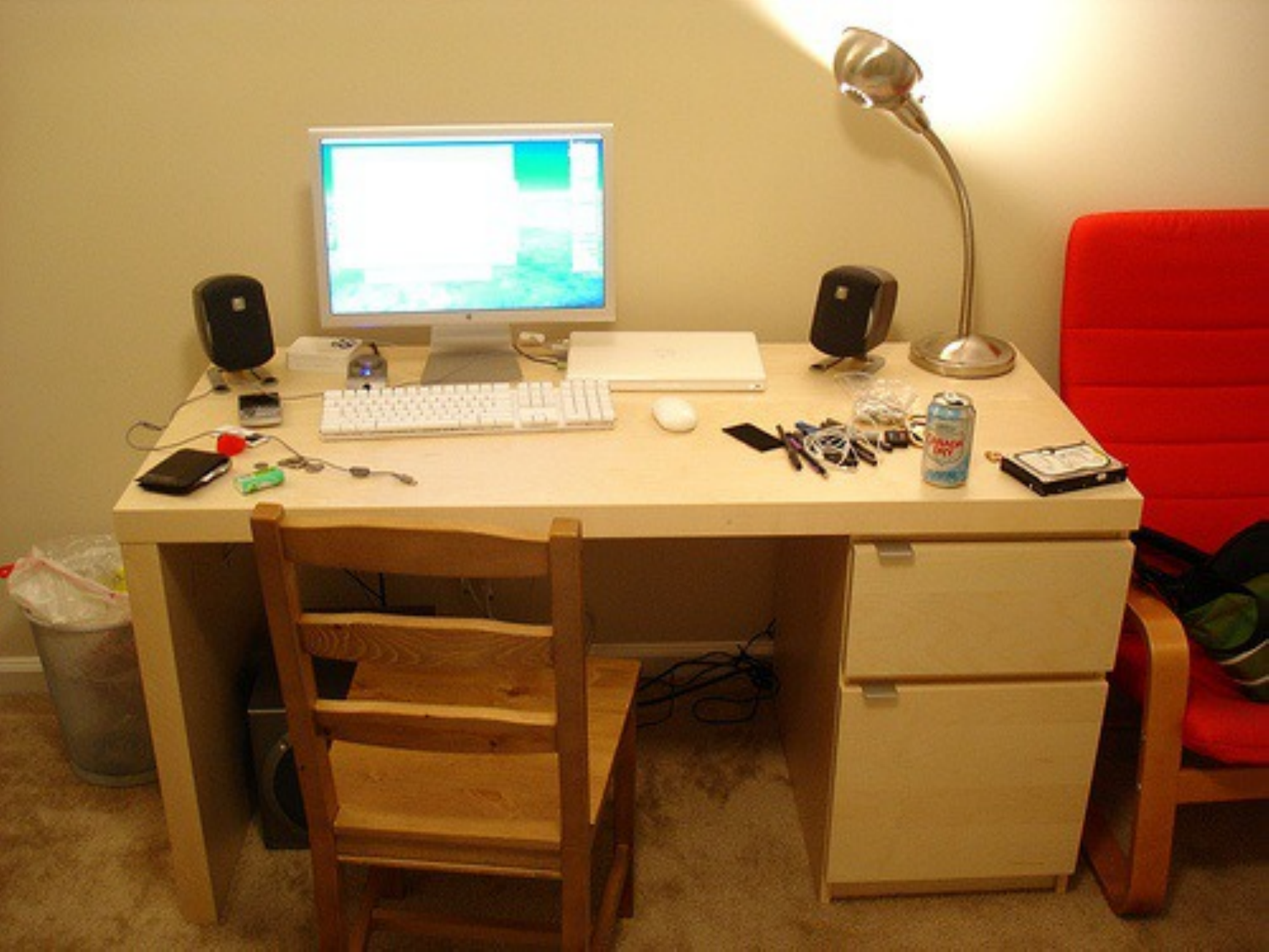}&\includegraphics[width=0.15\textwidth]{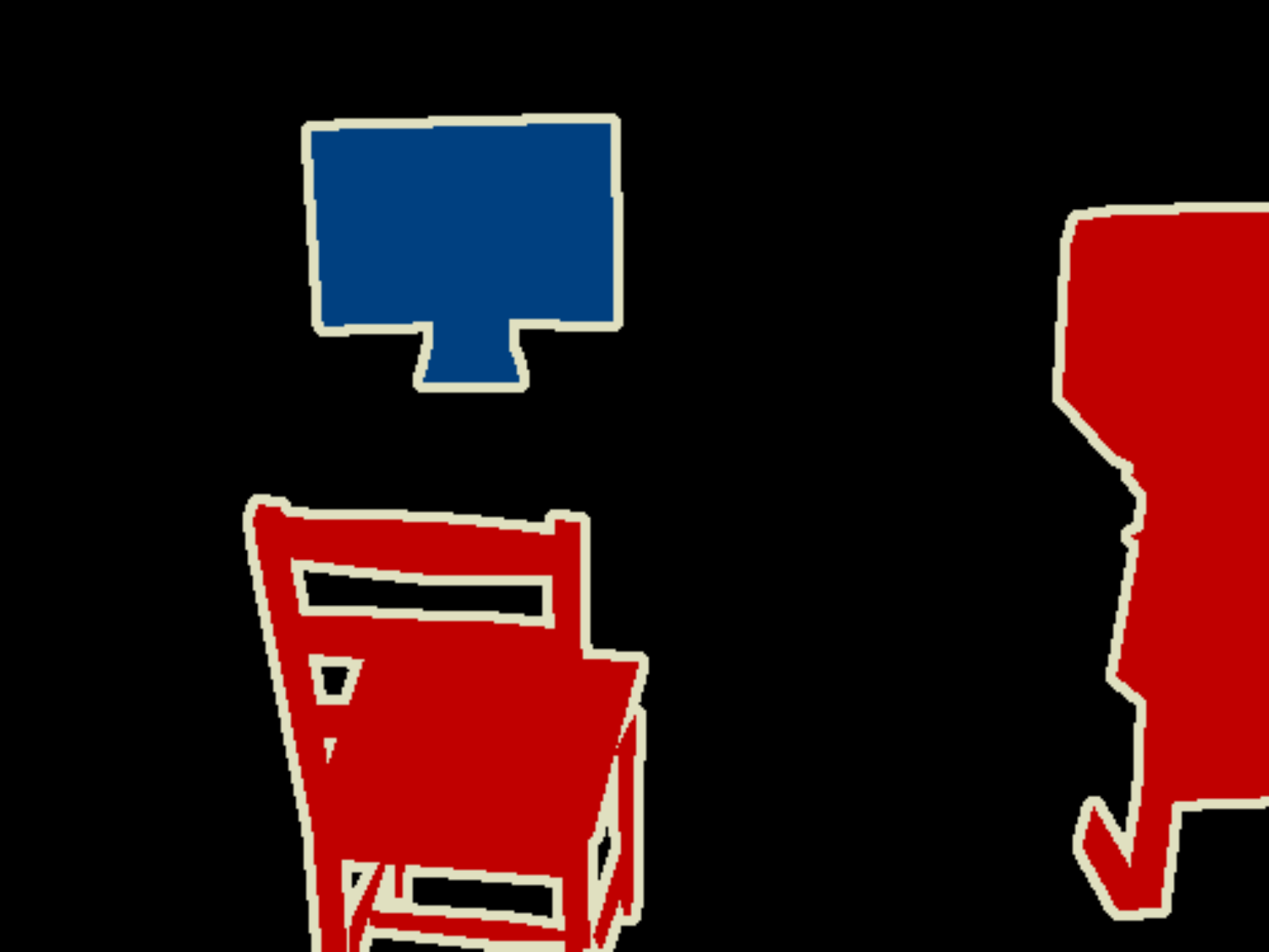}&\includegraphics[width=0.15\textwidth]{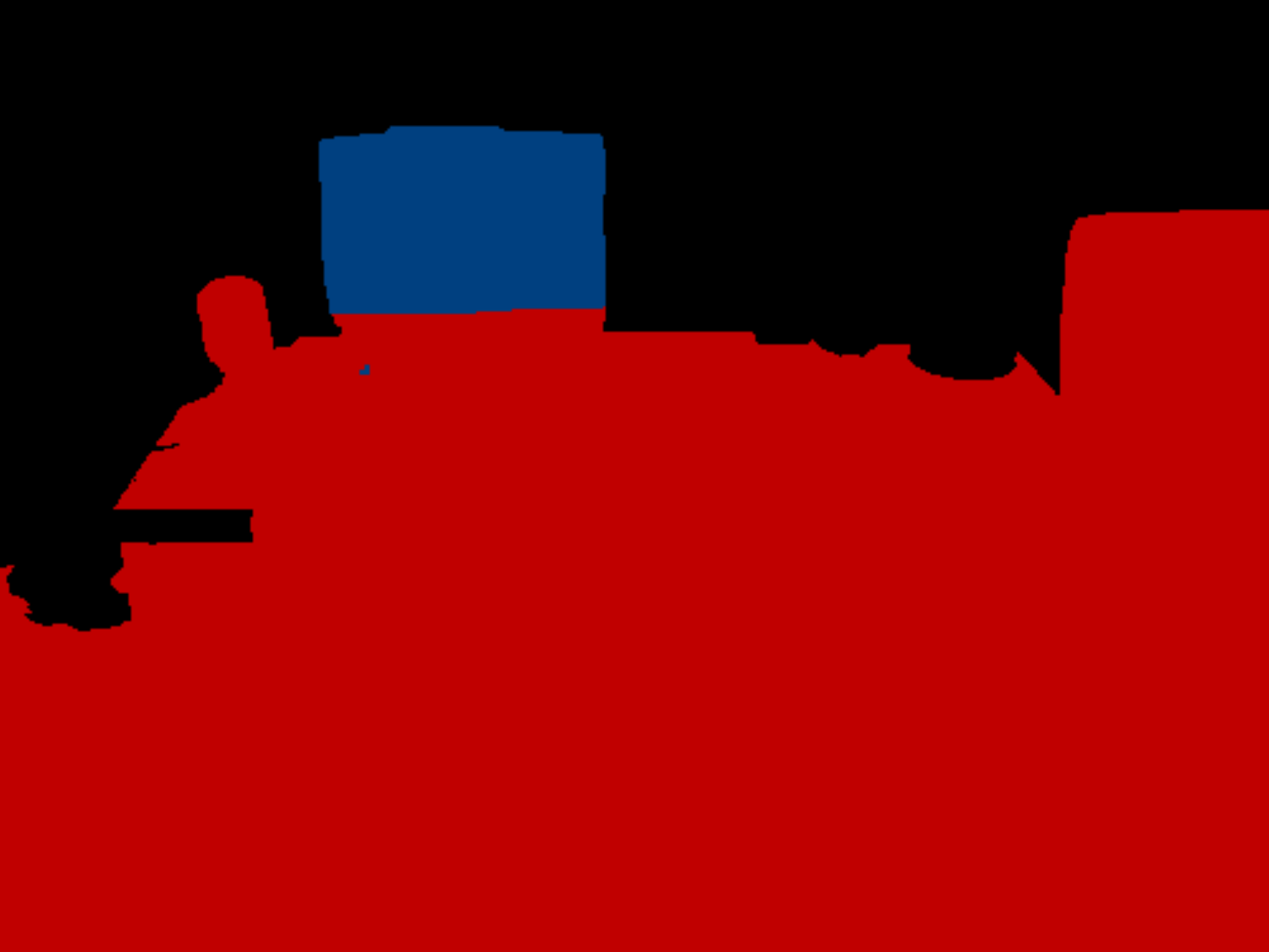}&\includegraphics[width=0.15\textwidth]{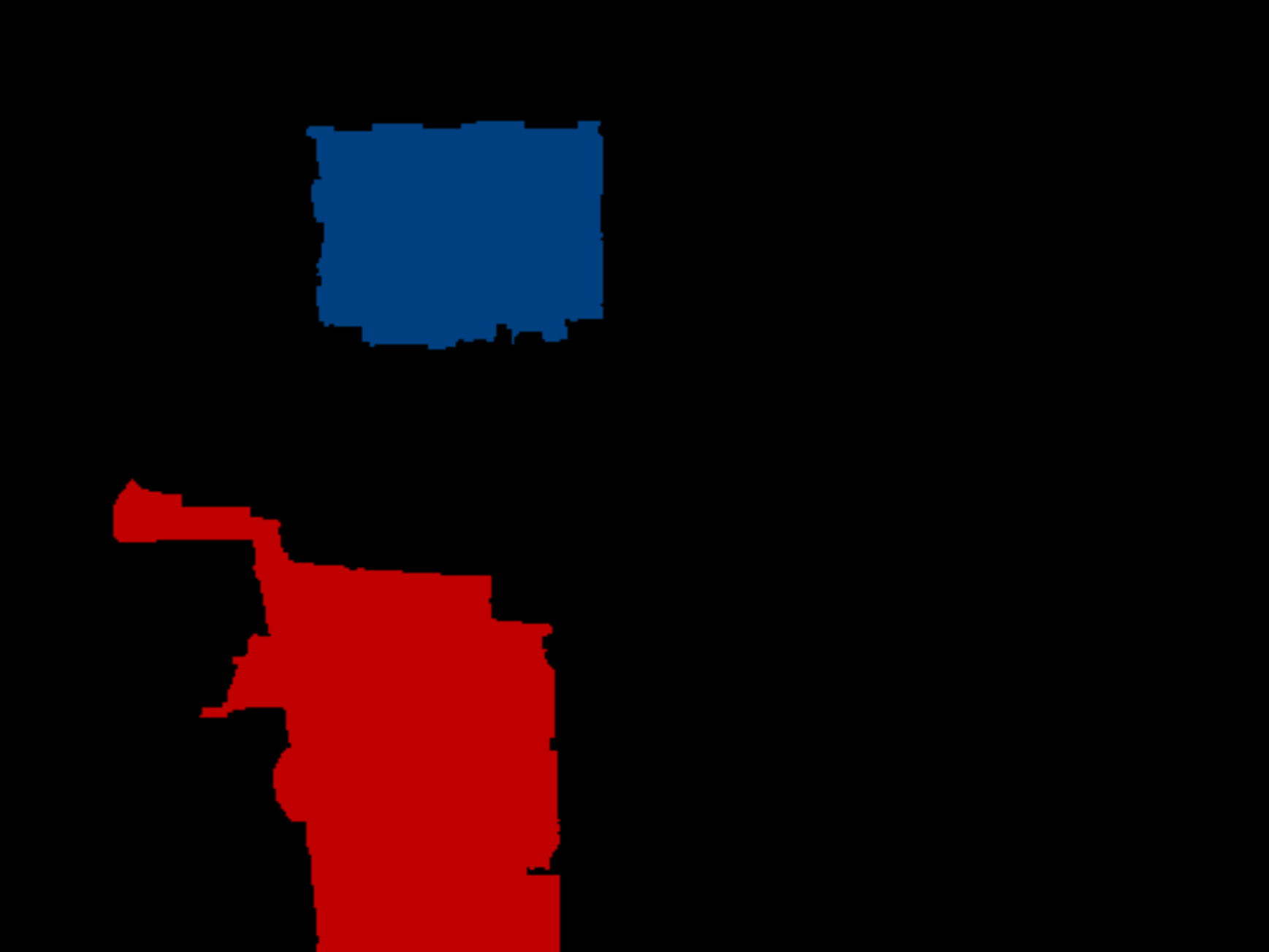}&\includegraphics[width=0.15\textwidth]{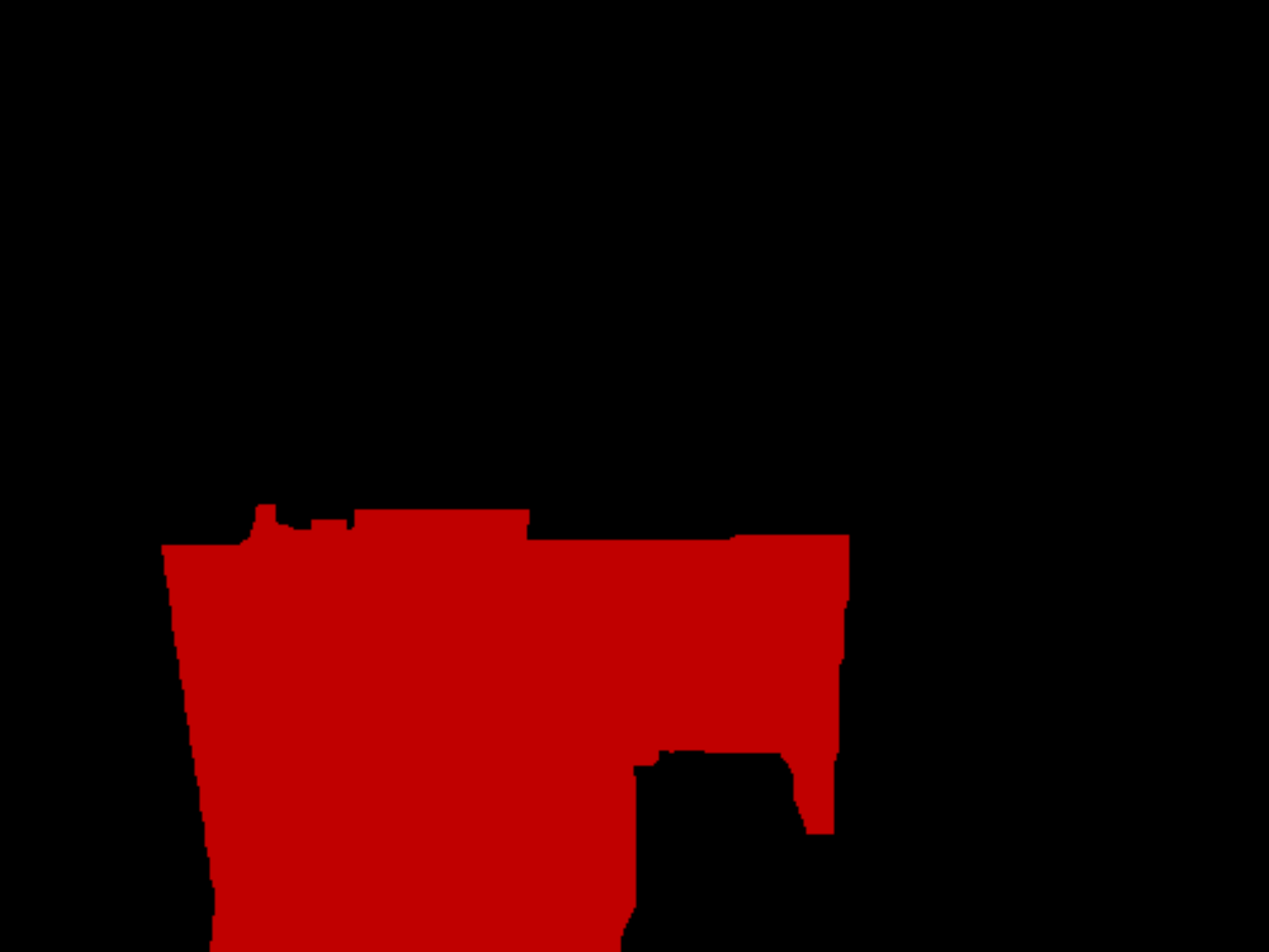}&\includegraphics[width=0.15\textwidth]{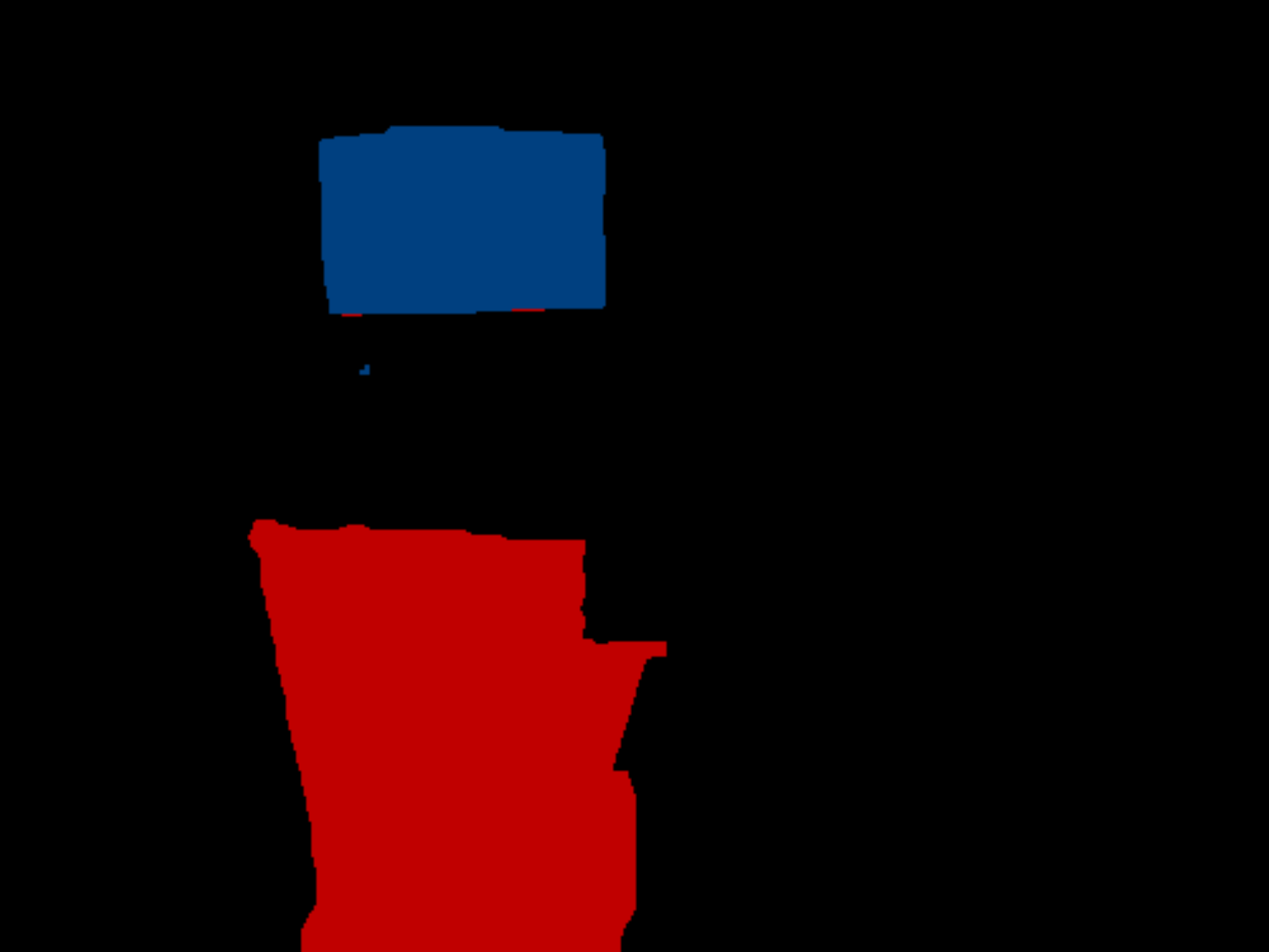}\\
		\includegraphics[width=0.15\textwidth]{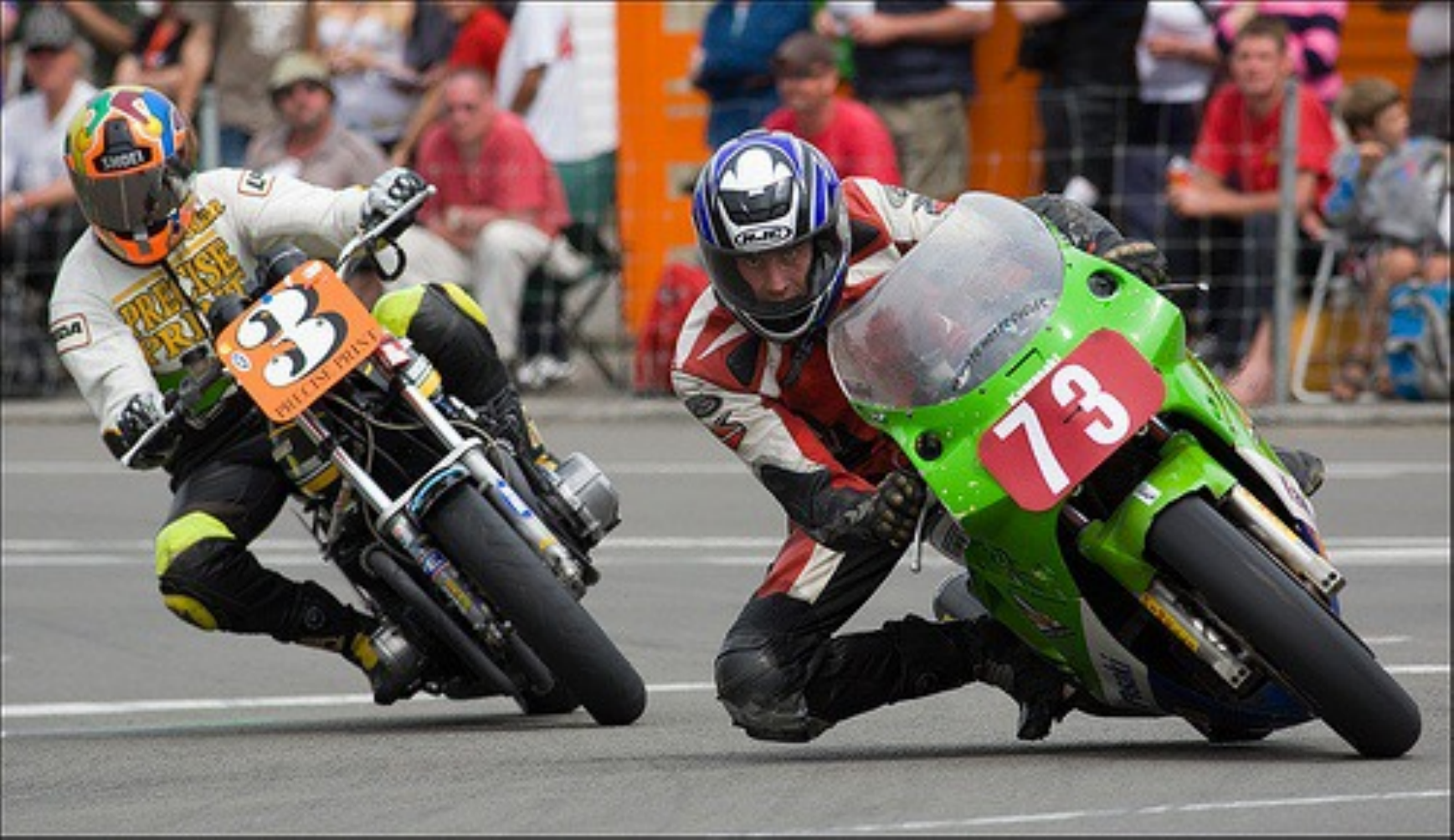}&\includegraphics[width=0.15\textwidth]{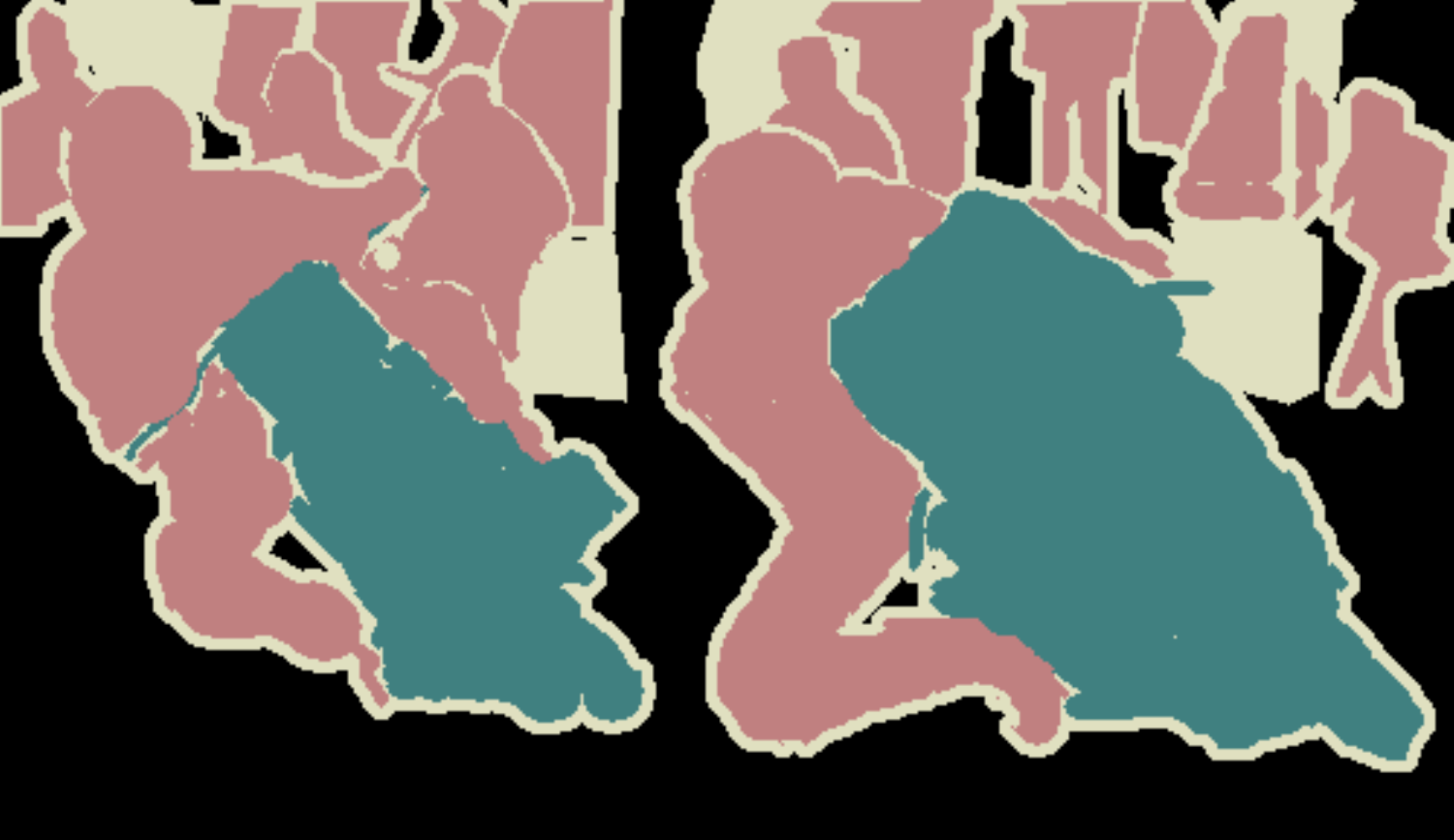}&\includegraphics[width=0.15\textwidth]{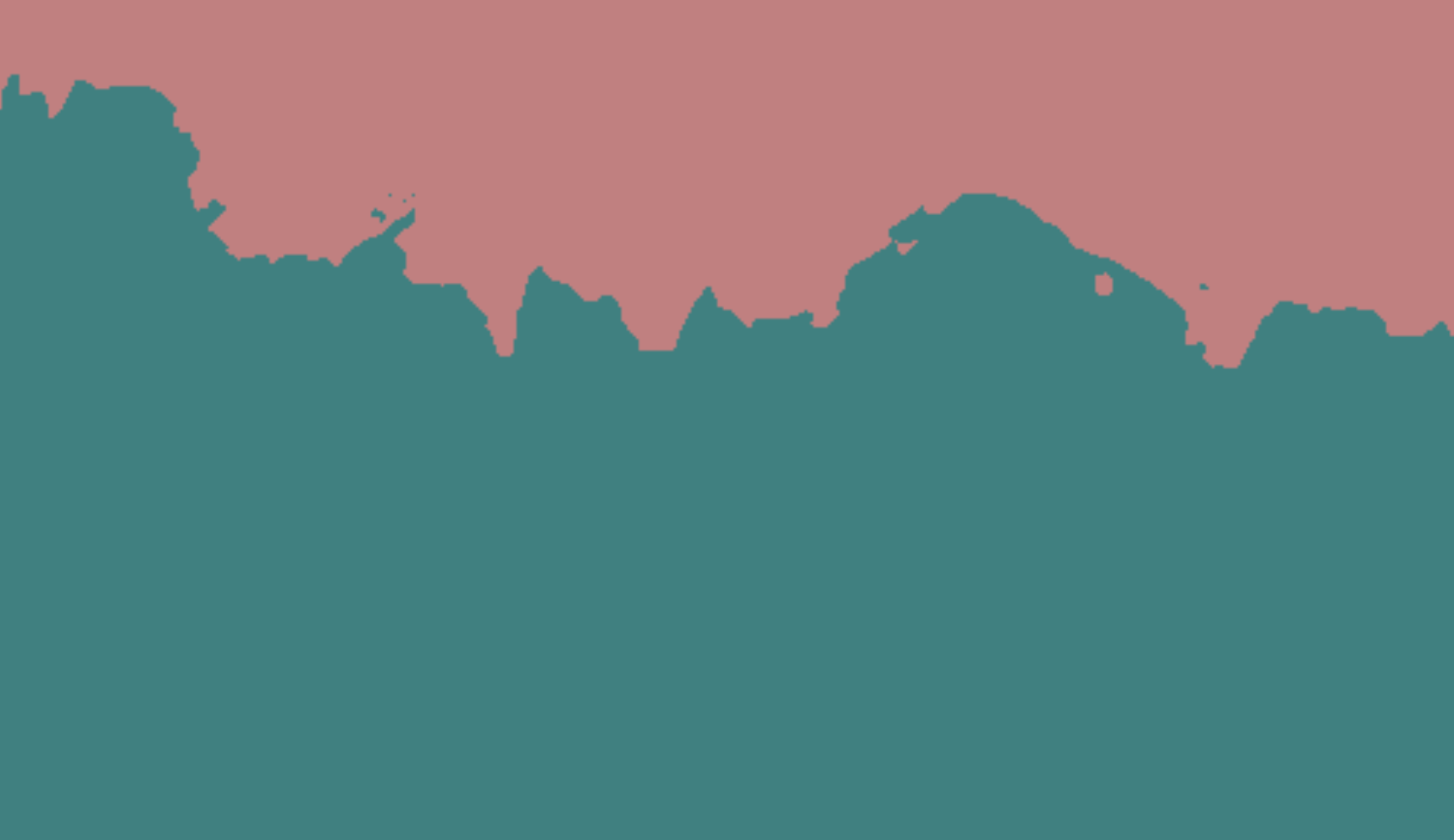}&\includegraphics[width=0.15\textwidth]{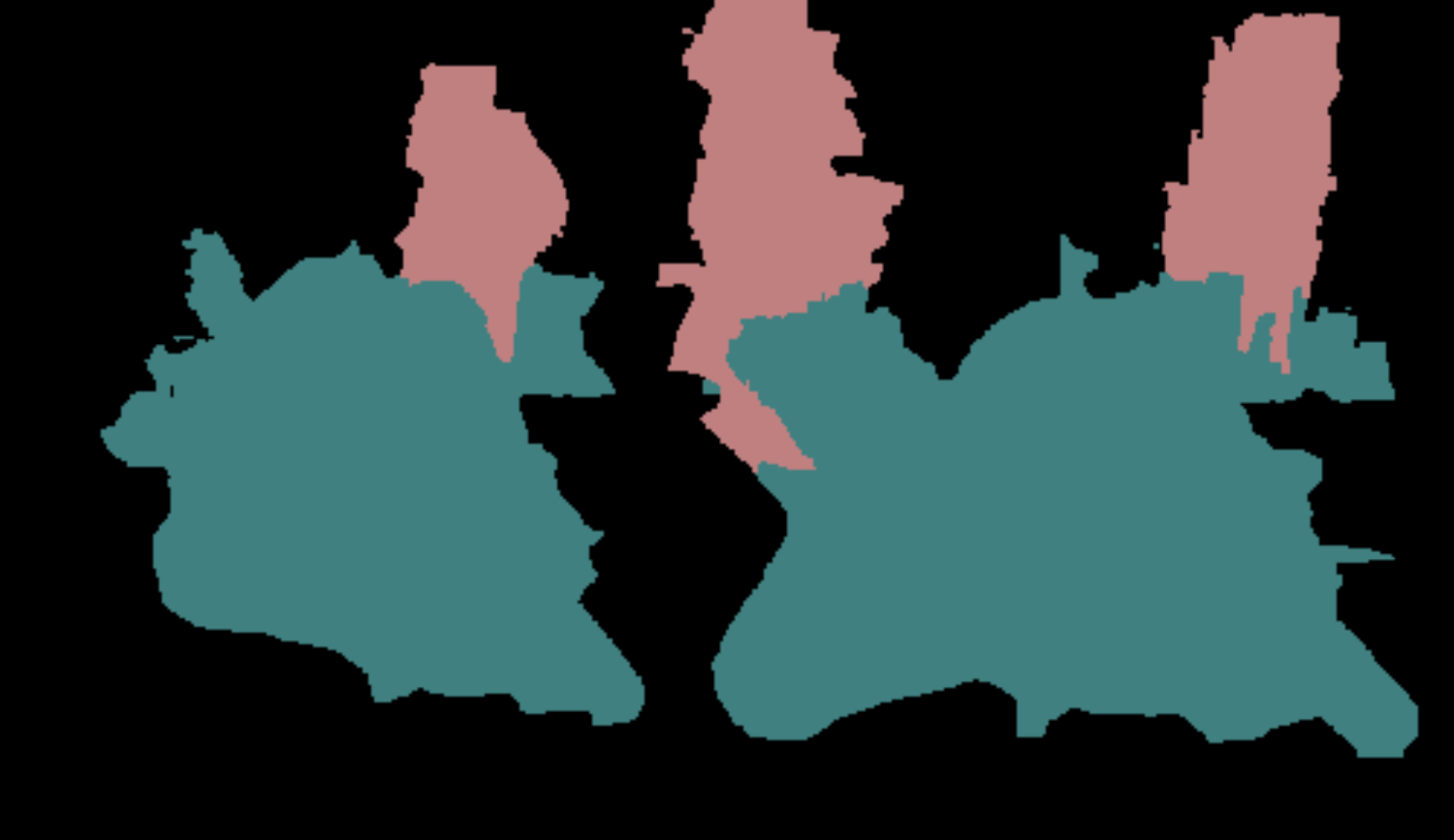}&\includegraphics[width=0.15\textwidth]{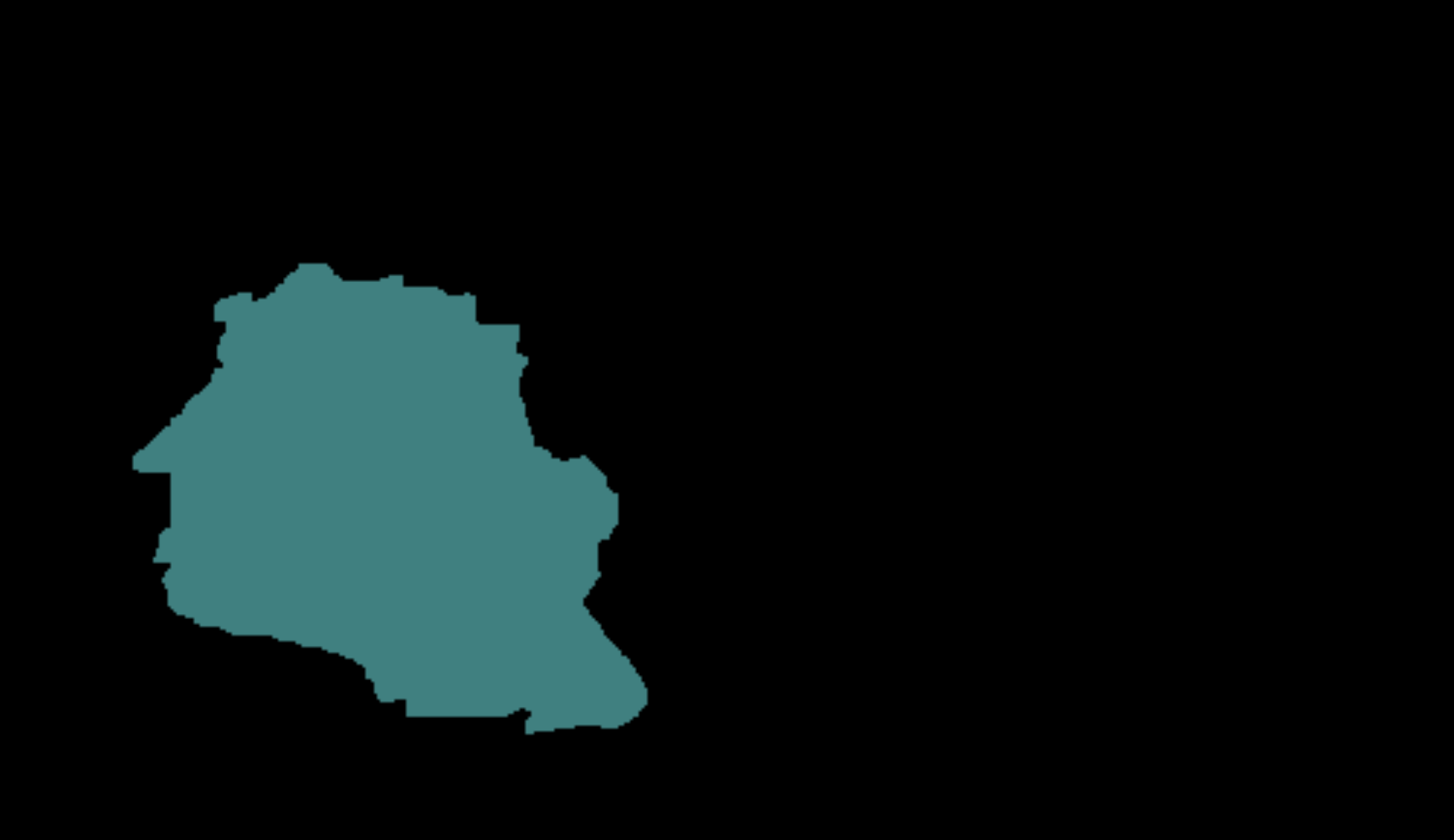}&\includegraphics[width=0.15\textwidth]{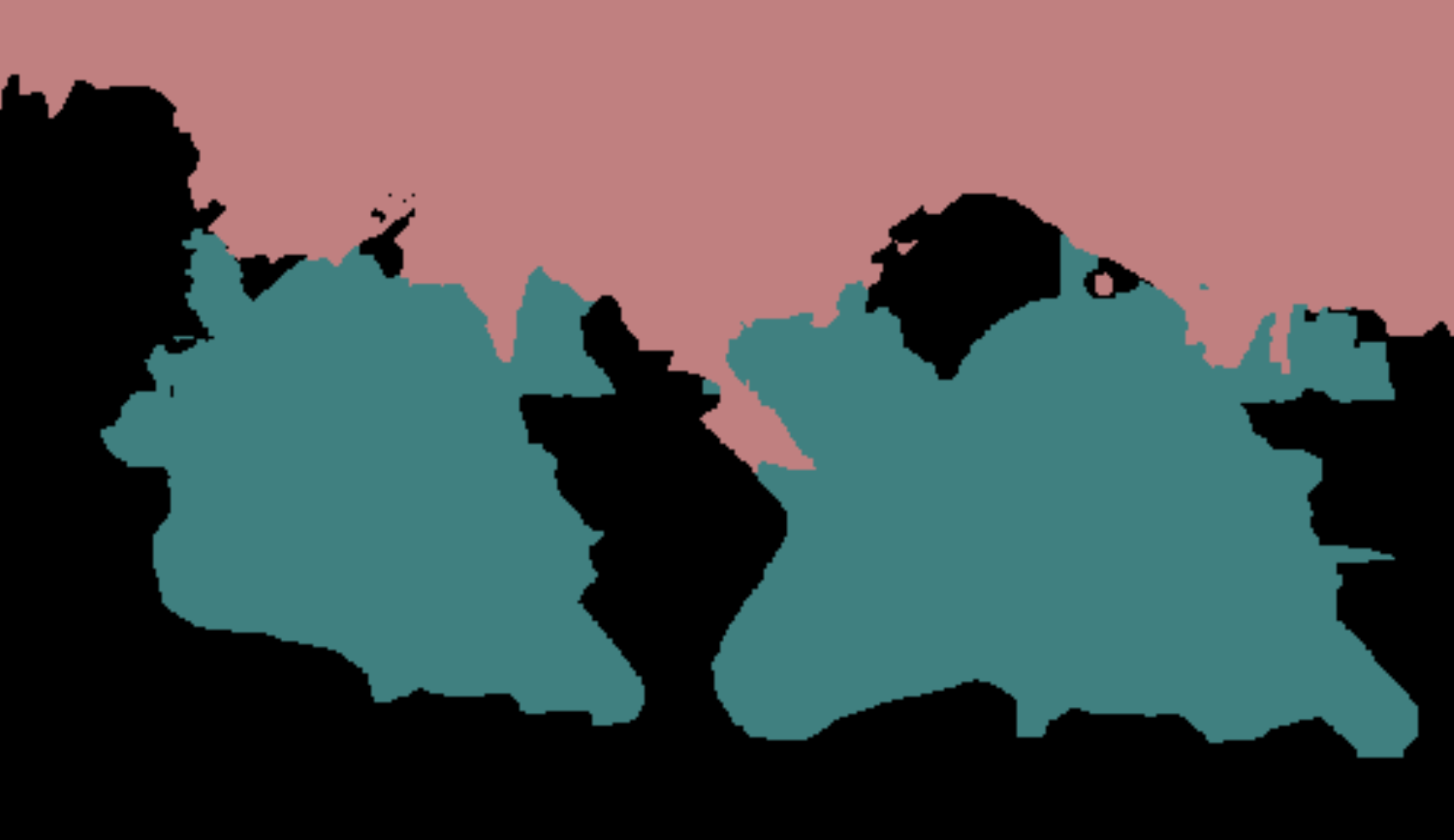}\\
		\includegraphics[width=0.15\textwidth]{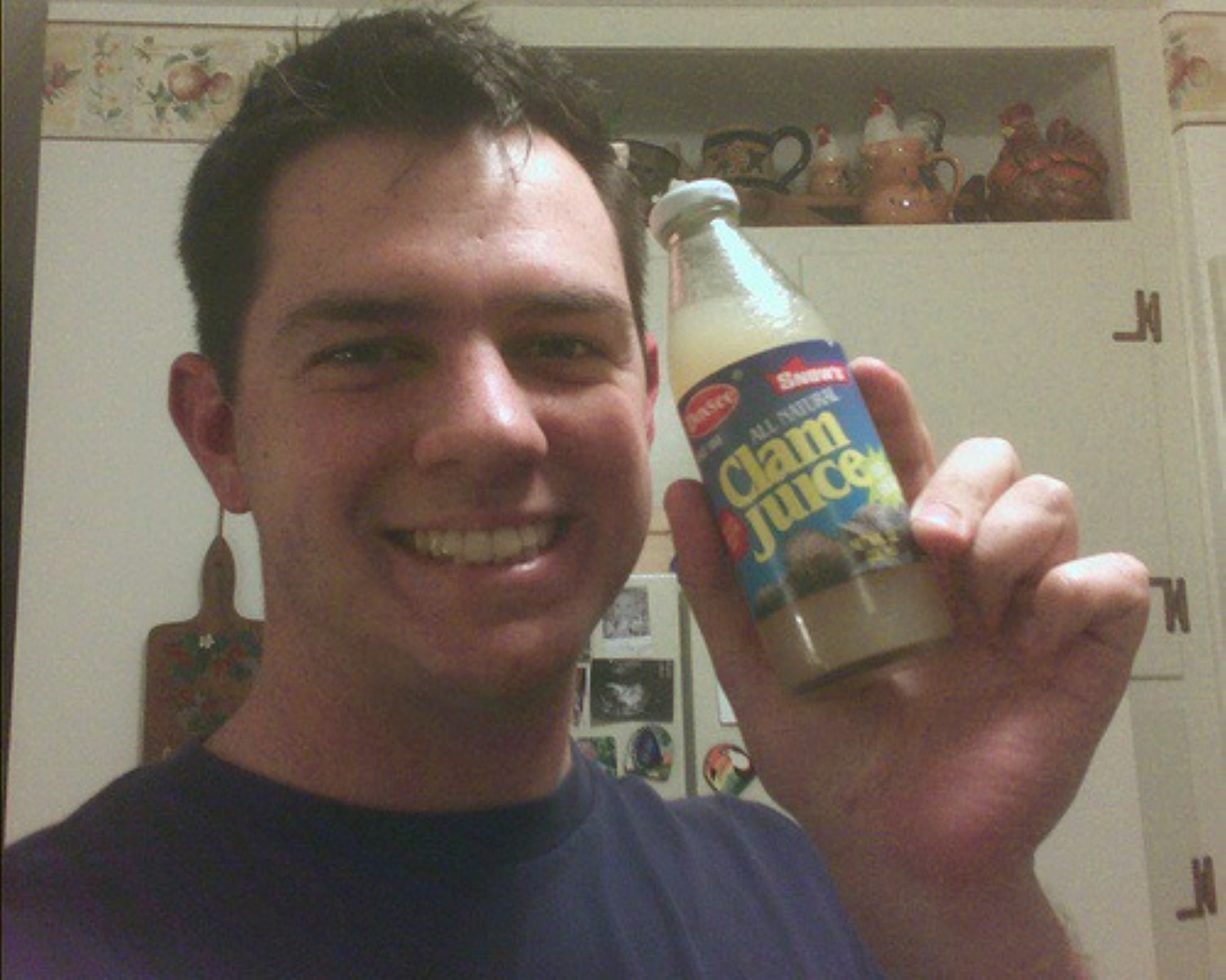}&\includegraphics[width=0.15\textwidth]{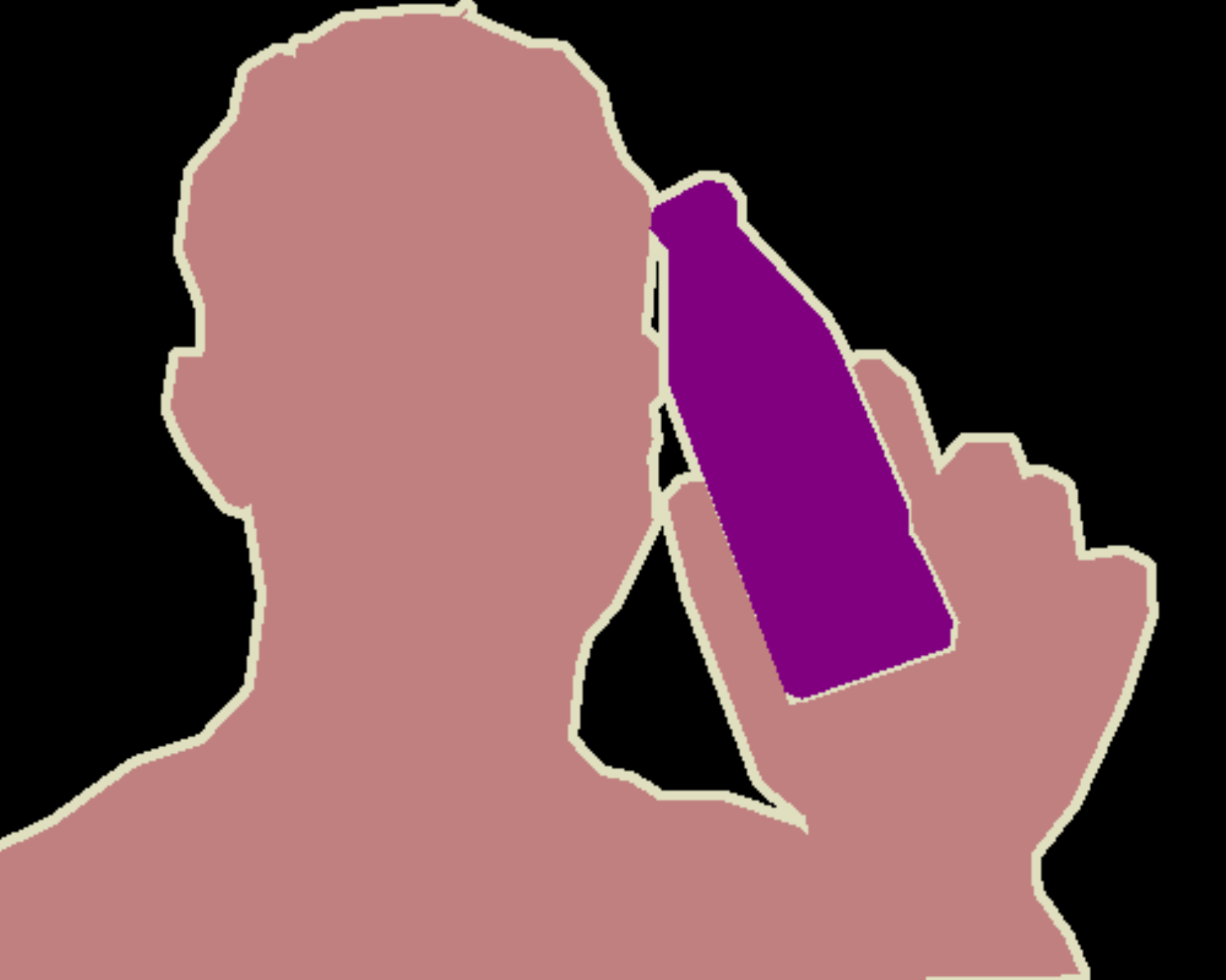}&\includegraphics[width=0.15\textwidth]{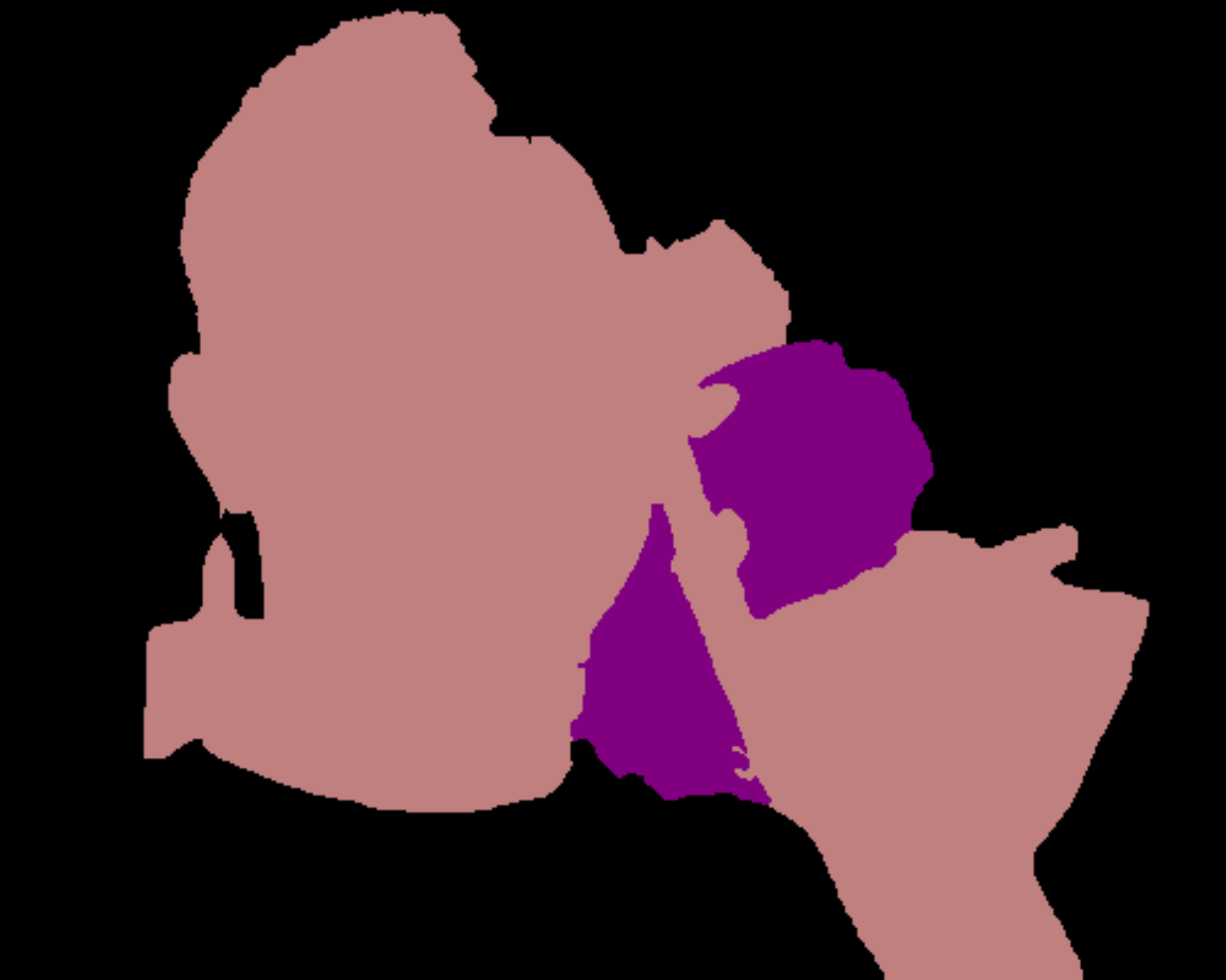}&\includegraphics[width=0.15\textwidth]{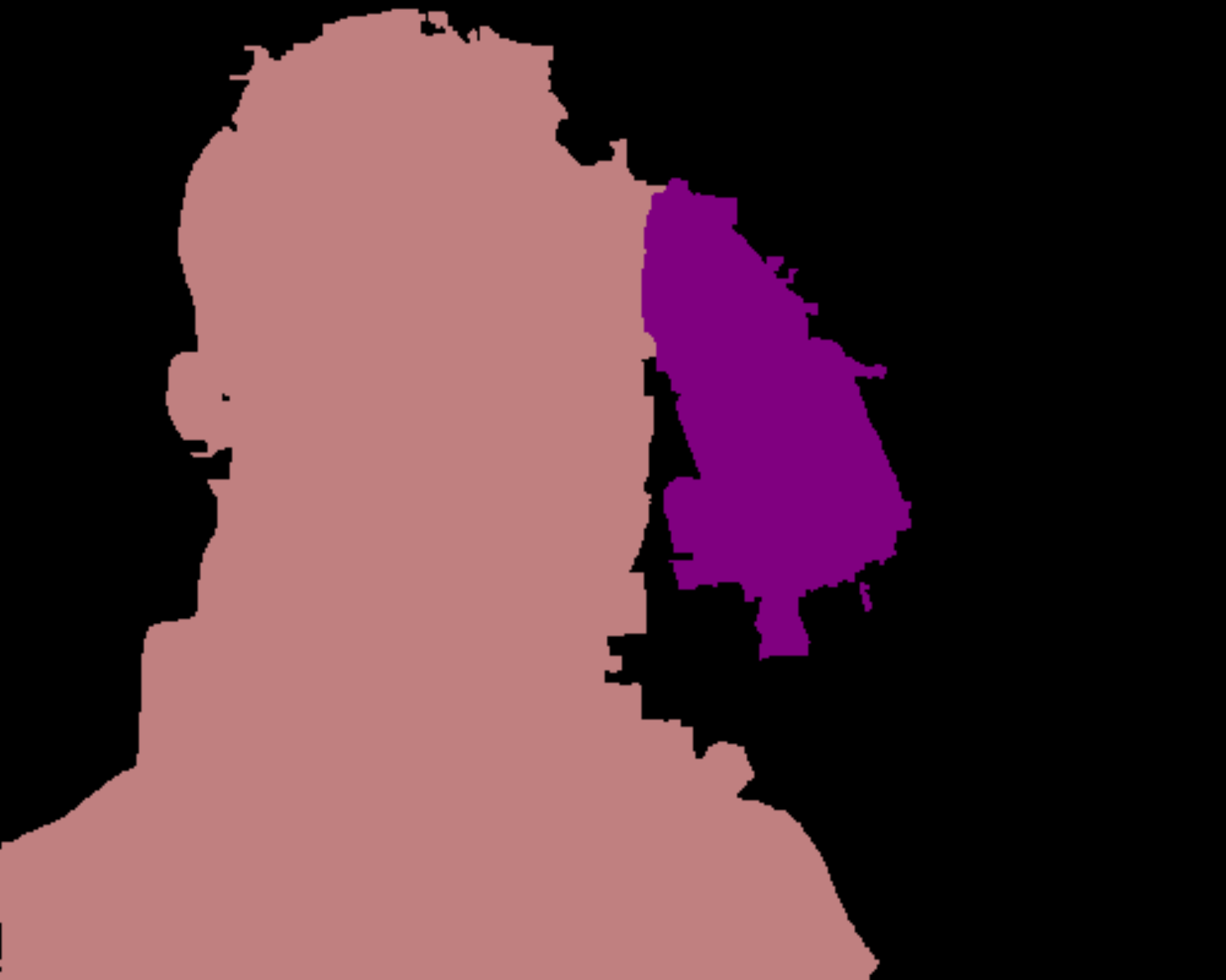}&\includegraphics[width=0.15\textwidth]{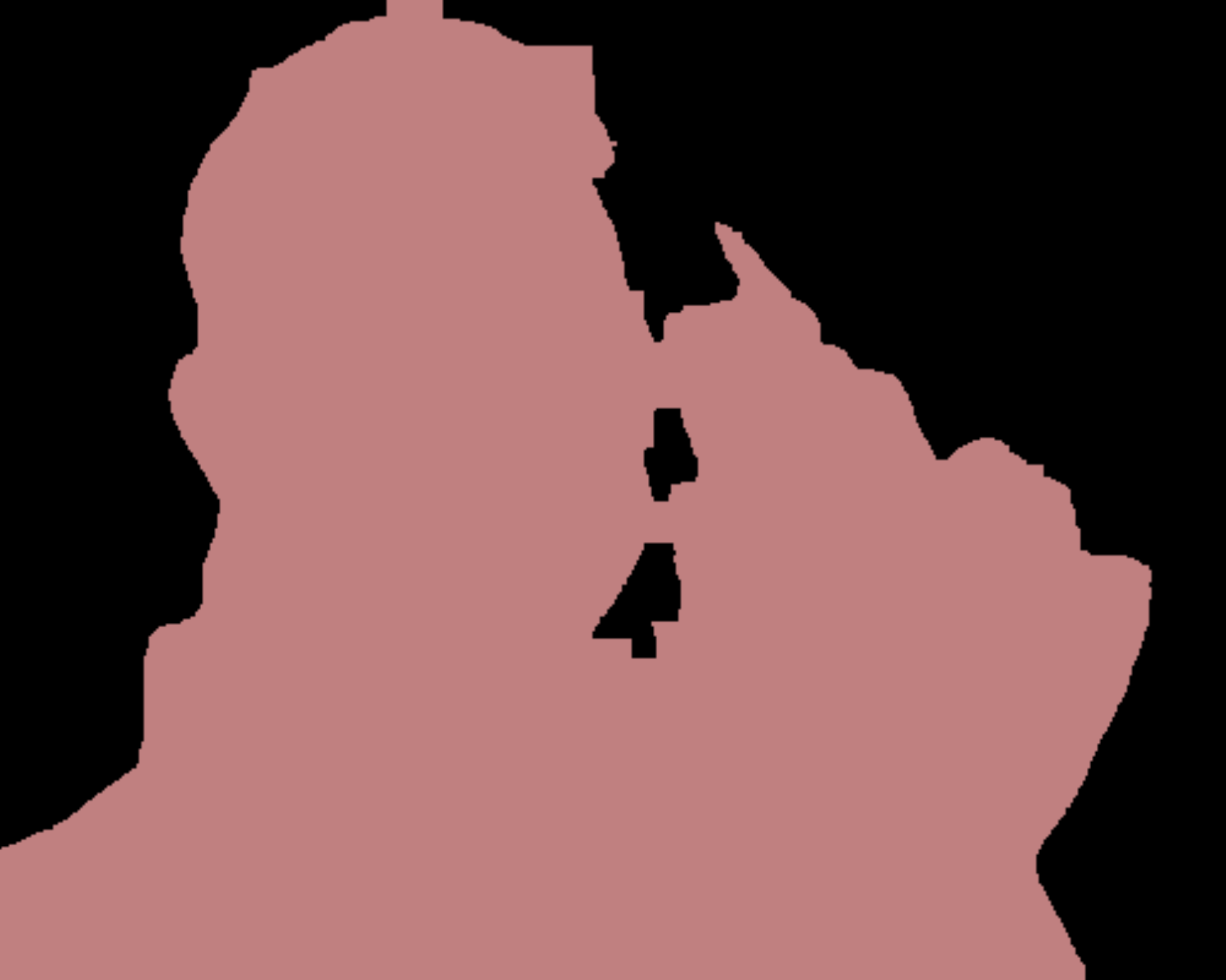}&\includegraphics[width=0.15\textwidth]{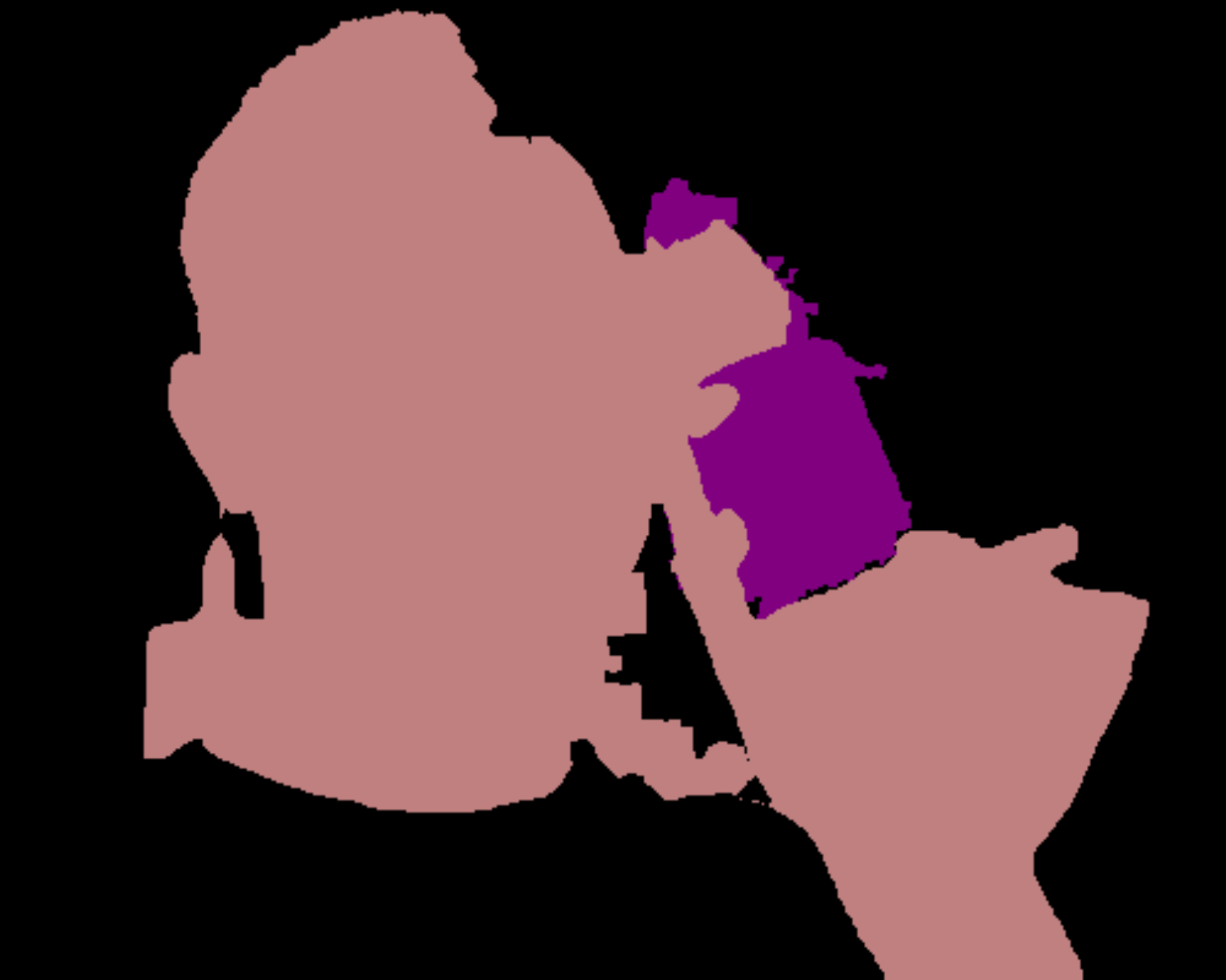}\\
		\includegraphics[width=0.15\textwidth]{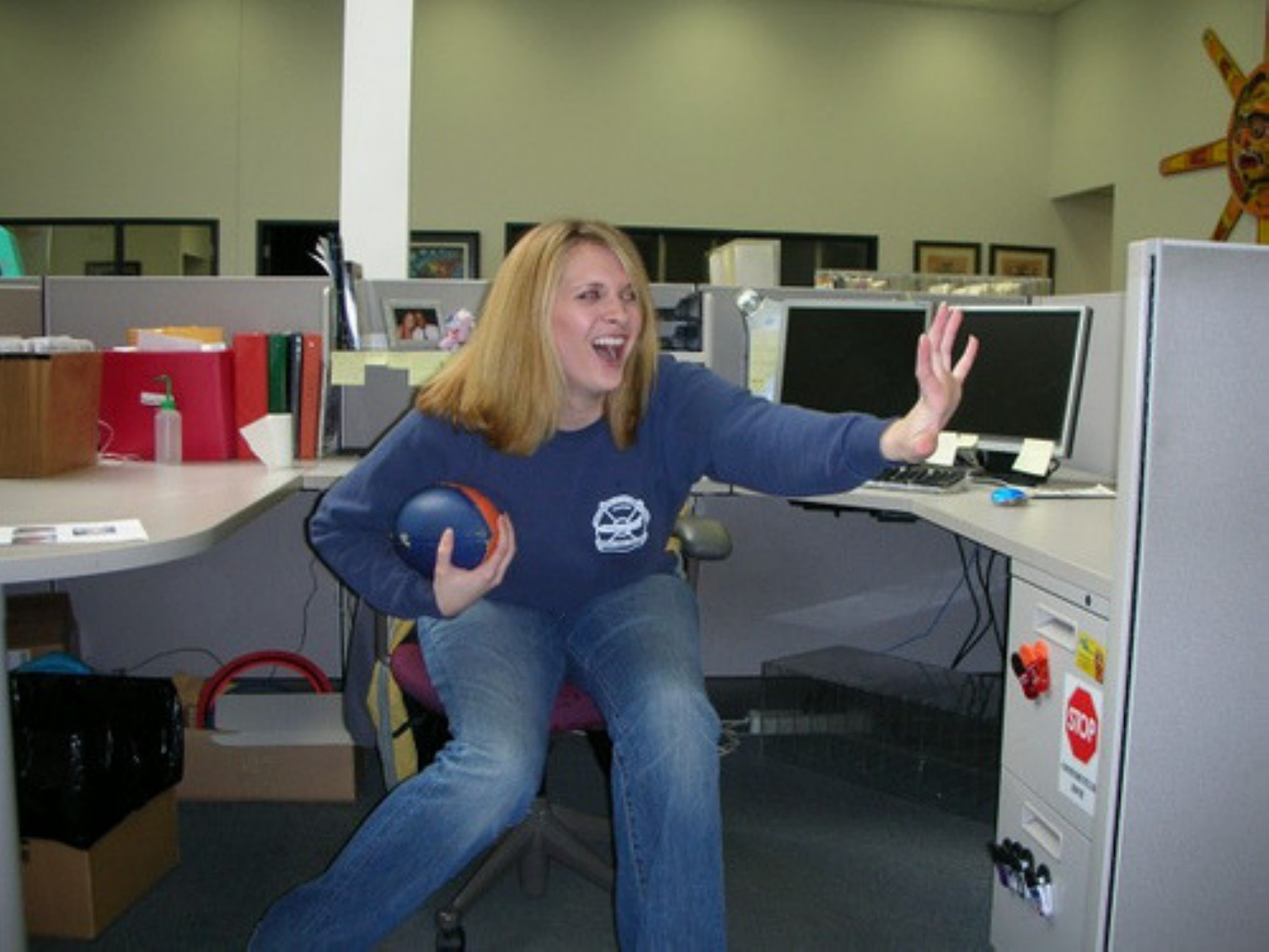}&\includegraphics[width=0.15\textwidth]{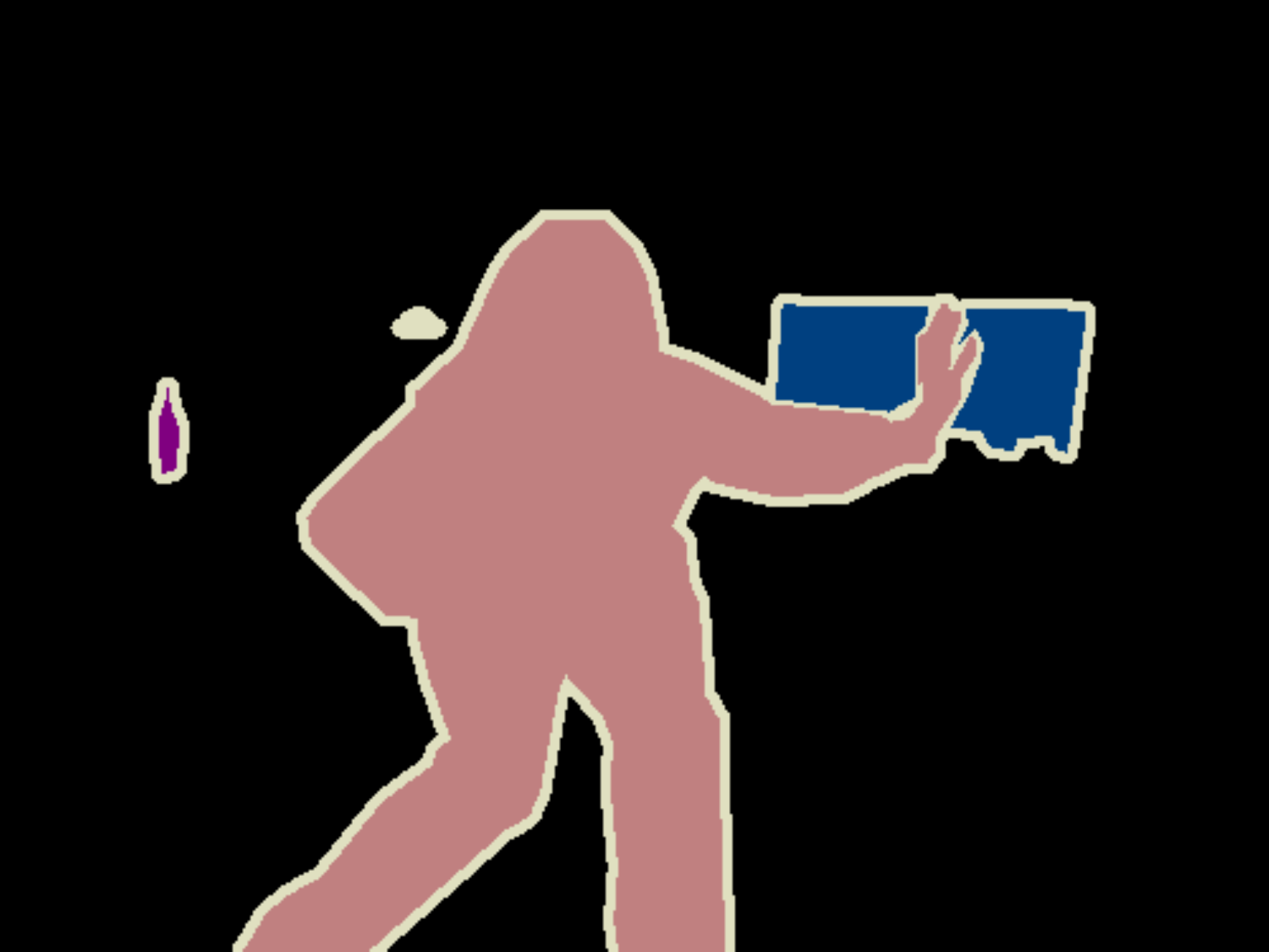}&\includegraphics[width=0.15\textwidth]{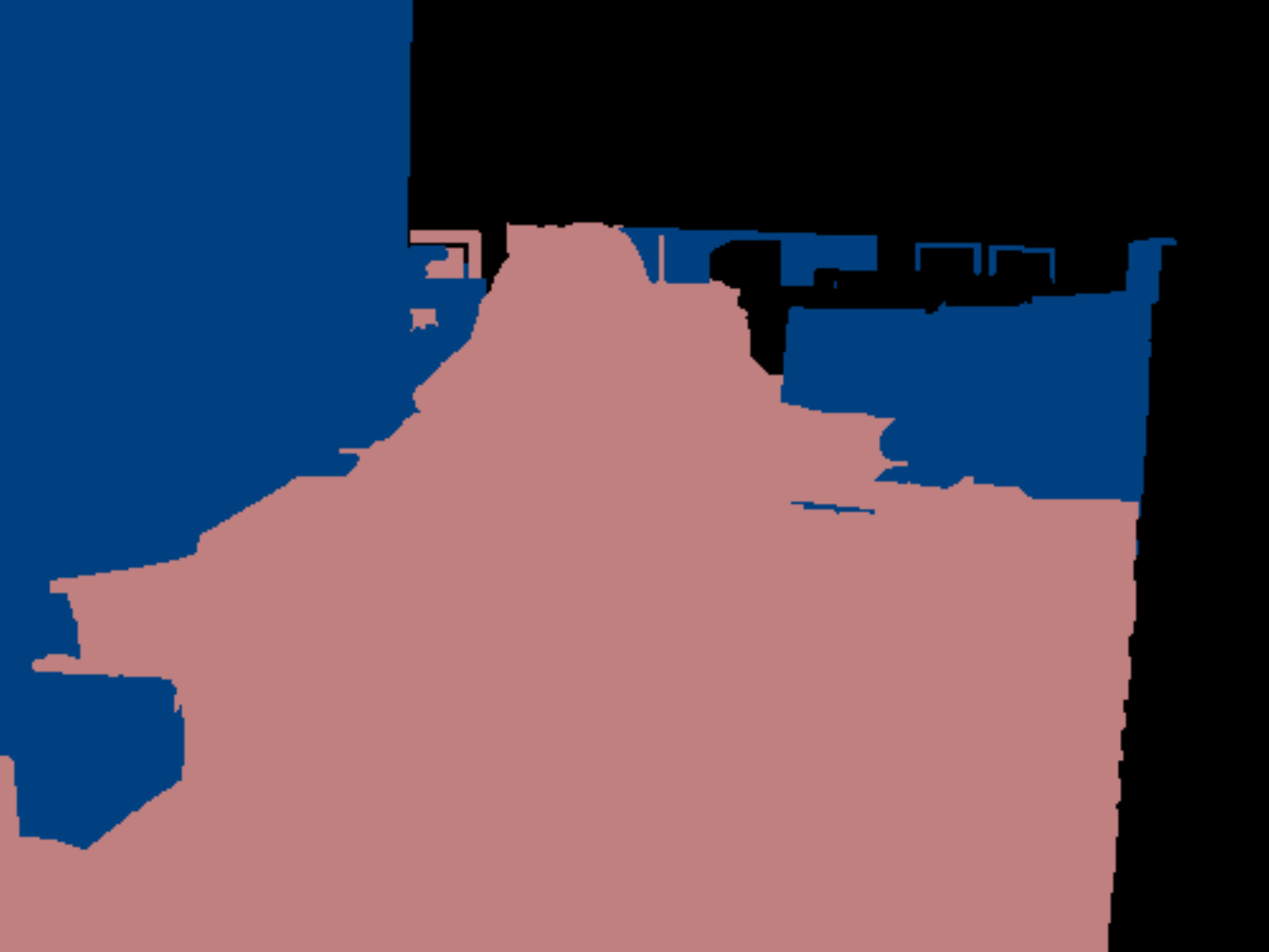}&\includegraphics[width=0.15\textwidth]{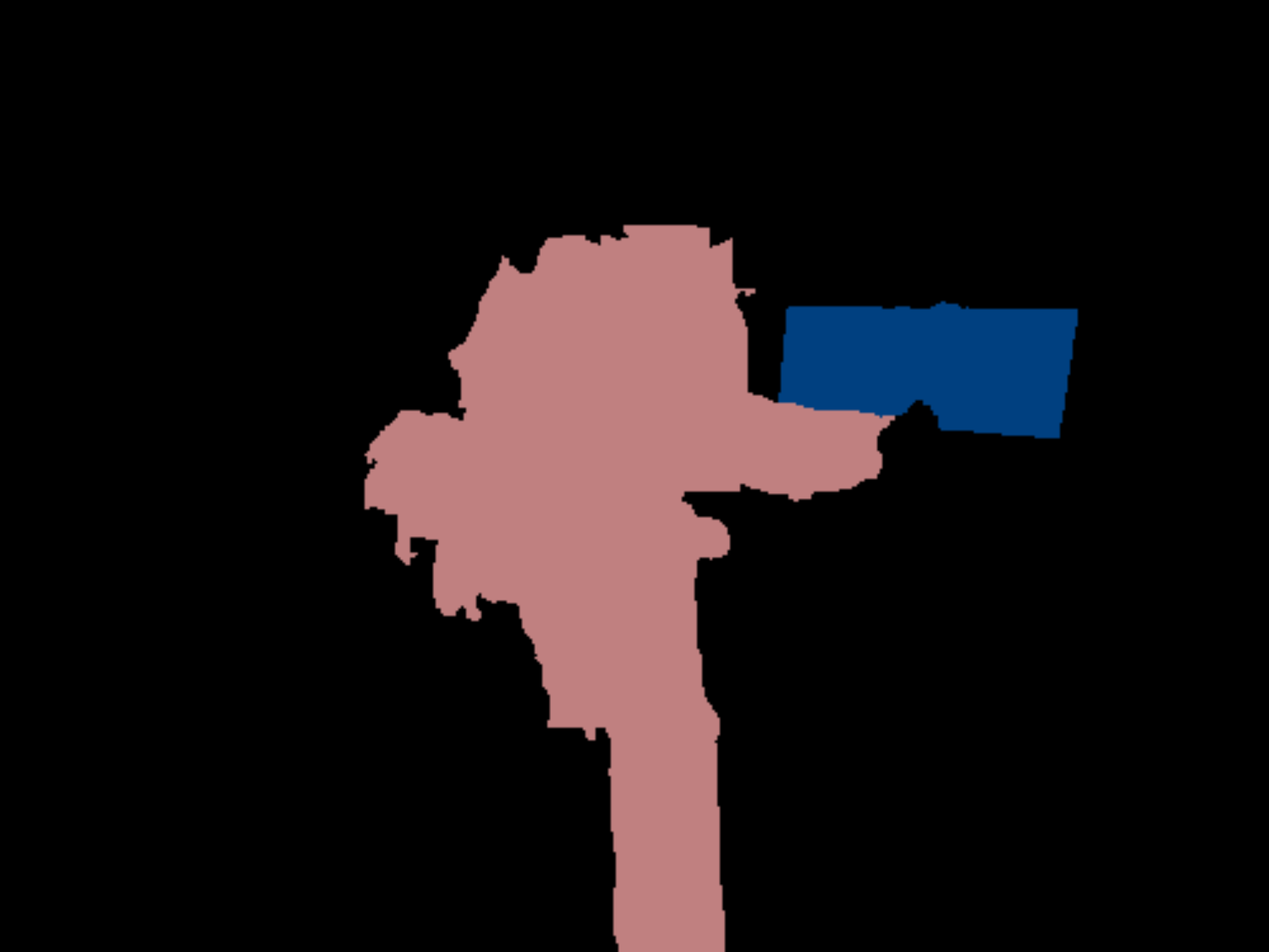}&\includegraphics[width=0.15\textwidth]{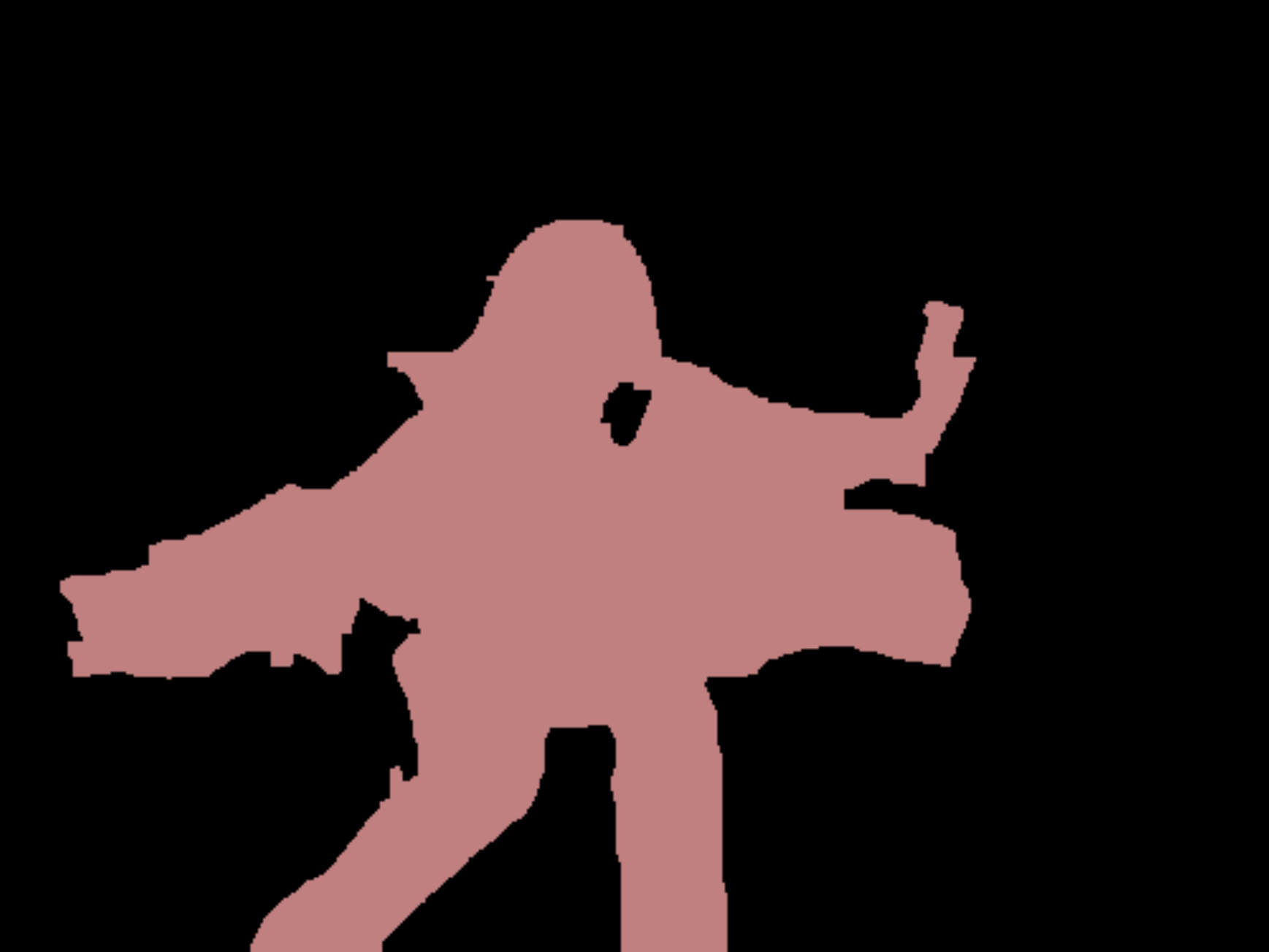}&\includegraphics[width=0.15\textwidth]{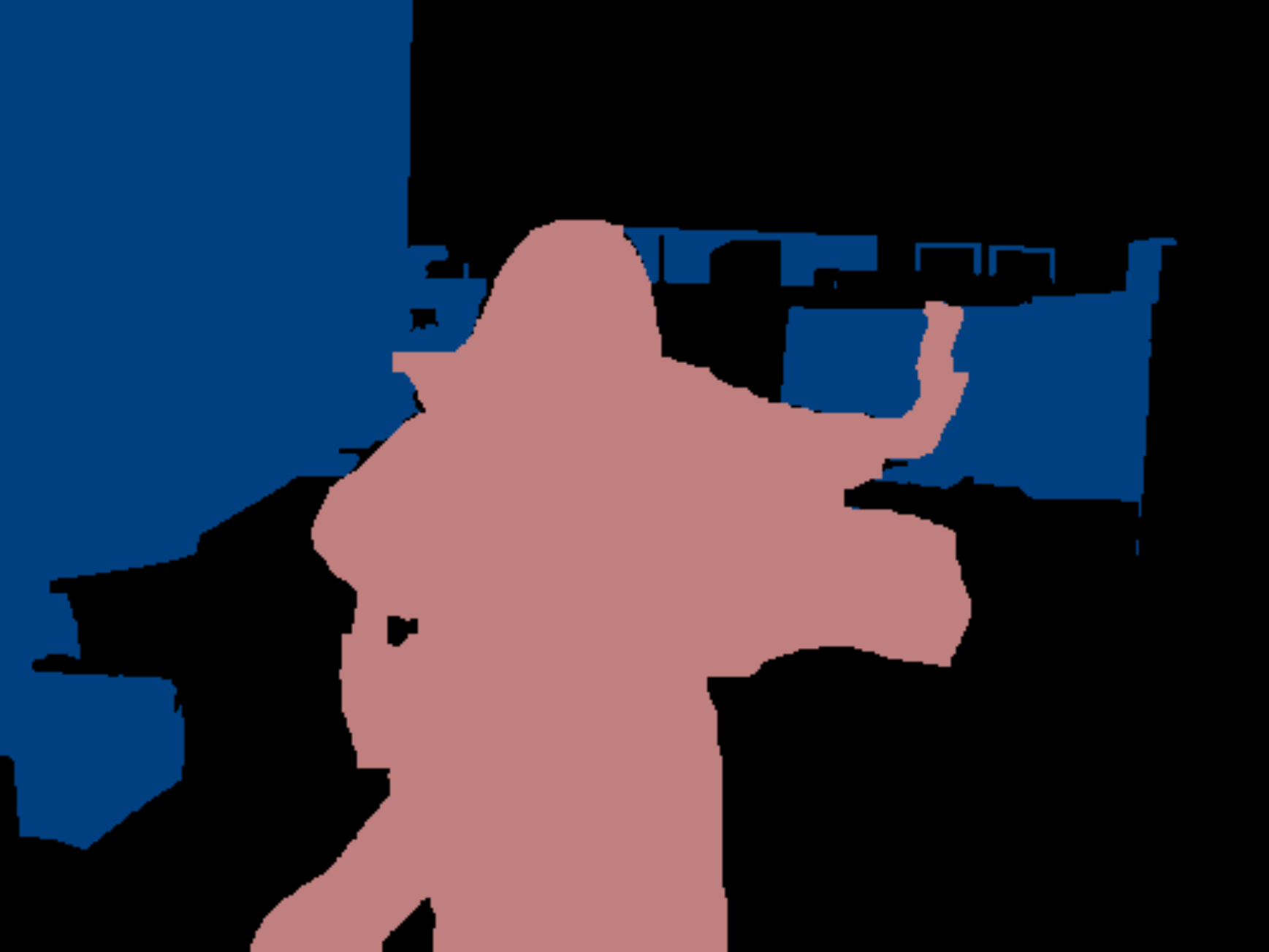}
	\end{tabular}
	\caption{\label{fig:results}Selected results form the IA platform. Each row represents one particular input. Column (1) shows the input image, column (2) shows the human generated ground truth, columns (3)-(5) shows the results of the three available algorithms in order ~\cite{ladicky:10,bharath:14,carreira:12} and last column shows the result obtained by IA platform.  }
\end{figure*}

Some examples of processing are shown in Figure~\ref{fig:results}. Notice that despite the low accuracy a number of images are improved by selecting the regions from each algorithm. 

To see how well the IA approach is performing we compare the average precision of each category of class. Comparison of each algorithm's results is shown in Table~\ref{tab:fval}. As can be seen due to the low level of learning our IA framework outperformed the highest classes precision only in three classes of objects: the boat, bus and dog. For the rest of the categories the IA approach was able to outperform most of the algorithms but one. This is due to the fact that the selection accuracy is relatively low.
\begin{table*}[bht]
\centering
\caption{\label{tab:fval} }
\begin{tabular}{|c|c|c|c|c|}
	\hline
	Class&\multicolumn{4}{c|}{Algorithms}\\
	\hline
      	&IA&CPMC&ALE&COMP6\\
	\hline
	background& 62.554\%&76.430\%&56.020\%&80.248\%\\
	aeroplane& 78.292\%&&61.750\%&78.292\%\\
	bicycle& 27.752\%&13.124\%&27.221\%&32.228\%\\
	bird& 15.190\%&28.443\%&13.318\%&25.932\%\\
	\textbf{boat}& 36.701\%&30.934\%&36.692\%&32.400\%\\
	bottle& 44.317\%&41.224\%&40.131\%&56.265\%\\
	\textbf{bus}& 75.846\%&51.433\%&74.244\%&72.300\%\\
	car& 49.148\%&28.672\%&49.696\%&39.259\%\\
	cat& 60.457\%&58.134\%&64.043\%&58.910\%\\
	chair& 14.664\%&3.885\%&19.565\%&18.576\%\\
	cow& 30.195\%&&31.334\%&2.943\%\\
	diningtable& 38.848\%&25.049\%&38.010\%&54.087\%\\
	\textbf{dog}& 49.950\%&38.775\%&49.949\%&41.761\%\\
	horse& 39.941\%&29.805\%&45.293\%&51.031\%\\
	motorbike& 50.562\%&39.305\%&50.991\%&34.006\%\\
	person& 44.531\%&43.712\%&42.786\%&61.879\%\\
	pottedplant& 22.978\%&22.065\%&27.603\%&38.502\%\\
	sheep& 67.456\%&39.326\%&72.930\%&71.171\%\\
	sofa& 26.792\%&24.301\%&28.751\%&16.612\%\\
	train& 46.152\%&32.323\%&46.195\%&4.945\%\\
	tvmonitor& 32.310\%&46.323\%&29.512\%&62.495\%\\
	\hline
	Average&43.554\%&32.060\%&43.144\%&44.469\%\\
	\hline
\end{tabular}
\end{table*}
Notice that according to the schematic of the IA platform the low accuracy of the algorithm selector could be compensated by a stronger verification and reasoning mechanism. Consider the third row in Figure~\ref{fig:results}. A better reasoning procedure would lead to a result as shown in the hypothetical and ideal case shown in Figure~\ref{fig:exm} rather to the result shown in the last column of the third row in Figure~\ref{fig:results}. The simplest heuristics that would prevent replacing regions directly reducing the f-value could increase the overall result without any significant computational overhead. Similar heuristics for improbable regions removal can also be implemented in parallel to the co-occurrence statistics. Thus even a relatively inaccurate algorithm selection with combined with simple high level verification would lead to better results. 


\section{Conclusion}
\label{sec:con}
In this paper we introduced a soft computing approach to the semantic segmentation problem. The method is based on an algorithm selection platform with the target to increase the quality of the result by reasoning on the content of algorithms outputs. The IA platform for image understanding iteratively improves the high level understanding and even with a very weak algorithm selector can outperform in many cases the best algorithm by combining the best results of each available algorithm. 

In the future several direct extensions and improvements are planned to the IA platform. First the algorithm selection accuracy must be improved. Second the high level verification also requires a more robust method of contradiction detection and hypothesis generation. Co-occurrence statistics are not sufficient because their dependence on the training data. Finally the result merging requires more flexible and robust mechanism in order to avoid decrease in result quality.
{\small
\bibliographystyle{ieee}
\bibliography{../../main}
}

\end{document}